\documentclass[10pt,journal,compsoc]{IEEEtran}
\ifCLASSOPTIONcompsoc
  \usepackage[nocompress]{cite}
\else
  \usepackage{cite}
\fi
\usepackage[pdftex]{graphicx}
\ifCLASSINFOpdf
\else
\fi
\usepackage{algorithm}
\usepackage{algpseudocode}
\usepackage{amsmath}
\usepackage{amssymb}
\usepackage{amsfonts}
\usepackage{multicol}
\usepackage{multirow}
\usepackage{color, graphics}
\usepackage{epsfig}
\usepackage{epstopdf}
\usepackage{balance}
\usepackage{verbatim}

\def\blue{\textcolor{black}}
\def\red{\textcolor{red}}

\def\blue{\textcolor{black}}
\def\red{\textcolor{black}}

\def\magenta{\textcolor{black}}
\def\ie{\emph{i.e.}}
\def\eg{\emph{e.g.}}
\def\yifan{\textcolor{black}}
\def\shuai{\textcolor{black}}

\def\revise{\textcolor{black}}

\usepackage{mathrsfs}
\usepackage{epsfig}

\usepackage{caption}

\begin{document}

\title{Online Adaptive Asymmetric Active Learning with Limited Budgets}

%


\author{Yifan Zhang, Peilin Zhao, Shuaicheng Niu, Qingyao Wu, Jiezhang Cao, Junzhou Huang, Mingkui Tan
\IEEEcompsocitemizethanks{
\IEEEcompsocthanksitem Y. Zhang, S. Niu, J. Cao, Q. Wu and M. Tan are with South China University of Technology, China. E-mail: $\{$sezyifan, sensc, secaojiezhang$\}$@mail.scut.edu.cn; $\{$qyw, mingkuitan$\}$@scut.edu.cn.
\IEEEcompsocthanksitem P. Zhao and J. Huang are with Tencent AI Lab, China. Email: peilinzhao@hotmail.com; joehhuang@tencent.com.
\IEEEcompsocthanksitem P. Zhao, S. Niu and Q. Wu are co-first authors;  Corresponding to M. Tan.}
}

\markboth{IEEE TRANSACTIONS ON KNOWLEDGE AND DATA ENGINEERING}%
{Zhang \MakeLowercase{\textit{et al.}}}

\IEEEtitleabstractindextext{
\begin{abstract}

Online Active Learning (OAL) aims to manage unlabeled datastream by selectively querying the label of data. OAL is applicable to many real-world problems, such as anomaly detection in health-care and finance. In these problems, there are two key challenges: the query budget is often limited; the ratio between classes is highly imbalanced. In practice, it is quite difficult to handle imbalanced unlabeled datastream when only a \revise{limited} budget of labels can be queried for training. To solve this, previous OAL studies adopt either asymmetric losses or queries (an isolated asymmetric strategy) to tackle the imbalance, and use first-order methods to optimize the cost-sensitive measure. However, the isolated strategy limits their performance in class imbalance, while first-order methods restrict their optimization performance. In this paper, we propose a novel Online Adaptive Asymmetric Active learning algorithm, based on a new asymmetric strategy (merging both asymmetric losses and queries strategies), and second-order optimization. We theoretically analyze its mistake bound and cost-sensitive metric bounds. Moreover, to better balance performance and efficiency, we enhance our algorithm via a sketching technique, which significantly accelerates the computational speed with quite slight performance degradation. Promising results demonstrate the effectiveness and efficiency of the proposed methods.
\end{abstract}

}

\maketitle

\IEEEdisplaynontitleabstractindextext

\IEEEpeerreviewmaketitle

\IEEEraisesectionheading{\section{Introduction}\label{Introduction}}
%
%
%
\IEEEPARstart{D}{ue} to rapid growth of data and computational resources, machine learning addresses more and more practical problems, powering many aspects of modern society \cite{Abe2006outlier,hoi2006batch,Moepya2014applying,Zhang2016Online,Zhao2013Costd}. Nevertheless, \yifan{many machine learning methods require the availability of sufficient off-line data before training, while \revise{those off-line samples} are required to be i.i.d. \cite{Dundar2007Learning,Hulten2001Mining}. However, in many real-world applications, data comes in an online manner} and the i.i.d. assumption may not hold. To address these limitations, online learning has emerged as a powerful tool \cite{Freund1999Large,Crammer2006Online,Cesa-Bianchi2005A,wang2014cost}. \revise{It} makes no assumptions about the distribution of data and thus is data efficient and adaptable~\cite{Dundar2007Learning,Hulten2001Mining}.

Most existing online methods assume all samples are labeled, \revise{and ignore} the labeling cost as well as the budget control problem. However, in many applications \revise{like} medical diagnosis \cite{hoi2006batch} and malicious URL detection \cite{Zhao2013Costd,Zhao2013Cost}, the cost of annotation is often expensive. \revise{Hence}, \yifan{it is important to find out samples, which deserve to be labeled from data streams}. To handle this task, online active learning (OAL) \cite{Sheng2008Get,Cesa-Bianchi2006Worst} has emerged. \revise{It aims to train a well-performed model by selectively querying only a small number of labels for data streams}. Many studies \cite{Sheng2008Get,Cesa-Bianchi2006Worst,Zhang2016Online} have found that different query rules result in very different performance, which means that \revise{the query strategy} is very important. \revise{Meanwhile}, real-world companies usually expect to spend as few funds as possible for data annotation. In other words, we only have a limited budget for label querying. Given \revise{a} limited budget, we have to select the most informative samples \revise{to query} so that they can help to train a \revise{well-performed} model.

In addition, the class-imbalanced issue seriously affects algorithm performance in real-world applications, such as cancer diagnosis \cite{hoi2006batch}, financial credit monitoring \cite{Moepya2014applying} and network fraud detection \cite{Zhao2013Costd}. Existing OAL methods usually \revise{train models} using balanced \emph{accuracy} or \emph{mistake rate} as metrics. However, they \yifan{cannot handle} the imbalance issue well \cite{wang2014cost}. \revise{To solve this}, researchers have proposed more informative metrics, such as the weighted \emph{sum} of \emph{sensitivity} and \emph{specificity}, and the weighted \emph{misclassification cost} \cite{wang2014cost}.

Based on these metrics,  a pioneering cost-sensitive online active learning algorithm (CSOAL) \cite{Zhao2013Costd} was proposed to directly optimize asymmetrically cost-sensitive metrics for OAL. However, this method only adopts a symmetric query rule \cite{Cesa-Bianchi2006Worst} \yifan{and \revise{ignores}} the imbalance problem in data selection. \yifan{Recently}, online asymmetric active learning algorithm (OAAL) \cite{Zhang2016Online} discovered that using asymmetric query strategy helps to handle imbalanced data better. However, this method overlooks imbalance issues in the optimization process and tends to query more majority data due to the recommended parameter settings \cite{Zhang2016Online}. \revise{Hence, this method may lead to poor performance on minority data.} In comparison, CSOAL is ``asymmetric update plus symmetric query", and OAAL is ``symmetric update plus asymmetric query". Both algorithms \revise{devise the asymmetric strategy from a different and isolated perspective, and thus may perform insufficiently  in class-imbalance problems.}

\revise{In addition, both algorithms only consider} first-order information of data streams. \revise{However, when scales of different features vary significantly, these methods} may converge slowly \cite{Cesa-Bianchi2005A}. As a result, it is difficult for them to achieve \revise{a good  solution} when labeled data is quite limited. \revise{To deal with this issue}, recent studies \cite{Crammer2009Adaptive,zhao2018cost} have found second-order information (\ie,  correlations between features) helps to enhance  online methods significantly.

In this paper, we propose a novel online adaptive asymmetric active (\textbf{OA3}) learning algorithm.  \revise{By exploiting samples' second-order information, we develop a new asymmetric strategy, which considers both model optimization and label queries simultaneously. As a result, the proposed strategy addresses the class imbalance better and thus improves model performance.} \yifan{Moreover, we theoretically analyze the metric bounds of our proposed algorithm for the cases within budgets and over budgets, respectively.}

\revise{Although enjoying the advantage of second-order information, our proposed algorithm may run slower than first-order methods, because \yifan{the update} of the correlation matrix is time-consuming.} \revise{Therefore}, it may be inappropriate for applications with quite high-dimensional datasets. \revise{To address this issue, we further propose two efficient variants of OA3} based on sketching techniques\cite{luo2016efficient,woodruff2014sketching,Krummenacher2016Scalable,Wang2014High}.

\yifan{We empirically evaluate the proposed methods} on real-world datasets. Encouraging results confirm their effectiveness, efficiency and \yifan{stability}. We also examine the influences of algorithm parameters. Extensive results validate the algorithm characteristics.

The rest of this paper is organized as follows. We first present the problem formulation and the proposed algorithm in Section 2, \yifan{followed by} the theoretical analyses in Section 3.
Next, we propose two efficient variants based on sketching techniques in Section 4. \yifan{We empirically evaluate the proposed algorithms in Section 5,
and conclude the paper in Section 6.} Due to the page limitation, we \yifan{put related work} in Supplementary D.

%
%
\vspace{-0.15in}
\section{Setup and Algorithm}
\magenta{In this section, we firstly introduce the \red{problem formulation of the online active learning problem for budgeted imbalanced data}. 
Next, we present the scheme of the proposed Online Adaptive Asymmetric Active (OA3) Learning algorithm. Lastly, relying on samples' second-order information, we propose a new asymmetric strategy, which consists of an asymmetric update rule and an asymmetric query rule.}
\subsection{Problem Formulation}
\yifan{Without loss of generality, we consider online binary classification under limited query budgets here. Streaming data comes in one by one $\{x_1,x_2,...,x_T\}$, where $x_t \small{\in} \mathbb{R}^d$ is a $d$-dimensional sample at time $t$, and $T$ is the total quantity of samples. Note that all samples are unlabeled, and there is a limited query budget $B$ for obtaining class label $y \small{\in} \{-1,1\}$. The main task is to learn a well-performed online linear classifier $w \in \mathbb{R}^d$ with only limited labeled data. Moreover, the prediction of the classifier is $\hat{y}={\rm sign(w^{\top}x)}$}.

Primarily, we define some notations: $\mathcal{M} = \{t \ | y_t \neq {\rm sign}(w_t^{\top} x_t), \forall t \in [T] \}$ is the mistake index set, $\mathcal{M}_p = \{t \in \mathcal{M}$ and $y_t=+1\}$ is the positive set of mistake index and $\mathcal{M}_n = \{t \in \mathcal{M}$ and $y_t=-1\}$ is the negative one. In addition, we set $M=|\mathcal{M}|$, $M_p= |\mathcal{M}_p|$ and $M_n= |\mathcal{M}_n|$ to denote the number of total mistakes, positive mistakes and negative mistakes. Moreover, we denote the index sets of all positive samples and all negative samples by $\mathcal{I}^p_T = \{i \in [T]| y_i = +1 \}$ and $\mathcal{I}^n_T = \{i \in [T]| y_i = -1 \}$, where $T_p = |\mathcal{I}^p_T|$ and $T_n = |\mathcal{I}^n_T|$ denote the number of positive samples and negative samples. \blue{For convenience}, we assume the positive class as the minority class, \ie, $T_p\leq T_n$.

Traditional online algorithms often optimize \emph{accuracy} or \emph{mistake rate}, which treats samples from different class equally. These metrics, however, may be inappropriate for imbalanced data, since a trivial learner by simply classifying all data as negative can still achieve high accuracy. To address this issue, a more suitable metric is to measure the \emph{sum} of weighted \emph{sensitivity} and \emph{specificity}, \ie,
\begin{equation}\label{sum_metric}
  sum =\alpha_p \times \frac{T_p-M_p}{T_p}+ \alpha_n \times \frac{T_n- M_n}{T_n}, 
\end{equation}
where $\alpha_p, \alpha_n \in [0,1]$ are trade-off parameters between \emph{sensitivity} and \emph{specificity}, and $\alpha_p+\alpha_n=1$. Note that when $\alpha_p=\alpha_n=0.5$, the $sum$ metric becomes the famous \emph{balanced accuracy}. In general, the higher the \emph{sum} value, the better the classification performance.

In addition, another metric is to measure the weighted misclassification \emph{cost} suffered by the model, \ie,
\begin{equation}\label{cost_metric}
  cost =c_p \times M_p+ c_n \times M_n, 
\end{equation}
where $c_p, c_n \in [0,1]$ denote the cost weights for positive and negative instances, and $c_p+c_n=1$. The lower the \emph{cost} value, the better the classification performance.

\subsection{Algorithm Scheme}

Inspired by the adaptive confidence weight technique \yifan{\cite{Dredze2008Confidence,Crammer2009Adaptive}, we exploit samples' second order information. Assume that the online model satisfies a multivariate Gaussian distribution \cite{Crammer2009Adaptive}, \ie, $w \sim \mathcal{N}(\mu, \Sigma)$, where $\mu$ is the mean vector and $\Sigma$ is the covariance matrix of distribution. Without loss of generality, the mean value $\mu_i$ represents the model's knowledge of the weight for feature $i$, while $\Sigma_{i,i}$ encodes the confidence of feature $i$. Generally, the smaller $\Sigma_{i,i}$, the more confidence the model has in the mean weight value $\mu_i$. The covariance term $\Sigma_{i,j}$ keeps the correlations between weights $i$ and $j$. \yifan{Given a definite Gaussian distribution, it is more practical to simply use the mean vector $\mu=\mathbb{E}[w]$ to make predictions, \ie, $\hat{y}= {\rm sign}(\mu^{\top} x)$} \cite{zhao2015cost,Crammer2009Adaptive,zhao2018cost}, where we denote $p=\mu^{\top}x$ as the predictive margin.}

Formally, \yifan{when receiving a new sample $x_t$ in the $t$-th round, the \revise{learner} needs to make a prediction $\hat{y}_t$ and decide \revise{whether} to query the true label $y_t$. If deciding to query, the learner will consume one unit budget, and then update the predictive vector $\mu_t$ based on the received painful loss from $(x_t,y_t)$. Otherwise, the model will ignore $x_t$.  Note that the above process is performed within the limited query budget $B$. Once the available budget goes down to zero, the learner will stop querying true labels and thus stop updating. We summarize the algorithm scheme in Algo.~\ref{al:oa3_scheme}.}

\begin{algorithm}
    \caption{Online Adaptive Asymmetric Active (OA3) Learning algorithm.}\label{al:oa3_scheme}
    \begin{algorithmic}[1]
        \Require budget $B$; learning rate $\eta$; regular parameter $\gamma$.
        \Ensure $\mu_1=0$, $\Sigma_1=I$, $B_1=0$.
        \For{$t = 1 \to T$}
            \State Receive an example $x_t \in \mathbb{R}^d;$
            \State Compute $p_t = \mu_t^{\top}x_t;$
            \State Make the prediction $\hat{y}_t =$ sign$(p_t);$
            \State Draw a variable $Z_t = \red{\textbf{Query}}(p_t) \in \{0,1\};$
            \If {$Z_t = 1$ and $B_t < B$}
                \State Query the true label $y_t \in \{-1,+1\};$
                \State $B_{t+1} = B_t +1;$
                \State $\mu_{t+1}, \Sigma_{t+1}  = \red{\textbf{Update}}(\mu_{t},\Sigma_{t};x_t,y_t).$
            \Else
                \State $  B_{t+1} = B_t , \mu_{t+1} = \mu_{t}, \Sigma_{t+1}= \Sigma_{t}.$
            \EndIf
        \EndFor
    \end{algorithmic}
\end{algorithm}

\yifan{Considering the class-imbalanced issue, there are two main challenges when designing this active algorithm.}

1) \textbf{How to update} the model in an asymmetric way to obtain \revise{a well-performed model}, which is described in \blue{Subsection 2.3.}

2) \textbf{How to query} the most informative samples asymmetrically, which is described in \blue{Subsection 2.4.}

\subsection{Adaptive Asymmetric Update Rule}

\revise{To solve the class imbalance issue}, our objective is  either  to maximize \emph{sum} \yifan{ metric in Eq.~(\ref{sum_metric})} or to minimize \emph{cost} \yifan{metric in Eq.~(\ref{cost_metric})}. Both objectives are equivalent to minimizing the following objective~\cite{wang2014cost}:
\begin{equation}
  \sum_{y_t=+1}\rho \mathbb{I}_{(y_t (\mu^{\top} x_t) <0)} +\sum_{y_t=-1} \mathbb{I}_{(y_t (\mu^{\top}x_t) <0)},
\end{equation}
where $\rho = \frac{\alpha_p T_n }{\alpha_n T_p}$ for maximizing \emph{sum} metric and $\rho = \frac{c_p}{c_n}$ for minimizing \emph{cost} metric, while $\mathbb{I}_{(\cdot)}$ is the indicator function.

However, this objective is non-convex. To facilitate the optimization, we replace the indicator function with its convex variants, \ie,
\begin{equation}\label{loss}
  \ell_t \big{(}\mu\big{)}= \big{(}\rho  \mathbb{I}_{(y=+1)}\small{+}\mathbb{I}_{(y=-1)}\big{)}  {\rm max}\big{\{}0, 1\small{-}y_t(\mu^{\top} x_t)\big{\}}.
\end{equation}

At round $t$, when \yifan{receiving a sample $x_t$ and querying its label $y_t$, we can naturally exploit second-order information of data by minimizing the following objective~ \cite{Crammer2009Adaptive,zhao2018cost}, \ie,}
\begin{equation}\label{eq:objective}
  D_{KL}(\mathcal{N}(\mu, \Sigma)||\mathcal{N}(\mu_t, \Sigma_t)) + \eta \ell_t(\mu) + \frac{1}{2\gamma}x_t^{\top}\Sigma x_t, 
\end{equation}
where $\eta$ is the learning rate, $\gamma$ is the regularized parameter and $D_{KL}$ denotes the Kullback-divergence, \ie,
\begin{align}
  D_{KL}\big{(}&\mathcal{N}(\mu, \Sigma)||\mathcal{N}(\mu_t, \Sigma_t)\big{)}    \nonumber  \\
  =&\frac{1}{2}{\rm log}\Big{(}\frac{{\rm det}\Sigma_t}{{\rm det}\Sigma}\Big{)}+ \frac{1}{2}{\rm Tr}(\Sigma^{-1}_t \Sigma)+\frac{1}{2}||\mu_t-\mu||_{\Sigma^{-1}_t}^2 -\frac{d}{2}.  \nonumber
\end{align}
\blue{Specifically}, 
the objective of Eq.~(\ref{eq:objective}) helps to reach the trade-off between distribution divergence (first term), loss function (second term) and model confidence (third term). In other words, the objective tends to make the least adjustment to minimize the painful loss and optimize the model confidence. However, this optimization does not have \blue{closed-form} solution. To address this issue, we replace the loss $\ell(\mu)$ with its first order Taylor expansion $\ell(\mu_t) + g_t^{\top}(\mu - \mu_t)$, where $g_t=\partial \ell_t(\mu_t)$. We then obtain the final optimization objective by removing constant terms, \ie,
\begin{equation}\label{5}
  f_t(\mu,\Sigma) = D_{KL}(\mathcal{N}(\mu, \Sigma)||\mathcal{N}(\mu_t, \Sigma_t)) \small{+} \eta g_t^{\top}\mu  \small{+} \frac{1}{2\gamma}x_t^{\top}\Sigma x_t,
\end{equation}
which is much easier to be solved. We solve this optimization problem in two steps, \yifan{iteratively}:

$\bullet$ \hspace{2ex} Update the mean parameter:
\begin{equation}
  \mu_{t+1} = {\rm arg} \ \min_{\mu} f_t(\mu,\Sigma);   \nonumber
\end{equation}

$\bullet$ \hspace{2ex} If $\ell_t(\mu_t) \neq 0$, update the covariance matrix:
\begin{equation}
  \Sigma_{t+1} = {\rm arg} \  \min_{\Sigma} f_t(\mu,\Sigma).   \nonumber
\end{equation}

For the first step, setting the derivative $\partial_{\mu}f_t(\mu_{t+1},\Sigma)$ to zero will give:
\begin{align}
  \Sigma_t^{-1}(\mu_{t+1}-\mu_t) + \eta g_t=0 \hspace{1ex} \Longrightarrow \hspace{1ex} \mu_{t+1} = \mu_t- \eta \Sigma_t g_t.  \nonumber
\end{align}
Second, setting derivative $\partial_{\Sigma}f_t(\mu,\Sigma_{t+1})$ to zero gives:
\begin{align}\label{eq:inverse_sigma}
  -\Sigma_{t+1}^{-1}+\Sigma_{t}^{-1}+\frac{x_tx_t^{\top}}{\gamma}=0 \hspace{1ex} \Longrightarrow \hspace{1ex}  \Sigma_{t+1}^{-1} = \Sigma_{t}^{-1}+\frac{x_tx_t^{\top}}{\gamma}. 
\end{align}
Based on the Woodbury identity \cite{Horn1985matrix}, we have:
\begin{align}\label{eq:sigma_t+1}
  \Sigma_{t+1} = \Sigma_{t} - \frac{\Sigma_{t}x_tx_t^\top\Sigma_{t}}{\gamma+x_t^\top\Sigma_{t}x_t}.
\end{align}

Apparently, the update of the predictive vector relies on the \blue{confidence $\Sigma$}. Thus, we  update the mean vector $\mu_t$ based on the updated covariance matrix $\Sigma_{t+1}$, which will be more accurate and aggressive \cite{zhao2015cost,zhao2018cost}, \ie,
\begin{align}\label{eq:mu_t+1}
  \mu_{t+1} = \mu_t- \eta \Sigma_{t+1} g_t.
\end{align}

We summarize the adaptive asymmetric update strategy in Algo.~\ref{al:oa3_update}.

\begin{algorithm}
    \caption{Adaptive Asymmetric Update Strategy: \hspace{3ex} \textbf{\red{\textbf{Update}}$(\mu_{t},\Sigma_{t};x_t,y_t)$}.} \label{al:oa3_update}
    \begin{algorithmic}[1]
        \Require $\rho=\frac{\alpha_pT_n}{\alpha_nT_p}$ for $sum$ or $\rho=\frac{c_p}{c_n}$ for $cost$;
            \State Receive a sample $(x_t,y_t);$
            \State Compute the loss $\ell_t(\mu_t)$, based on Equation (4)$;$
            \If {$\ell_t(\mu_t) > 0$}
                \State $\Sigma_{t+1} = \Sigma_{t} - \frac{\Sigma_{t}x_tx_t^\top\Sigma_{t}}{\gamma+x_t^\top\Sigma_{t}x_t};$
                \State $\mu_{t+1} = \mu_{t} - \eta\Sigma_{t+1}g_t,$   where $ g_t=\partial\ell_t(\mu_t).$
            \Else
                \State $\mu_{t+1} = \mu_{t}, \Sigma_{t+1}= \Sigma_{t}.$
            \EndIf
         \State Return \hspace{0.5ex} $\mu_{t+1}, \Sigma_{t+1}.$
    \end{algorithmic}
\end{algorithm}

\textbf{\blue{Time Complexity Analysis.}} \yifan{Time complexities of updating for} $\mu$ and $\Sigma$ are both $\mathcal{O}(T d^2)$, so the overall time complexity of this update strategy is $\mathcal{O}(T d^2)$, where $d$ is the dimension of data. Nevertheless, the update efficiency of OA3 is slower than first-order algorithms \big{(}$\mathcal{O}(Td)$\big{)}, especially when handling high-dimensional datasets. To promote the efficiency, we propose a \textbf{diagonal version}  of the update strategy (the pseudo-code is put in Supplementary B.1), which accelerates the efficiency to $\mathcal{O}(Td)$. Specifically, in this \revise{variant}, only the diagonal entries of $\Sigma$ are maintained and updated in each round.

\textbf{Remark.} \  \emph{We employ the adaptive asymmetric update rule for OA3 to pursue high performance with faster convergence. Nevertheless, it is not the only choice. Other classic techniques can also be used, \eg, online gradient descent \cite{wang2014cost} and online margin-based strategies \cite{Crammer2006Online,Cesa-Bianchi2005A}}.

\subsection{Asymmetric Query Strategy}
In a pioneering study \cite{Cesa-Bianchi2006Worst}, one classic sampling method was proposed to query  labels based on a Bernoulli random variable $Z_t \in \{0,1\}$. \revise{That is,
querying the label for a given instance} only when $Z_t = 1$.
\revise{To be specific}, the probability of sampling $Z_t$ depends on the absolute value of the predictive margin $|p_t|$, \ie,
\begin{equation}
    \text{Pr}\big{(}Z_t=1\big{)} = \frac{\delta}{\delta + |p_t|}, \nonumber
\end{equation}
where $\delta > 0$ is the query bias. \revise{In this equation, when the absolute predictive margin $|p_t|$ is low, the probability to query the label of sample $x_t$ is relatively high. This is because when the sample's prediction is close to the classification hyperplane (small $|p_t|$), the sample is difficult to be classified and hence is more valuable to be queried (high $\text{Pr}(Z_t=1)$).}

However, this query rule ignores the imbalance of data and treats predictions of two imbalanced classes equally. To address this limitation, inspired by recent work \cite{Zhang2016Online}, we employ an asymmetric strategy to query labels, \ie,
$$\text{Pr}\big{(}Z_t=1\big{)}=
\begin{cases}
    \frac{\delta_{+}}{\delta_{+}+|p_t|}, & \text{if $p_t \geq 0$}; \\
    \frac{\delta_{-}}{\delta_{-}+|p_t|}, & \text{if $p_t < 0$};
\end{cases}$$
where $\delta_+ > 0$ and $\delta_- > 0$ denote query biases for positive and negative predictions, respectively.

However, \blue{this} asymmetric query strategy heavily depends on the absolute value of margin $p_t$, which is directly calculated by the model $\mu_t$. As a result, the query decisions may be inaccurate when $\mu_t$ is not precise enough~\cite{Hao2016second}.

To address this issue, \yifan{we use samples' second order information to enhance the query strategy and improve the robustness of the query judgement}. We first define the variance of a model on the sample $x_t$ as $v_t = x_t^{\top}\Sigma_t x_t$. It represents the familiarity of the model with the current sample through previous experience. Based on $v_t$, we then define the query confidence:
\begin{equation}\label{eq:c_t}
  c_t = - \frac{1}{2} \frac{\eta\rho_{max}}{\frac{1}{v_t}+\frac{1}{\gamma}},
\end{equation}
where $\rho_{max} = \max \{1,\rho \}$. We highlight that this equation is helpful for the theoretical analysis. Moreover, the confidence $c_t$ directly depends on the variance $v_t$. \yifan{Based on this equation}, when the model has been well trained on some instances similar to the current sample $x_t$ (\ie, low variance $v_t$), the model would be confident of this sample (\ie, large confidence $c_t$).

\revise{Based on} the predictive margin and the confidence, we obtain the final query parameter:
\begin{equation}\label{eq:q_t}
    q_t = |p_t|+c_t.
\end{equation}
\yifan{Moreover, both} the learning rate $\eta$ and regularized parameter $\gamma$ in Eq.~(\ref{eq:c_t}) can be understood as trade-off factors in Eq.~(\ref{eq:q_t}).

Relying on above analyses, we propose an improved asymmetric query strategy:
\begin{equation}
\text{Pr}\big{(}Z_t=1\big{)}=
\begin{cases}
    \frac{\delta_{+}}{\delta_{+}+q_t}, & \text{if $p_t \geq 0$}; \\
    \frac{\delta_{-}}{\delta_{-}+q_t}, & \text{if $p_t < 0$}.
\end{cases}  \nonumber
\end{equation}

\revise{To be specific}, when $q_t \small{>}0$, the query \revise{decision} of the model is very confident, so we directly draw a Bernoulli variable based on this equation. If $q_t\small{\leq} 0$, the query decision is unconfident of the current sample, so we decide to query the true label whatever the value of $p_t$, \ie, obtaining $Z_t\small{=}1$ by setting $q_t \small{=} 0$ (see the above equation).

We summarize the proposed asymmetric query strategy in Algo.~\ref{al:oa3_query}.
\begin{algorithm}
    \caption{Asymmetric Query Strategy: \textbf{\red{\textbf{Query}}$(p_t)$}.}\label{al:oa3_query}
    \begin{algorithmic}[1]
    \Require $\rho_{max}=\max\{1,\rho\}$; query bias $(\delta_{+}\small{,} \delta_{-})$ for positive and negative predictions.
        \State Compute the variance $v_t= x_t^{\top} \Sigma_t x_t;$
        \State Compute the query parameter $q_t=|p_t|- \frac{1}{2} \frac{\eta\rho_{max}}{\frac{1}{v_t}+\frac{1}{\gamma}};$
        \If {$ q_t \leq 0$}
            \State Set $q_t=0;$
        \EndIf

        \If {$ p_t \geq 0$}
            \State $p_t^{+} = \frac{\delta_{+}}{\delta_{+}+q_t};$
            \State Draw a Bernoulli variable $Z_t \small{\in} \{0,1\}$ with $p_t^{+}.$
        \Else
            \State $p_t^{-} = \frac{\delta_{-}}{\delta_{-}+q_t};$
            \State Draw a Bernoulli variable $Z_t \small{\in} \{0,1\}$ with $p_t^{-}.$
        \EndIf

        \State Return \hspace{0.5ex} $Z_t.$
    \end{algorithmic}
\end{algorithm}

We can obtain the expected number of queried samples without budget limitations as follows.

\textbf{Proposition 1.} Based on the proposed asymmetric query strategy, the expected number of requested samples without a budget is:
\begin{equation}
 \sum \mathbb{I}_{(q_t \leq 0)}  + \sum_{\substack{q_t > 0\\ p_t \geq 0}}\frac{\delta_{+}}{\delta_{+}+q_t}+ \sum_{\substack{q_t > 0\\ p_t < 0}}\frac{\delta_{-}}{\delta_{-}+q_t}.    \nonumber
\end{equation}

\section{Theoretical Analysis}

We next analyze the proposed algorithm in terms of its mistake bound and two cost-sensitive metric bounds, for the cases within budgets and over budgets, respectively. Before that, we first show a lemma, which facilitates the analysis within budgets. Due to the page limitation, all proofs are put in \blue{Supplementary A}.

For convenience, we introduce the following notations:
\begin{align}
   &M_t  = \mathbb{I}_{(\hat{y}_t\neq y_t)}, \ \rho = \frac{\alpha_p T_n}{\alpha_n T_p} \ {\rm or} \ \frac{c_p}{c_n}, \nonumber  \\
    \rho_t \small{=} \rho \mathbb{I}_{(y_t =+1)}\small{+}&\mathbb{I}_{(y_t =-1)} ,  \rho_{max}\small{=}\max\{1,\rho\},   \rho_{min}\small{=}\min\{1,\rho\}.  \nonumber
\end{align}

\textbf{Lemma 1.} \ \emph{Let $(x_1,y_1),...,(x_T,y_T)$ be a sequence of input samples, where $x_t \in \mathbb{R}^{d}$ and $y_t \in \{-1,+1\}$ for all t. Let $T_B$ be the round that runs out of the budgets, \ie,  $B_{T_{B}+1}=B$. For any $\mu \in \mathbb{R}^{d}$ and any $\delta>0$, OA3 algorithm satisfies:}
\begin{align}
   \sum_{t=1}^{T_B} M_t  Z_t(\delta \small{+} q_t) \small{\leq} \frac{\delta}{\rho_{min}}\sum_{t=1}^{T_B} \ell_t(\mu)\small{+} & \frac{1}{\eta \rho_{min}}\emph{Tr}(\Sigma_{T_{B}+1}^{-1})\times    \nonumber \\
   & \big{[}M(\mu) +(1-\delta)^2 ||\mu||^2 \big{]}, \nonumber
\end{align}
where $M(\mu)= \max_t||\mu_t \small{-}\mu||^2$.

Based on Lemma 1, we obtain the following three theorems for the case \textbf{within budgets}.

\textbf{Theorem 1.} \ \emph{Let $(x_1,y_1),...,(x_T,y_T)$ be a sequence of input samples, where $x_t \in \mathbb{R}^{d}$ and $y_t \in \{-1,+1\}$ for all t. Let $T_B$ be the round that runs out of the budgets, \ie,  $B_{T_{B}+1}=B$. For any $\mu \in \mathbb{R}^d$, the expected mistake number of OA3 within budgets is bounded by:}
\begin{align}
    \mathbb{E}\bigg{[}\sum_{t=1}^{T_B} M_t\bigg{]} &= \mathbb{E}\bigg{[}\sum_{\substack{t=1\\ y_t = +1}}^{T_B}M_t+\sum_{\substack{t=1\\ y_t = -1}}^{T_B}M_t\bigg{]}    \nonumber \\
    & \leq \frac{1}{\rho_{min}}\bigg{[}\sum_{t=1}^{T_B}\ell_t(\mu) + \frac{1}{\eta} D(\mu)  \emph{Tr}(\Sigma_{T_{B}+1}^{-1})\bigg{]},  \nonumber
\end{align}
where $D(\mu)\small{=} \max \Big{\{}\frac{M(\mu)\small{+} (1\small{-}\delta_{+})^2 ||\mu||^2}{\delta_{+}}, \frac{M(\mu)\small{+} (1\small{-}\delta_{-})^2 ||\mu||^2}{ \delta_{-}} \Big{\}}$.

This mistake bound helps to analyze the weighted $sum$ performance under limited budgets.

\textbf{Theorem 2.} \ \emph{Under the same condition in Theorem 1, by setting $\rho = \frac{\alpha_pT_n}{\alpha_nT_p}$, the proposed OA3 within budgets satisfies for any $\mu \in \mathbb{R}^d$:}
\begin{align}
   \mathbb{E}\Big{[}sum\Big{]}  \geq   1  -  \frac{\alpha_n \rho_{max}}{T_n \rho_{min}} \bigg{[}\sum_{t=1}^{T_B}\ell_t(\mu) \small{+}\frac{1}{\eta} D(\mu) \emph{Tr}(\Sigma_{T_{B}+1}^{-1})\bigg{]}.\nonumber
\end{align}

\textbf{Remark. } \emph{By setting $\alpha_p = \alpha_n = 0.5$, we can easily obtain the bound of the balanced accuracy.}

Note that $\alpha_n$ cannot be set to zero, because $\rho=\frac{\alpha_pT_n}{\alpha_nT_p}$. One restriction is that we could not acquire $\frac{T_n}{T_p}$ in advance in real-world tasks. To overcome this limitation, we can choose $cost$ metric as an alternative, where $\rho=\frac{c_p}{c_n}$. \revise{In this sense}, engineers need not worry $\frac{T_n}{T_p}$ any more. Next, we bound the cumulative $cost$ performance under limited budgets.

\textbf{Theorem 3.} \ \emph{Under the same condition in Theorem 1, by setting $\rho\small{=}\frac{c_p}{c_n}$, the proposed OA3 within budgets satisfies for any $\mu \in \mathbb{R}^d$:}
\begin{align}
   \mathbb{E}\Big{[}cost\Big{]} \leq  \frac{c_n \rho_{max}}{\rho_{min}} \bigg{[}\sum_{t=1}^{T_B}\ell_t(\mu) \small{+}\frac{1}{\eta} D(\mu) \emph{Tr}(\Sigma_{T_{B}+1}^{-1})\bigg{]}. \nonumber
\end{align}
Note that $c_n$ cannot be set to zero, since $\rho\small{=}\frac{c_p}{c_n}$.

By now, we have analyzed OA3 algorithm within budgets. Next, we analyze OA3 for the case \textbf{over budgets}.

\textbf{Theorem 4.} \ \emph{Let $(x_1,y_1),...,(x_T,y_T)$ be a sample stream, where $x_t \in \mathbb{R}^{d}$ and $y_t \in \{-1,+1\}$. Let $T_B$ be the round that uses up the budgets, \ie,  $B_{T_{B}+1}\small{=}B$. For any $\mu \small{\in} \mathbb{R}^d$, the expected mistakes of OA3 over budgets is bounded by:}
\begin{align}
   \mathbb{E}\bigg{[}\sum_{T_{B}+1}^{T} M_t\bigg{]} \leq \sum_{T_{B}+1}^{T}\bigg{[}\frac{\ell_t(\mu)}{\rho_{min}} + y_tx_t^{\top}\mu_{T_{B}+1}\bigg{]}, \nonumber
\end{align}
where $\mu_{T_{B}+1}$ is the predictive vector of model, trained by all the previous queried samples.

Now, we bound the weighted $sum$ and misclassification $cost$ after running out of budgets.

\textbf{Theorem 5.} \ \emph{Under the same condition in Theorem 4, by setting $\rho = \frac{\alpha_pT_n}{\alpha_nT_p}$, the sum performance of OA3 over budgets satisfies for any $\mu \in \mathbb{R}^d$:}
\begin{align}
   \mathbb{E}\Big{[} sum\Big{]}\geq 1  - \frac{\alpha_n \rho_{max}}{T_n}\sum_{T_{B}+1}^{T}\bigg{[}\frac{\ell_t(\mu)}{\rho_{min}} + y_tx_t^{\top}\mu_{T_{B}+1}\bigg{]}. \nonumber
\end{align}

\textbf{Theorem 6.} \ \emph{Under the same condition in Theorem 4, by setting $\rho\small{=}\frac{c_p}{c_n}$, the misclassification cost of OA3 over budgets satisfies for any $\mu \in \mathbb{R}^d$:}
\begin{align}
   \mathbb{E}\Big{[}cost\Big{]} \leq c_n \rho_{max}\sum_{T_{B}+1}^{T}\bigg{[}\frac{\ell_t(\mu)}{\rho_{min}} + y_tx_t^{\top}\mu_{T_{B}+1}\bigg{]}. \nonumber
\end{align}

\section{Enhanced Algorithm With Sketching}

As mentioned above, OA3 \yifan{requires $\mathcal{O}(Td^2)$ time complexity.  The diagonal version accelerates the time complexity to $\mathcal{O}(Td)$}. However, it cannot enjoy the correlation information between different dimensions of samples. When instances have low \emph{effective rank}, the regret bound of OA3$_{diag}$ may be much worse than its full-matrix version, \revise{because it lacks} enough dependence on data dimensionality\cite{Krummenacher2016Scalable}.
Unfortunately, many real-world high-dimensional datasets have such low-rank settings with abundant correlations \blue{among} features. For these datasets, it is more appropriate to consider the complete feature correlations \shuai{(\ie, adopting its full-matrix version)} \revise{and also the efficiency issue.}

\revise{To better balance performance and efficiency}, we propose two efficient variants of our OA3 algorithm, which use \yifan{the} sketch method to approximate the \shuai{inverse of the covariance matrix.}  \yifan{Specifically, we first propose Sketched Online Adaptive Asymmetric Active (SOA3) Learning algorithm in Subsection 4.1 and then present its sparse version (SSAO3) in Subsection 4.2.}

\subsection{Sketched Algorithm}

\yifan{By exploiting the Oja's sketch method, we propose a SAO3 algorithm \cite{luo2016efficient,Oja1982Simplified} to accelerate our algorithm when facing high-dimensional datasets.}

\shuai{The Oja's sketch~\cite{oja1985stochastic} is a method to compute the dominant eigenvalues and corresponding eigenvectors of a $n\small{\times} n$ matrix $A$. In Oja's method, matrix $A$ has the following property: $A$ itself is unknown but there is an available sequence $A_k, k\small{=}1,2,...$ with $E\{A_k\}\small{=}A$ for all $k$. In our proposed OA3 algorithm, the computation of inverse covariance matrix $\Sigma^{\small{-}1}$ in Eq. \ref{eq:inverse_sigma} can be transformed into this formula. Thus, the Oja's sketch method can be introduced to our OA3 algorithm.}

\yifan{In SOA3, we exploit Oja's sketch to search $m$ carefully selective directions and use them to  approximate our second-order \shuai{inverse covariance matrix}. \revise{Here}, $m$ is a small constant and called as the sketch size.} According to Eqs.~(\ref{eq:sigma_t+1}-\ref{eq:mu_t+1}), we know the updating rule of the model weight $\mu$:
\begin{align}
  \mu_{t+1}= \mu_{t}-\eta\Sigma_{t+1}g_{t}, \nonumber
\end{align}
and the incremental formula of the covariance matrix:
\begin{align}
  \Sigma_{t+1}^{-1}= \Sigma_{t}^{-1}+ \frac{x_t x_t^{\top}}{\gamma},\nonumber
\end{align}
which can be expressed in another way:
\begin{align}\label{eq:sigma_t+1_inv}
  \Sigma_{t+1}^{-1}= I_d + \sum_{i=1}^{t}\frac{x_i x_i^{\top}}{\gamma},
\end{align}
where $d$ is the dimensionality of instance.
Let $X\small{\in}\mathbb{R}^{t\times d}$ be a matrix, whose $t$-th row is $\hat{x}_t^{\top}$, where we define $\hat{x}_t\small{=}\frac{x_t}{\sqrt{\gamma}}$ as the \emph{to-sketch vector}. Hence,  Eq.~(\ref{eq:sigma_t+1_inv}) can be written as:
\begin{align}
  \Sigma_{t+1}^{-1} =I_d + X_t^{\top}X_t.\nonumber
\end{align}

The idea of sketching is to maintain a sketch matrix $S_t\in\mathbb{R}^{m\times d}$, where $m\ll d$. When $m$ is chosen so that $S_t^{\top}S_t$   can  approximate $X_t^{\top}X_t$ well, the Eq.~(\ref{eq:sigma_t+1_inv}) can be redefined as:
\begin{align}
  \Sigma_{t+1}^{-1} =I_d + S_t^{\top}S_t.\nonumber
\end{align}
By Woodbury identity \cite{Horn1985matrix}, we have:
\begin{align}\label{eq:sigma_t+1_s}
  \Sigma_{t+1} =I_d - S_t^{\top}H_t S_t,
\end{align}
where $H_t=(I_m+S_tS_t^{\top})^{-1}\in\mathbb{R}^{m\times m}$. We rewrite the updating rule of $\mu$:
\begin{align}\label{eq:mu_t+1_s}
    \mu_{t+1} = \mu_t-\eta(g_t-S_t^{\top}H_tS_tg_t).
\end{align}

We summarize SOA3 in Algo.~\ref{al:soa3_scheme}.
\begin{algorithm}
    \caption{Sketched Online Adaptive Asymmetric Active (SOA3) Learning algorithm.} \label{al:soa3_scheme}
    \begin{algorithmic}[1]
        \Require budget $B$; learning rate $\eta$; regularized parameter $\gamma$; sketch size $m$; \yifan{bias $\rho\small{=}\frac{\alpha_p*T_n}{\alpha_n*T_p}$ for $sum$ and $\rho\small{=}\frac{c_p}{c_n}$ for $cost$}.
        \Ensure \yifan{$\mu_1\small{=}0$, $B_1\small{=}0$}.
        \Ensure   $(S_0,H_0)\leftarrow$ \textbf{SketchInit}$(m);$ (See Algo.~\ref{al:soa3_oja})
        \For{$t = 1 \to T$}
            \State Receive sample $x_t$;
            \State Compute $p_t = \mu_t^{\top}x_t$;
            \State Make the prediction $\hat{y}_t = sign(p_t)$;
            \State Draw $Z_t \small{=} \textbf{SketchQuery}(p_t)\small{\in} \{0,1\};$ (See Algo.~\ref{al:soa3_query})
            \If {$Z_t = 1$ and $B_t < B$}
                \State Query the true label $y_t\in\{-1,+1\}$, $B_{t+1}\small{=}B_t\small{+}1$;
                \State Compute the loss $\ell_t(\mu_t)$, based on Equation (4)$;$
                \State Compute the $t$-$sketch$ $vector$ $\hat{x}_t= \frac{x_t}{\sqrt{\gamma}};$
                \State $(S_t,H_t) \leftarrow$ \textbf{SketchUpdate}$(\hat{x});$  (See Algo.~\ref{al:soa3_oja})
                \If {$\ell_t(\mu_t)>0$}
                    \State $\mu_{t\small{+}1}\small{=} \mu_{t}\small{-}\eta\small{(}g_{t}\small{-}S_t^{\top}H_t S_t g_t\small{)}, \text{where} \ g_t \small{=}\partial_{\mu}\ell_t\small{(}\mu_t\small{)};$
                \Else
                     \State $\mu_{t+1} \small{=} \mu_{t};$
                \EndIf
            \Else
                \State $\mu_{t+1} \small{=} \mu_{t}$,  $B_{t+1}\small{=}B_t$, $S_t\small{=}S_{t-1}$, $H_t\small{=}H_{t-1}.$
            \EndIf
        \EndFor
    \end{algorithmic}
\end{algorithm}

We \revise{next} discuss how to maintain matrices $S_t$ and $H_t$ efficiently via sketching technique. Specifically, with \emph{to-sketch vector} $x_t$ as input, the eigenvalues and eigenvectors of sequential data are computed by online gradient descent.

\yifan{At round $t$}, let the diagonal matrix $\Lambda_t\in\mathbb{R}^{m\times m}$ contain $m$  estimated eigenvalues and let $V_t\in\mathbb{R}^{m\times d}$ denote the corresponding \yifan{eigenvectors}. The update rules of $\Lambda_t$ and $V_t$ using Oja's algorithm are defined as:
\begin{align}
  \Lambda_{t}=&(I_m - \Gamma_{t})\Lambda_{t-1} + \Gamma_{t} diag\{V_{t-1}\hat{x}_t\}^2 , \label{eq:Lambda_t}\\
  &V_{t} \xleftarrow{orth} V_{t-1} + \Gamma_{t}V_{t-1} \hat{x}_t \hat{x}_t^{\top}, \label{eq:V_t}
\end{align}
where $\Gamma_t\in\mathbb{R}^{m\times m}$ is a diagonal matrix whose diagonal elements are learning rates. In this paper, we set $\Gamma_t = \frac{1}{t}I_m$. The "$\xleftarrow{orth}$" operator represents an orthonormalizing step\footnote{\blue{For sake of simplicity, $V_t+ \Gamma_{t+1}V_t\hat{x}_t\hat{x}_t^{\top}$ is assumed as full rank with rows all the way, so that the $\xleftarrow{orth}$ operation always keeps the same dimensionality of $V_t$.}}. Hence, the sketch matrices can be obtained by:
\begin{align}\label{13}
    &S_t = (t\Lambda)^{\frac{1}{2}}V_t, \\
    H_t = diag\{&\frac{1}{1+t\Lambda_{1,1}},...,\frac{1}{1+t\Lambda_{m,m}}\}.   \nonumber
\end{align}

The rows of $V_t$ are always orthonormal and \yifan{thus $H_t$ is an efficiently maintainable diagonal matrix}. We summarize the Oja's sketching technique in Algo.~\ref{al:soa3_oja}.
\begin{algorithm}
    \caption{Oja's Sketch for SOA3} \label{al:soa3_oja}
    \begin{algorithmic}
        \Require  $m$, $\hat{x}$ and stepsize matrix $\Gamma_t$.
        \State \hspace{-2.7ex} \textbf{Internal State} \ $t$, $\Lambda$, $V$ and $H.$
        \State \hspace{-2.2ex}\textbf{SketchInit}$(m)$
        \State \hspace{1ex}1: Set $t=0, S = 0_{m \times d},  H = I_m , \Lambda = 0_{m\times m}$
        \State \hspace{3ex} and $V$ to any $ m \times d$ matrix with orthonormal rows$;$
        \State \hspace{1ex}2: Return $(S,H).$
        \\

        \State \hspace{-2.6ex} \textbf{SketchUpdate}$(\hat{x})$
        \State \hspace{1ex}1: Update $t \leftarrow t+1;$
        \State \hspace{1ex}2: Update $\Lambda=(I_m - \Gamma_{t})\Lambda + \Gamma_{t} diag\{V\hat{x}\}^2 ;$
        \State \hspace{1ex}3: Update $V \xleftarrow{orth} V + \Gamma_{t}V \hat{x} \hat{x}^{\top};$
        \State \hspace{1ex}4: Set $S = (t\Lambda)^{\frac{1}{2}}V;$
        \State \hspace{1ex}5: Set $H = diag\{\frac{1}{1+t\Lambda_{1,1}},...,\frac{1}{1+t\Lambda_{m,m}}\};$
        \State \hspace{1ex}6: Return $(S,H).$
    \end{algorithmic}
\end{algorithm}

\yifan{\textbf{Remark.} The time complexity of this sketched updating strategy is $O(m^2d)$ per round because of the orthonormalizing operation. One can update the sketch every $m$ rounds to improve time complexity to $O(md)$\cite{Hardt2014The}.}

Last, we discuss the asymmetric query strategy in SOA3. In Section 2.4, we compute variance $v_t = x_t^{\top}\Sigma_t x_t$ to enhance the query strategy. For the same purpose, in SOA3, we compute variance based on Eq.~(\ref{eq:sigma_t+1_s}):
\begin{align}\label{eq:variance_s}
  v_t =x_t^{\top}(I_d - S_{t-1}^{\top}H_{t-1} S_{t-1})x_t.
\end{align}
\yifan{Based on Eq.~(\ref{eq:variance_s}) and Algo.~\ref{al:oa3_query}, we} summarize the sketched query strategy in Algo.~\ref{al:soa3_query}.

\begin{algorithm}
    \caption{Sketched Asymmetric Query Strategy: \textbf{\red{\textbf{SketchQuery}}$(p_t)$}.}\label{al:soa3_query}
    \begin{algorithmic}[1]
    \Require $\rho_{max}=\max\{1,\rho\}$; query bias $(\delta_{+}\small{,} \delta_{-})$ for positive and negative predictions.
        \State Compute the variance $v_t =x_t^{\top}(I_d - S_{t-1}^{\top}H_{t-1} S_{t-1})x_t;$
        \State Compute the query parameter $q_t=|p_t|- \frac{1}{2} \frac{\eta\rho_{max}}{\frac{1}{v_t}+\frac{1}{\gamma}};$
        \If {$ q_t \leq 0$}
            \State Set $q_t=0;$
        \EndIf

        \If {$ p_t \geq 0$}
            \State $p_t^{+} = \frac{\delta_{+}}{\delta_{+}+q_t};$
            \State Draw a Bernoulli variable $Z_t \small{\in} \{0,1\}$ with $p_t^{+}.$
        \Else
            \State $p_t^{-} = \frac{\delta_{-}}{\delta_{-}+q_t};$
            \State Draw a Bernoulli variable $Z_t \small{\in} \{0,1\}$ with $p_t^{-}.$
        \EndIf

        \State Return \hspace{0.5ex} $Z_t.$
    \end{algorithmic}
\end{algorithm}

\subsection{Sparse Sketched Algorithm}

In many real-world applications, \revise{data is sparse that} $||x_t||_0 \leq  s$ for all $t$, and $s\small{\ll} d$ is a small constant. For most first-order online methods,  \revise{computational complexities depend} on $s$ rather than $d$. However, SOA3 cannot enjoy this sparsity, because the sketch matrix $S_t$ will become dense quickly due to the orthonormalizing updating of $V_t$. To address this, we propose a sparse variant of SOA3 to achieve a purely sparsity-dependent time cost, named Sparse Sketched Online Adaptive Asymmetric (SSOA3) Learning.

The key of SSOA3 is to decompose the estimated eigenvectors $V_t$ and predictive vector $\mu_t$ \yifan{so that they can keep sparse. Specifically, there are two main modifications: (1) the eigenvectors $V_t$ are modified as $V_t = F_t\yifan{U}_t$, where $F_t\in \mathbb{R}^{m\times m}$ is an orthonormalizing matrix so that $F_t\yifan{U}_t$ is orthonormal, and $\yifan{U}_t\in\mathbb{R}^{m\times d}$ is a sparsely updatable direction. (2) the weight $\mu_t$ are split as $\bar{\mu}_t + \yifan{U}_{t-1}^{\top}b_t$, where $b_t\in \mathbb{R}^m$ maintains the weight on the subspace captured by $V_{t-1}$ (the same as $\yifan{U}_{t-1}$), and $\bar{\mu}_t$ captures the weight on the complementary subspace.}

Next, we describe the way to update $\bar{\mu}_t$ and $b_t$ sparsely in detail. Firstly, according to Eq.~(\ref{eq:mu_t+1_s}) and $S_t=(t\Lambda)^{\frac{1}{2}}V_t=(t\Lambda)^{\frac{1}{2}}F_t\yifan{U}_t$, we have:
\begin{footnotesize}
    \begin{align}
        \mu_{t+1} =& \mu_t - \eta (I_d -S_t^{\top}H_tS_t)g_t  \nonumber \\
        = &\bar{\mu}_t +\yifan{U}_{t-1}^{\top}b_t - \eta g_t+  \eta \yifan{U}_t^{\top} F_t^{\top}(t\Lambda H_t)F_t \yifan{U}_t g_t \nonumber \\
        = &[ \underbrace{\bar{\mu}_t - \eta g_t- (\yifan{U}_t-\yifan{U}_{t-1})^{\top}b_t}_{\bar{\mu}_{t+1}}] + \yifan{U}_t^{\top}[ \underbrace{b_t + \eta F_t^{\top}(t\Lambda H_t)F_t \yifan{U}_t g_t}_{b_{t+1}}]. \nonumber
    \end{align}
\end{footnotesize}
\yifan{Based on this equation, we define the updating rule of $\bar{\mu}_t$:
\begin{align}
    \bar{\mu}_{t+1} =   \bar{\mu}_t - \eta g_t - (\yifan{U}_t-\yifan{U}_{t-1})^{\top}b_t,  \nonumber
\end{align}
and define the updating rule of $b_t$:}
\begin{align}
    b_{t+1} = b_t + \eta F_t^{\top}(t\Lambda_tH_t)F_t\yifan{U}_tg_t. \nonumber
\end{align}
Based on the above, we summarize SSOA3 in Algo.~\ref{al:ssoa3_scheme}, where the pseudo-code of sparse Oja's algorithms is provided in Supplementary B.2 due to the page limitation.
\begin{algorithm}
    \caption{Sparse Sketched Online Adaptive Asymmetric Active (SSOA3) Learning Algorithm.}\label{al:ssoa3_scheme}
    \begin{algorithmic}[1]
        \Require budget $B$; learning rate $\eta$; regularized parameter $\gamma$; sketch size $m$; bias $\rho\small{=}\frac{\alpha_p*T_n}{\alpha_n*T_p}$ for ``sum``, $\rho\small{=}\frac{c_p}{c_n}$ for ``cost``.
        \Ensure $\bar{\mu}_1=0_{d\times 1}$, $b_1=0_{m \times 1}$, $B_1=0$;
        \Ensure $(\Lambda_0, F_0, \yifan{U}_0 ,H_0)\leftarrow$ \textbf{SparseSketchInit}$(m);$
        \For{$t = 1 \to T$}
            \State Receive sample $x_t;$
            \State Compute $p_t = \mu_t^{\top}x_t$;
            \State Make the prediction $\hat{y}_t=sign(p_t)$;
            \State Draw a variable $Z_t\small{=} \textbf{SparseSketchQuery}(p_t)\small{\in} \{0,1\}$;
            \If {$Z_t=1$ and $B_t<B$}
                \State Query the true label $y_t\small{\in}\{-1,+1\}$, $B_{t+1}\small{=}B_t\small{+}1$;
                \State Compute the loss $\ell_t(\mu_t)$, based on Equation (4);
                \State Compute the $t$-$sketch$ $vector$ $\hat{x}_t= \frac{x_t}{\sqrt{\gamma}};$
                \State \small{$(\Lambda_t, F_t, \yifan{U}_t, H_t,\delta_t)\leftarrow$ \textbf{SparseSketchUpdate}$(\hat{x});$}
                \If {$\ell_t(\mu_t) > 0$}
                    \State $\bar{\mu}_{t+1} = \bar{\mu}_t -\eta g_t -\hat{x}_t \delta_t^{\top} b_t,$ where $ g_t=\partial\ell_t(\mu_t);$
                    \State $b_{t+1} = b_t + \eta F_t^{\top}(t\Lambda_t H_t)F_t\yifan{U}_tg_t;$
                    \State $\mu_{t+1}= \bar{\mu}_{t+1} + \yifan{U}_t^{\top}b_{t+1};$
                \Else
                    \State $\mu_{t+1} = \mu_{t}$, \ $\bar{\mu}_{t+1}= \bar{\mu}_{t}$, \ $b_{t+1}= b_t;$
                \EndIf
            \Else
                \State $\mu_{t+1} = \mu_{t}$, \ $\bar{\mu}_{t+1}= \bar{\mu}_{t}$, \ $b_{t+1}= b_t$, \ $B_{t+1}=B_t;$
                \State $(\Lambda_t, F_t, \yifan{U}_t , H_t,\delta_t)$$\small{=}$$(\Lambda_{t-1}, F_{t-1}, \yifan{U}_{t-1} , H_{t-1},\delta_{t-1}).$
            \EndIf
        \EndFor
    \end{algorithmic}
\end{algorithm}

Next, we discuss how to update $\Lambda_t, \yifan{U}_t$ and $F_t$. First, we rewrite Eq.~(\ref{eq:Lambda_t}) based on $V_t = F_t\yifan{U}_t$:
\begin{align}
    \Lambda_t = (I_m-\Gamma_t) \Lambda_{t-1}+\Gamma_tdiag\{F_{t-1}\yifan{U}_{t-1}\hat{x}_t\}^2. \nonumber
\end{align}
and rewrite Eq.~(\ref{eq:V_t}) in the same way:
\begin{align}
    F_t\yifan{U}_t \xleftarrow{orth} & F_{t-1}\yifan{U}_{t-1} + \Gamma_tF_{t-1}\yifan{U}_{t-1}\hat{x}_t \hat{x}_t^{\top},  \nonumber \\
    = &F_{t-1}(\yifan{U}_{t-1} + F_{t-1}^{-1}\Gamma_tF_{t-1}\yifan{U}_{t-1}\hat{x}_t\hat{x}_t^{\top}), \nonumber
\end{align}
where $\yifan{U}_{t} = \yifan{U}_{t-1}+\delta_t\hat{x}_t^{\top}$ and $\delta_t= F_{t-1}^{-1}\Gamma_tF_{t-1}\yifan{U}_{t-1}\hat{x}_t$. Note that
$\yifan{U}_t-\yifan{U}_{t-1}$ is a sparse rank-one matrix, \yifan{which makes the update of $\bar{\mu}_t$ efficient.}

Since the update of $F_t$ is to \yifan{enforce $F_t\yifan{U}_t$ orthonormal}, \blue{we apply the Gram-Schmidt algorithm to $F_{t-1}$ in a Banach space, where the inner product is defined as $\langle a,b\rangle=a^{\top}K_tb$ and $K_t=\yifan{U}_t\yifan{U}_t^{\top}$ is the Gram matrix (see Supplementary B.2).}  \revise{Consequently}, we can update $K_t$ efficiently based on the update of $\yifan{U}_t$:
 \begin{align}
        K_t = & \yifan{U}_t\yifan{U}_t^{\top},  \nonumber \\
        = &(\yifan{U}_{t-1}+\delta_t\hat{x}_t^{\top})(\yifan{U}_{t-1}+\delta_t\hat{x}_t^{\top})^{\top},  \nonumber \\
        = &K_{t-1} +\yifan{U}_{t-1} \hat{x}_t \delta_t^{\top}+ \delta_t\hat{x}_t^{\top}\yifan{U}_{t-1}^{\top}+ \delta_t \hat{x}_t^{\top} \hat{x}_t \delta_t^{\top}. \nonumber
\end{align}

We summarize the Sparse Oja's algorithm for OA3 in Supplementary B.2.

\yifan{\textbf{Remark.} Note that both the updates of $\bar{\mu}_t$ and $b_t$ require $O(m^2+ms)$ time complexity. The most time-consuming step is the update of $F_t$,  which needs $O(m^3)$. Furthermore, the prediction $\mu_t^{\top}x_t=\bar{\mu}_t^{\top}x_t+b_t^{\top}\yifan{U}_{t-1}x_t$ can be computed in $O(ms)$ time. So, the overall time complexity of the sparse sketched update rule is $O(m^3+ms)$.}

Like SOA3, SSOA3 also computes the variance $v_t$ in an approximate way. Based on the decomposition of estimated eigenvectors $V_t = F_t \yifan{U}_t$ and Eq.~(\ref{eq:variance_s}), we have:
\yifan{\begin{align}\label{eq:variance_ss}
  v_t =x_t^{\top}(I_d - \yifan{U}_{t-1}^{\top}F_{t-1}^{\top}(t-1)\Lambda_{t-1}H_{t-1}F_{t-1}\yifan{U}_{t-1})x_t.  \nonumber
\end{align}}

Based on this equation and Algo.~\ref{al:oa3_query}, we summarize the sparse sketched asymmetric query strategy in Supplementary B.2.

\section{Experiments}
In this section, we evaluate the performance and characteristics of the proposed algorithms\footnote{The codes will be released in https://github.com/Vanint/OA3.}.

\vspace{-2ex}
\subsection{Experimental Testbed and Setup}\label{experimental_setup}
We compare \textbf{OA3} and its variants (\textbf{OA3$_{diag}$}, \textbf{SOA3}, \textbf{SSOA3}) with several state-of-the-art online active learning methods: (1) Online Passive-aggressive Active Algorithm (\textbf{PAA}) \cite{Lu2016Online}; (2) Online Asymmetric Active Algorithm (\textbf{OAAL}) \cite{Zhang2016Online}; (3) Cost-Sensitive Online Active Algorithm (\textbf{CSOAL}) \cite{Zhao2013Costd}; (4) Second-order Online Active Algorithm (\textbf{SOAL}) \cite{Hao2016second} and its cost-sensitive variant (\textbf{SOAL-CS}) \cite{Hao2016second}.

\revise{All algorithms are evaluated on four benchmark datasets.}
The statistics are summarized in Table \red{1}. Specifically, the first three datasets are obtained from LIBSVM\footnote{https://www.csie.ntu.edu.tw/~cjlin/libsvmtools/datasets/.} and the fourth dataset is obtained from KDD Cup 2008\footnote{http://www.kdd.org/kdd-cup/view/kdd-cup-2008/Data.}.

For data preprocessing, all samples are normalized by $x_t \shuai{\leftarrow} \frac{x_t}{\|x_t\|_2}$, which is extensively used in online learning, since samples are obtained sequentially. When budgets are run out, \shuai{both the query and update of the corresponding method will stop.}

For fair comparisons, all algorithms use the same experimental settings. We set $\alpha_p=\alpha_n=0.5$ for $sum$, and $c_p=0.9$ and $c_n=0.1$ for $cost$. The value of $\rho$ is set to $\frac{\alpha_p*T_n}{\alpha_n*T_p}$ for $sum$ and $\frac{c_p}{c_n}$ for $cost$. In addition, query biases $(\delta, \delta_+, \delta_-)$ and learning rates $\eta$ for all algorithms are selected from $[10^{-5}, 10^{-4}, ...,10^4, 10^5]$ using \blue{cross validations}. By default, the regularization parameter $\gamma$ is set to 1 for all second order algorithms (\ie, SOAL and OA3 based algorithms). \yifan{For our sketched algorithms, we focus on the case that the sketch size $m$ is fixed as 5, although our methods can be easily generalized by setting different sketch size like \cite{luo2016efficient}, while other implementation details are similar to \cite{luo2016efficient}.}

On each dataset, experiments are conducted over 20 random permutations of data. Results are averaged over these 20 runs and 4 metrics are employed: \emph{sensitivity}, \emph{specificity}, weighted \emph{sum} of sensitivity and specificity, and weighted \emph{cost} of misclassification. All algorithms are implemented in \blue{MATLAB on a 3.40GHz Windows machine}.

\begin{table}[h]
	\caption{\small{Datasets for Evaluation of OA3 Algorithms}}
    \vspace{-0.1in}
    \begin{center}
	\begin{tabular}{|l|c|c|c|}\hline
		Dataset &\#Examples & \#Features & \#Pos:\#Neg  \\\hline \hline
        protein & 17766 & 357 & 1:1.7 \\
        Sensorless & 58509 & 48 & 1:10 \\
        w8a & 64700 & 300 & 1:32.5 \\
        KDDCUP08 & 102294 & 117 & 1:163.19 \\\hline
	\end{tabular}
    \end{center}

\end{table}

\vspace{-5ex}
\subsection{Evaluation on Fixed Query Budgets}

\begin{figure}
    \begin{minipage}{0.485\linewidth}
      \centerline{\includegraphics[width=4.7cm]{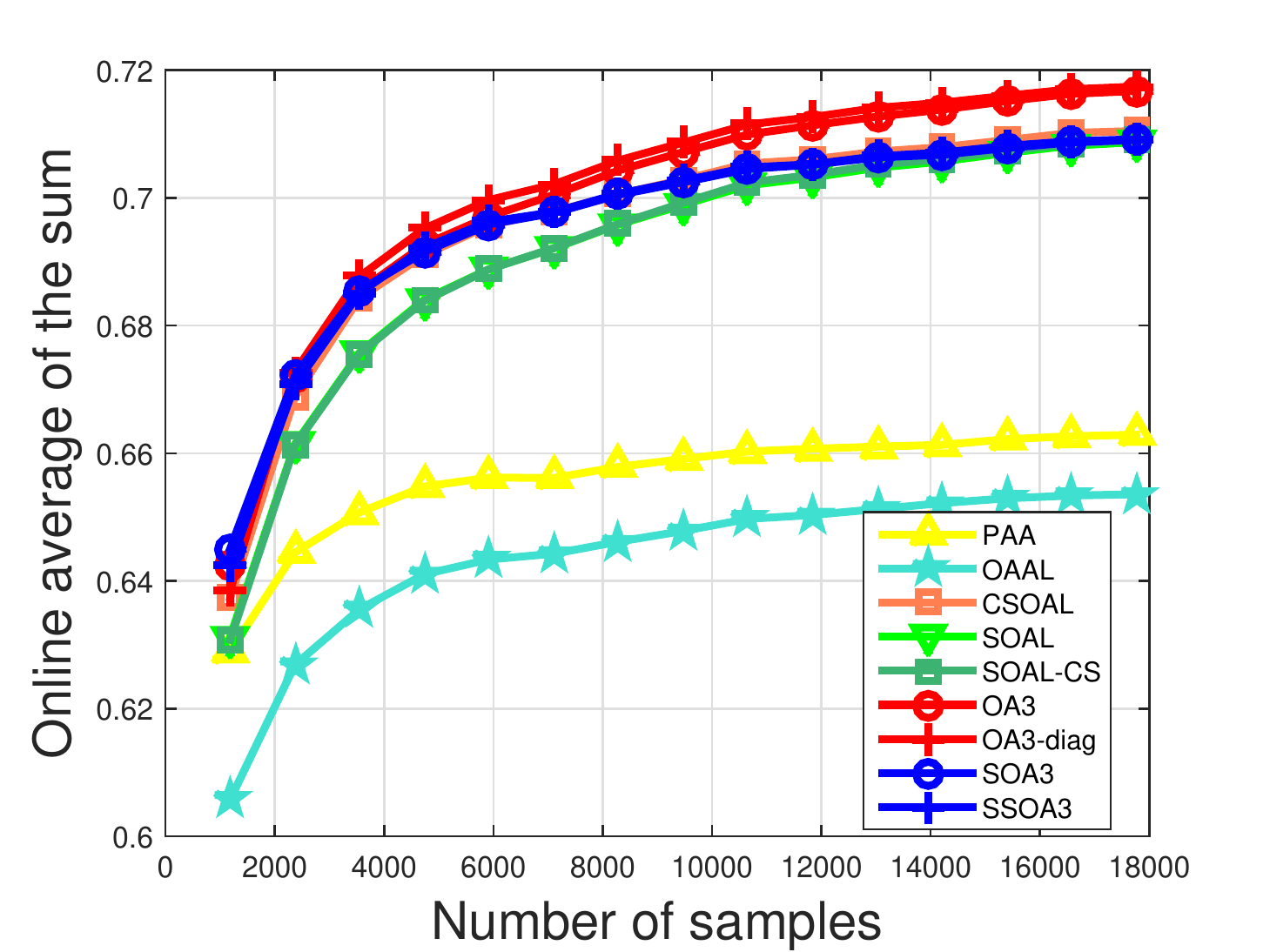}}
      \centerline{(a) protein, B=8500}
    \end{minipage}
    \hfill
    \begin{minipage}{0.485\linewidth}
      \centerline{\includegraphics[width=4.7cm]{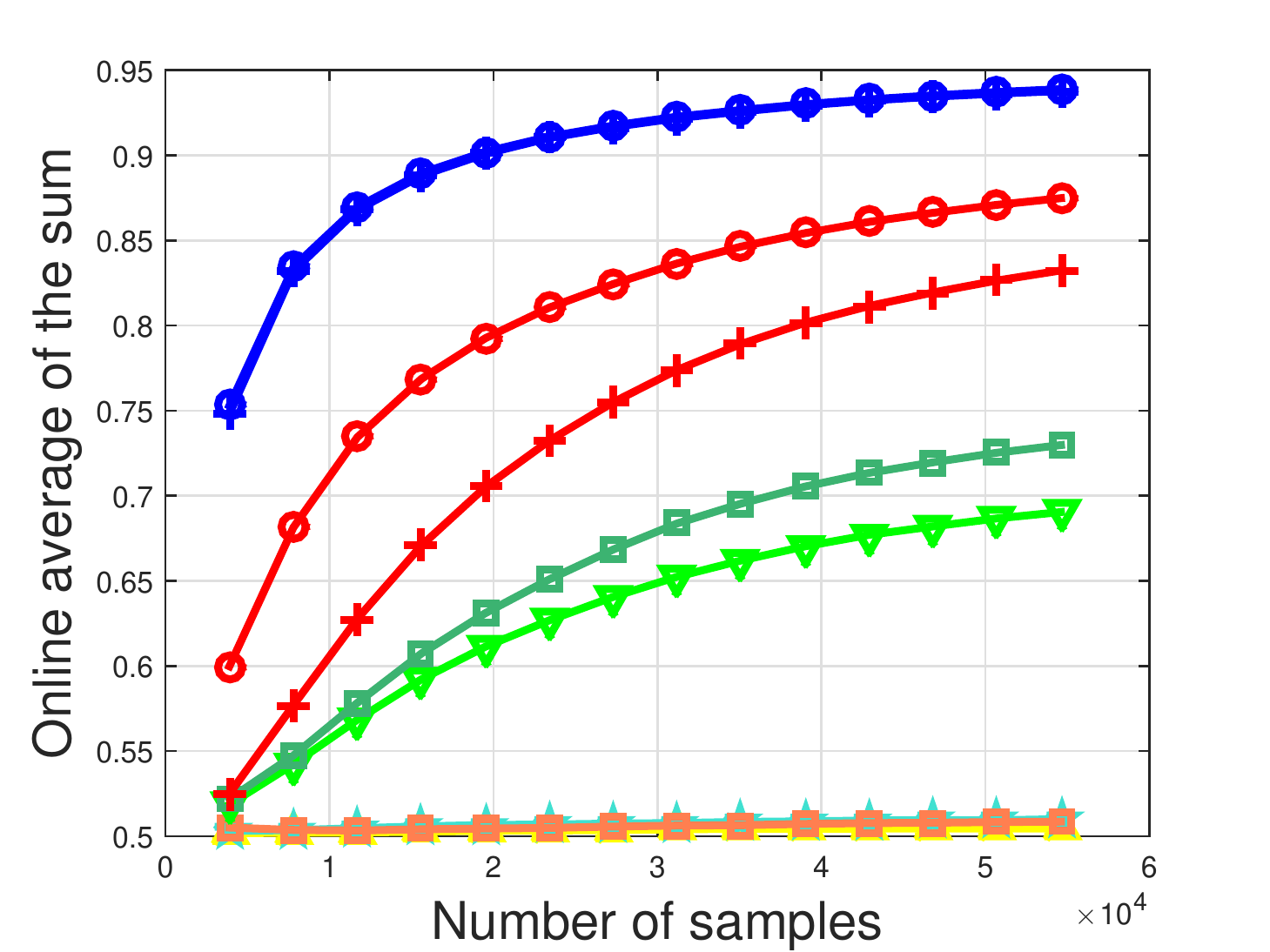}}
      \centerline{(b) Sensorless, B=29000}
    \end{minipage}
    \vfill
    \begin{minipage}{0.485\linewidth}
      \centerline{\includegraphics[width=4.7cm]{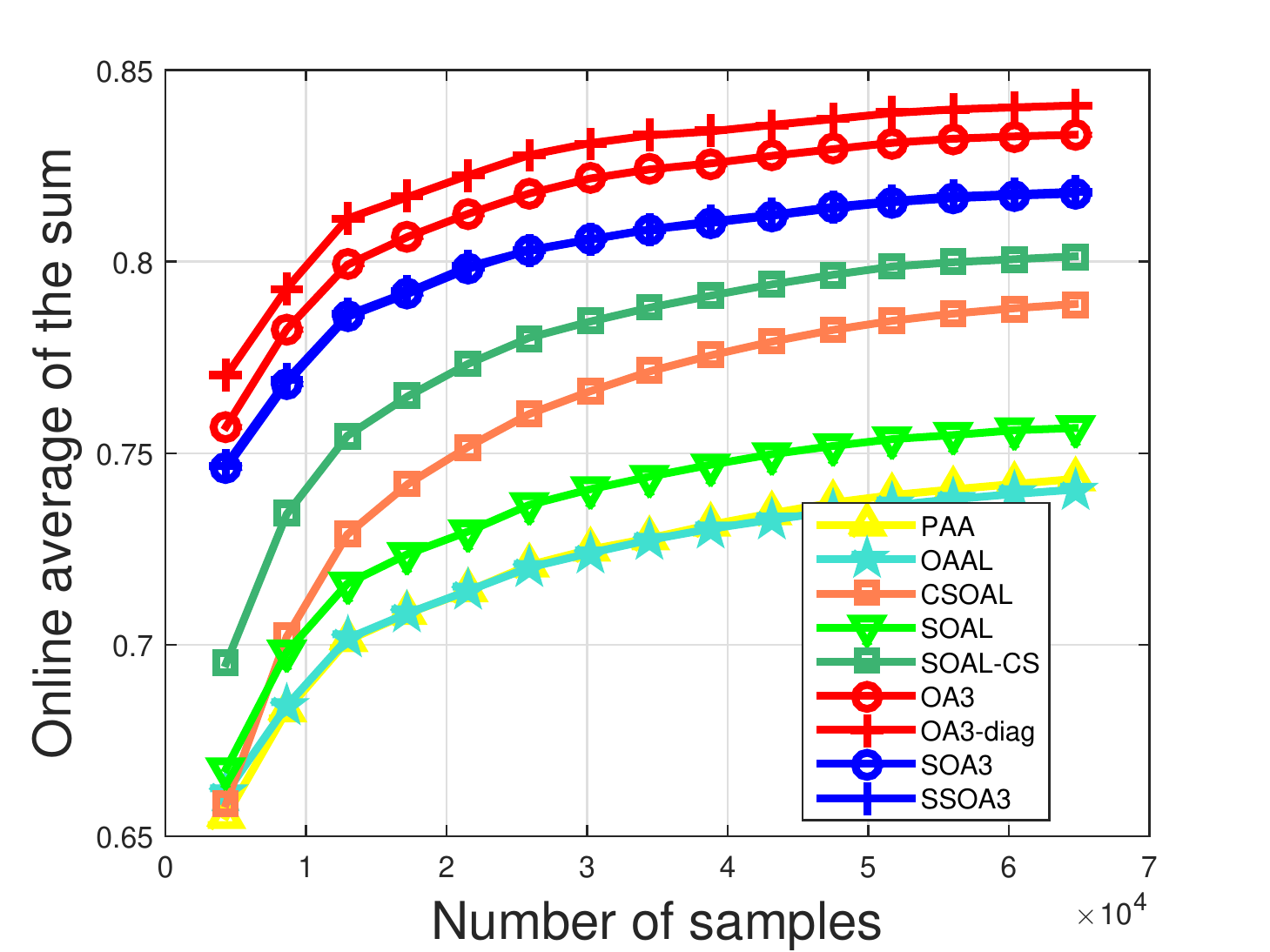}}
      \centerline{(c) w8a, B=32000}
    \end{minipage}
    \hfill
    \begin{minipage}{0.485\linewidth}
      \centerline{\includegraphics[width=4.7cm]{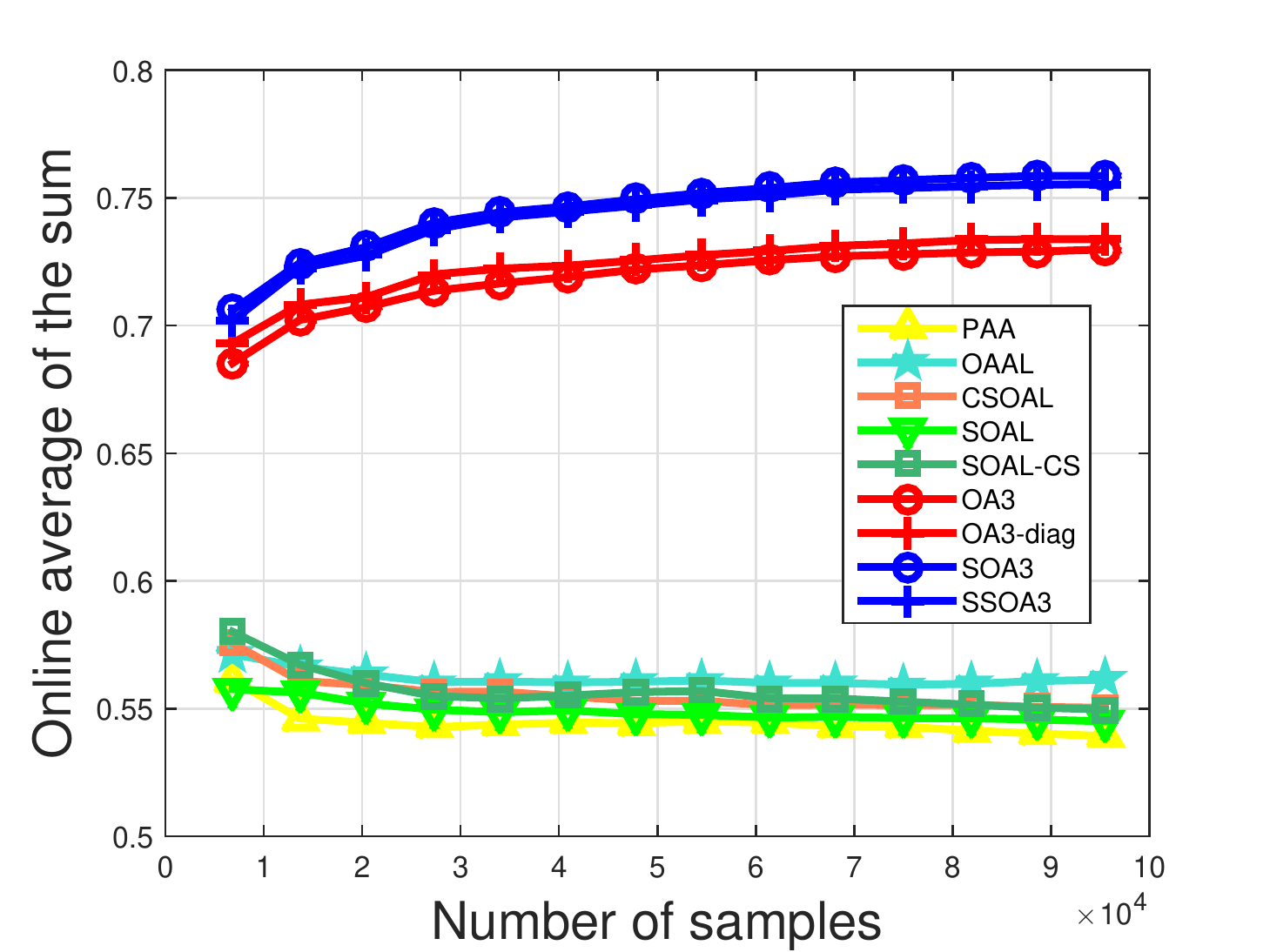}}
      \centerline{(d) KDDCUP08, B=50000}
    \end{minipage}
    \vspace{-0.1in}
    \caption{Evaluation of sum with fixed budget.}\label{sum_fix}
    \vspace{-0.15in}
\end{figure}

\begin{figure}
    \begin{minipage}{0.485\linewidth}
      \centerline{\includegraphics[width=4.7cm]{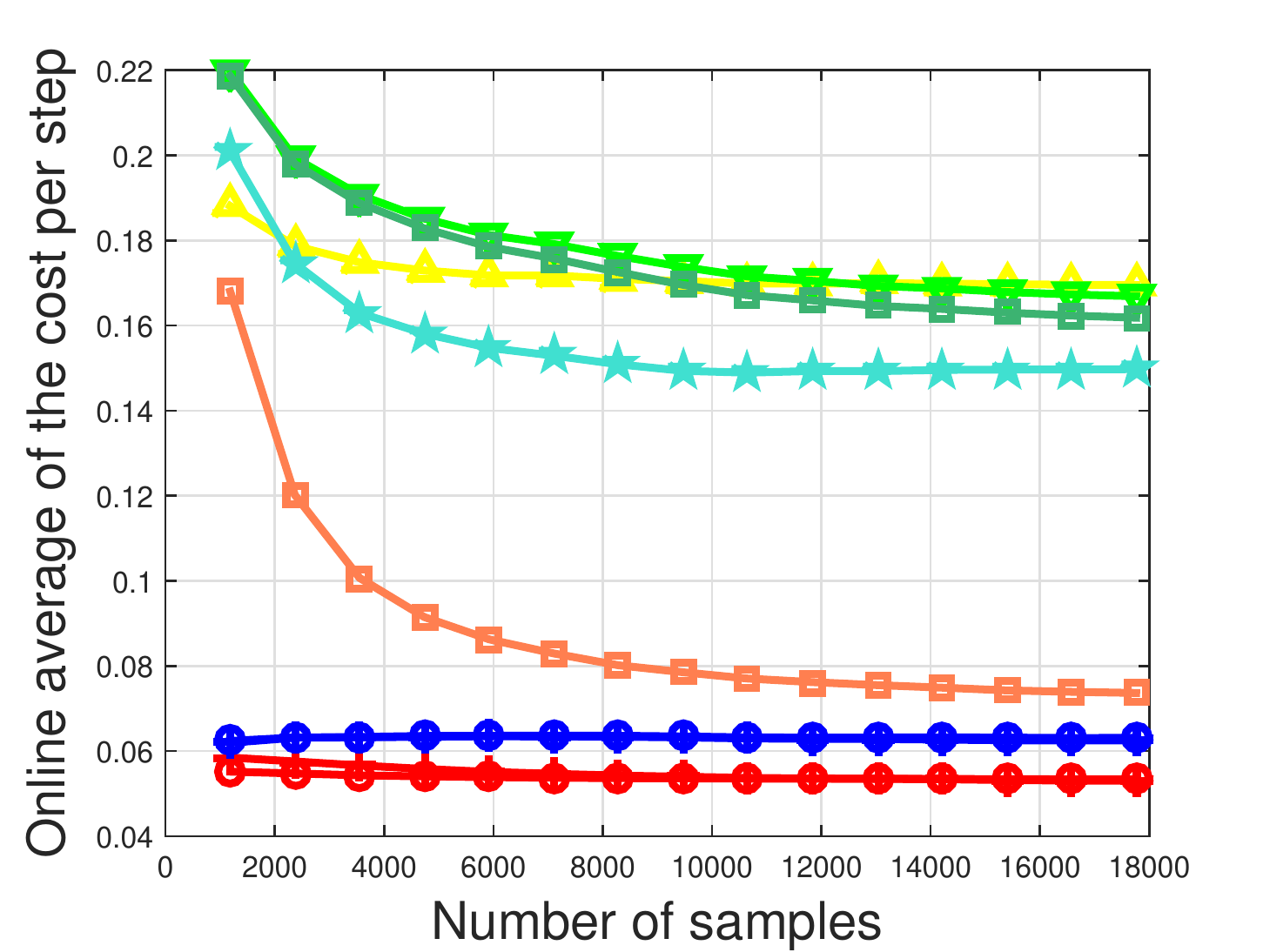}}
      \centerline{(a) protein, B=8500}
    \end{minipage}
    \hfill
    \begin{minipage}{0.485\linewidth}
      \centerline{\includegraphics[width=4.7cm]{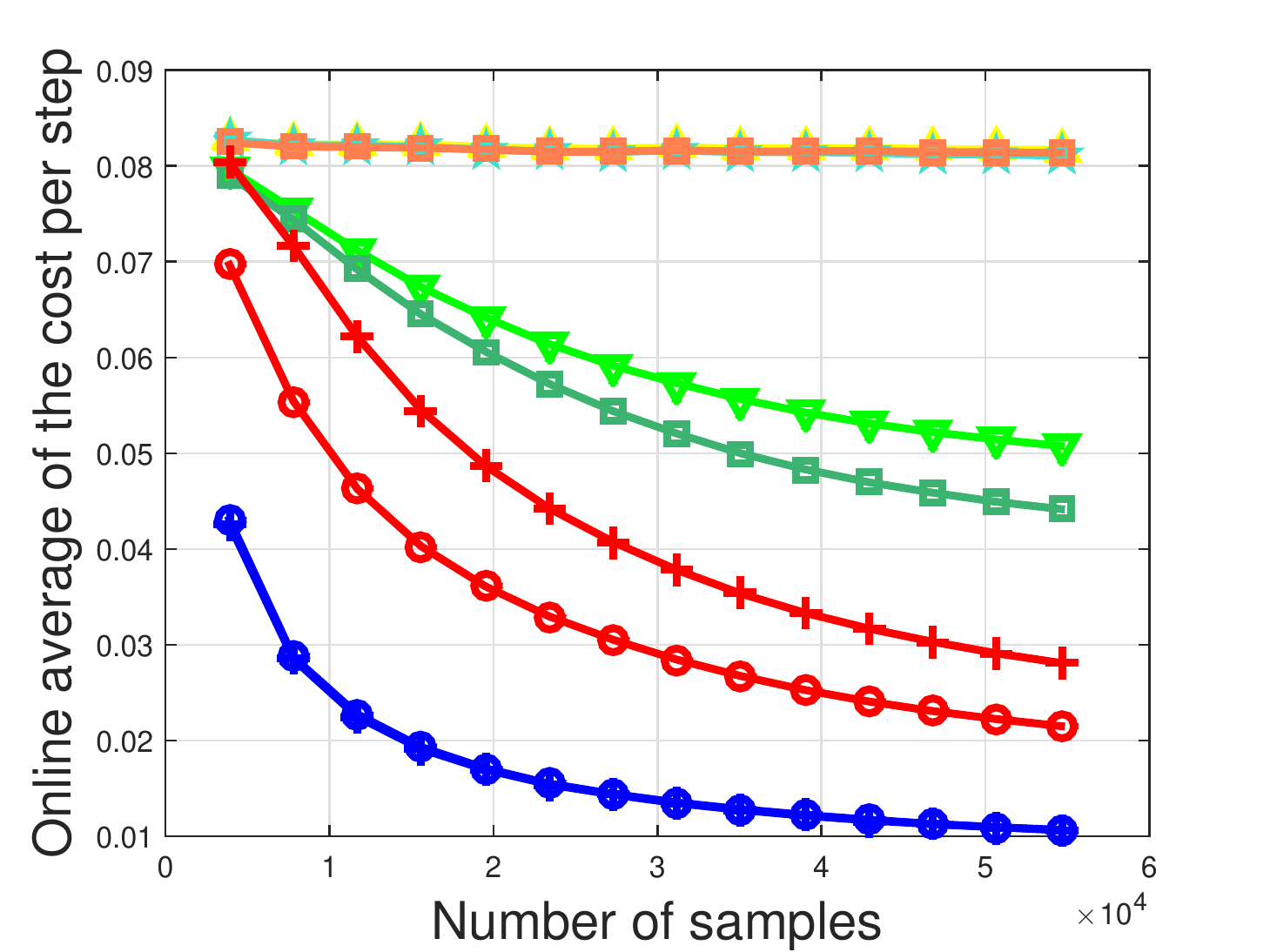}}
      \centerline{(b) Sensorless, B=29000}
    \end{minipage}
    \vfill
    \begin{minipage}{0.485\linewidth}
      \centerline{\includegraphics[width=4.7cm]{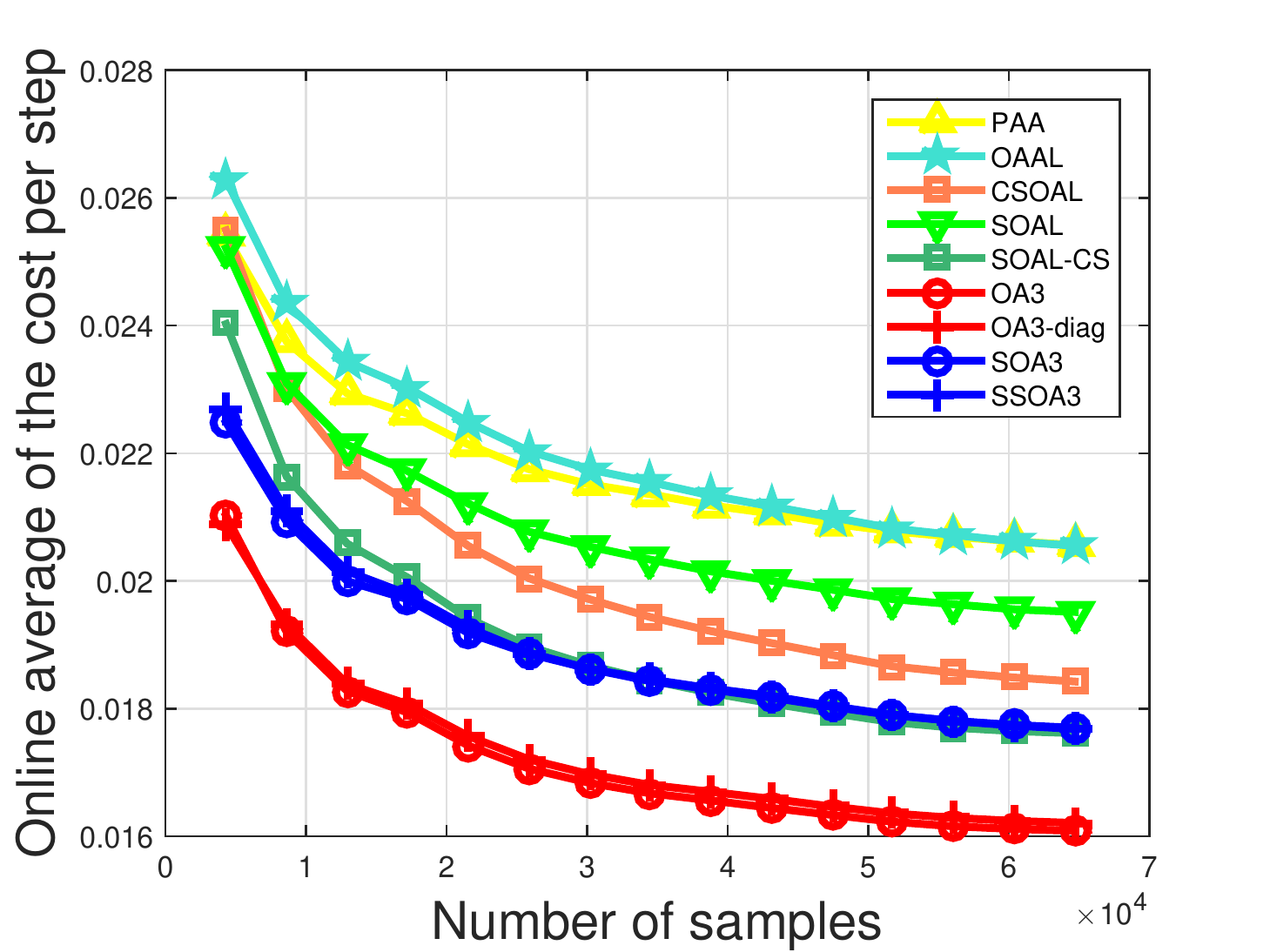}}
      \centerline{(c) w8a, B=32000}
    \end{minipage}
    \hfill
    \begin{minipage}{0.485\linewidth}
      \centerline{\includegraphics[width=4.7cm]{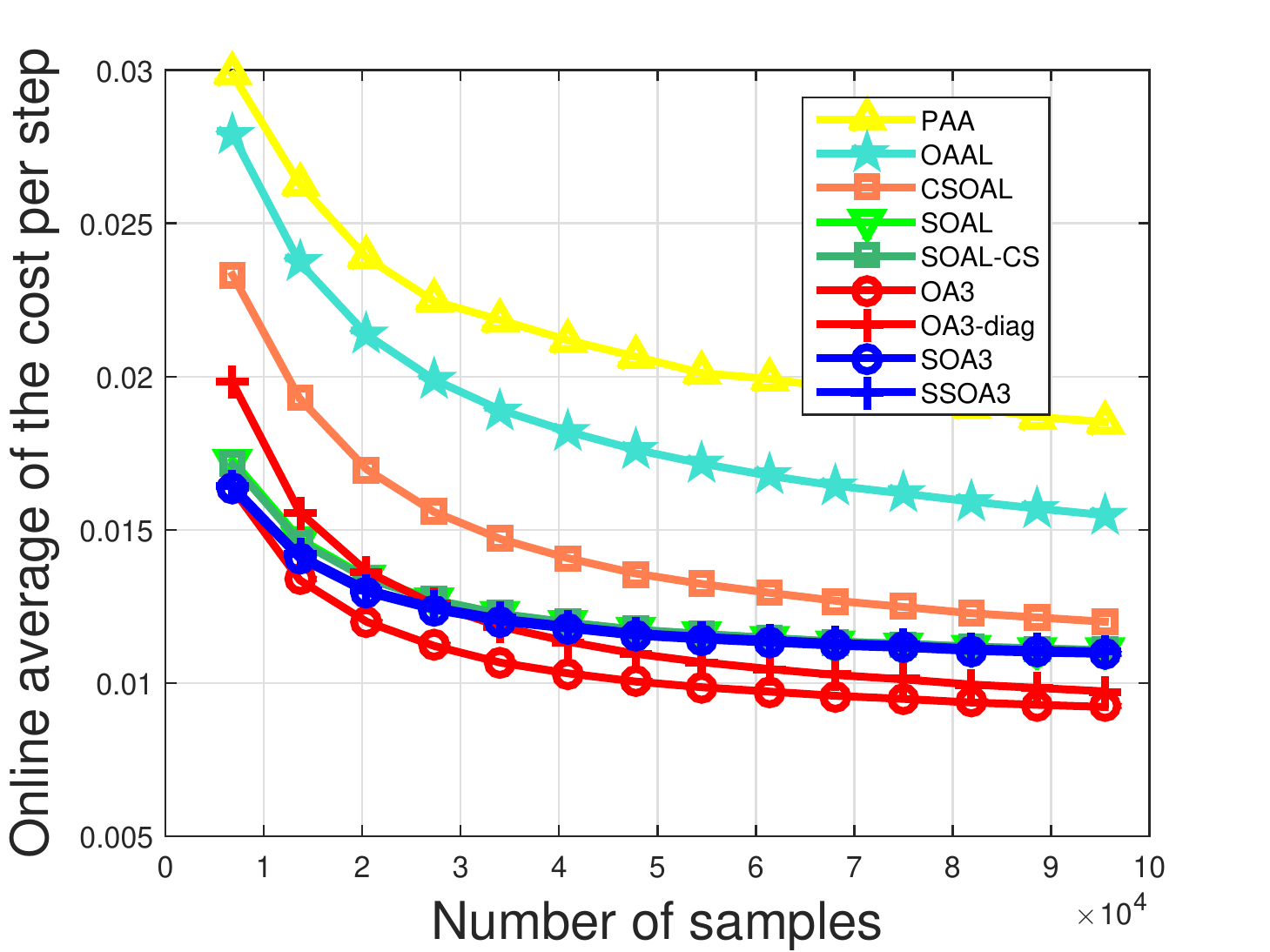}}
      \centerline{(d) KDDCUP08, B=50000}
    \end{minipage}
     \vspace{-0.1in}
    \caption{Evaluation of cost with fixed budget.}\label{cost_fix}
    \vspace{-0.15in}
\end{figure}

\begin{table*}[t]
	\caption{Evaluation of the OA3 algorithms }\label{performance}
    \vspace{-0.1in}
    \begin{center}
    \begin{scriptsize}
    \scalebox{0.96}{
    \renewcommand*\arraystretch{1.25}
	\begin{tabular}{|l|c|c|c|c|c|c|c|c|}\hline

        \multirow{2}{*}{Algorithm}&\multicolumn{4}{c|}{$``sum"$ on protein} &  \multicolumn{4}{c|}{$``cost"$ on protein}\cr\cline{2-9}
        &Sum(\%)&Sensitivity(\%)&Specificity  (\%)&Time(s)   & Cost&Sensitivity(\%)&Specificity  (\%)&Time(s)\cr
        \hline \hline
        PAA         &66.263 	$\pm$ 0.553 	& 61.455 	$\pm$ 4.455 	& 71.072 	$\pm$ 3.732 	& 0.274 &	3092.285 	$\pm$ 248.215 & 61.882 $\pm$ 3.777& 70.754 $\pm$ 3.357& 0.274 \\
        OAAL  &65.930 	$\pm$ 0.574 	& 68.953 	$\pm$ 3.442 	& 62.907 	$\pm$ 3.664 	& 0.275 	& 2657.710 	$\pm$ 210.120 	& 67.306 $\pm$ 3.142 & 74.343 $\pm$ 2.305 & 0.277 	\\
        CSOAL  &71.073 	$\pm$ 0.304 	& 69.565 	$\pm$ 3.144 	& 72.582 	$\pm$ 2.714 	& 0.258 	&1309.025 	$\pm$ 75.994 & 89.110 $\pm$ 1.285 & 47.167 $\pm$ 2.204 & 0.263 \\
        SOAL  &70.863 	$\pm$ 0.734 	& 62.576 	$\pm$ 5.437 	& \textbf{79.150 	$\pm$ 4.177} 	& 12.411 	& 2959.845 	$\pm$ 356.434 		& 62.586 $\pm$ 5.356 & \textbf{79.163 $\pm$ 4.104} & 13.095 \\
        SOAL-CS  &70.920 	$\pm$ 0.734 	& 63.115 	$\pm$ 5.490 	& 78.725 	$\pm$ 4.268 	& 12.616 &	2876.015 	$\pm$ 370.944 		& 63.853 $\pm$ 5.600& $ 78.152 \pm$ 4.478 & 13.736 \\
        OA3  &\textbf{71.673 	$\pm$ 0.348} 	& \textbf{72.608 	$\pm$ 4.700} 	& 70.738 	$\pm$ 4.838 	& 12.511  &949.855 	$\pm$ 5.615 	& \textbf{99.691 	$\pm$ 0.099} 	& 3.106 	$\pm$ 0.963 	& 8.870  \\
        OA3$_{diag}$  &71.531 	$\pm$ 0.334 	& 72.114 	$\pm$ 1.270 	& 70.949 	$\pm$ 1.393 	& 7.008  	&\textbf{938.795 	$\pm$ 9.116} 	& 99.144 	$\pm$ 0.188 	& 8.480 	$\pm$ 1.864 	& 5.318  \\
        SOA3  &71.028 	$\pm$ 0.247 	& 70.658 	$\pm$ 2.766 	& 71.398 	$\pm$ 2.485 	& 1.395  	&1117.690 	$\pm$ 38.451 	& 95.041 	$\pm$ 0.912 	& 21.421 	$\pm$ 3.461 	& 1.360  \\
        SSOA3  &70.907 	$\pm$ 0.304 	& 70.197 	$\pm$ 3.102 	& 71.616 	$\pm$ 2.967 	& 1.004 &1137.265 	$\pm$ 45.136 	& 94.474 	$\pm$ 0.971 	& 23.749 	$\pm$ 3.238 	& 0.944  \\

        \hline%
        \hline

        \multirow{2}{*}{Algorithm}&\multicolumn{4}{c|}{$``sum"$ on Sensorless} &  \multicolumn{4}{c|}{$``cost"$ on Sensorless}\cr\cline{2-9}
        &Sum(\%)&Sensitivity(\%)&Specificity  (\%)&Time(s)   & Cost&Sensitivity(\%)&Specificity  (\%)&Time(s)\cr
        \hline \hline
        PAA         &50.411 	$\pm$ 0.487 	& 8.049 	$\pm$ 9.730 	& 92.772 	$\pm$ 8.926 	& 0.426 &	 4789.080 	$\pm$ 85.260 	& 11.395 $\pm$ 14.573& $ 89.707 \pm$ 14.189 & 0.432 	\\
        OAAL  &51.661 	$\pm$ 0.377 	& 39.208 	$\pm$ 0.790 	& 64.114 	$\pm$ 0.683 	& 0.441 	& 4809.425 	$\pm$ 40.441 	& 38.309 $\pm$ 0.874 & 65.102 $\pm$ 0.578 & 0.445 	\\
        CSOAL  &50.582 	$\pm$ 0.279 	& 9.172 	$\pm$ 8.983 	& 91.993 	$\pm$ 8.670 	& 0.405 	&4774.760 	$\pm$ 30.883 	 	& 7.069 $\pm$ 2.585 & 93.870 $\pm$ 1.949 & 0.409 	 \\
        SOAL  &69.381 	$\pm$ 1.524 	& 40.079 	$\pm$ 3.141 	& \textbf{98.683 	$\pm$ 0.179} 	& 0.826 	 & 2904.685 	$\pm$ 145.436 		& 40.854 $\pm$ 3.215 & \textbf{98.621 $\pm$ 0.298} & 0.824 	\\
        SOAL-CS  &73.426 	$\pm$ 1.118 	& 49.253 	$\pm$ 2.339 	& 97.598 	$\pm$ 0.338 	& 0.874 	&  2555.670 	$\pm$ 113.284 		& 49.267 $\pm$ 2.570 & $ 97.612 \pm$ 0.362 & 0.871 	\\
        OA3   &87.944 	$\pm$ 0.516 	& 89.649 	$\pm$ 0.973 	& 86.238 	$\pm$ 1.148 	& 0.966  	&   1219.940 	$\pm$ 57.251 	& 89.113 $\pm$ 0.948& 86.863 $\pm$ 0.844 & 0.985 	\\
        OA3$_{diag}$  &86.268 	$\pm$ 0.744 	& 87.469 	$\pm$ 2.042 	& 85.067 	$\pm$ 1.374 	& 0.792 	&1364.925 	$\pm$ 63.246 	& 87.365 	$\pm$ 1.190 	& 85.710 	$\pm$ 1.682 	& 0.761  	\\
        SOA3  &\textbf{94.067 	$\pm$ 0.238} 	& 95.860 	$\pm$ 0.318 	& 92.274 	$\pm$ 0.578 	& 0.862 	&   607.540 	$\pm$ 34.277		& \textbf{95.458 $\pm$ 0.484} & 92.666 $\pm$ 0.823 & 0.863 	\\
        SSOA3  &94.011 	$\pm$ 0.349 	& \textbf{95.870 	$\pm$ 0.394} 	& 92.151 	$\pm$ 0.747 	& 0.920 	&  \textbf{605.860 	$\pm$ 29.903} 	& 95.441$\pm$ 0.472& 92.713 $\pm$ 0.707 & 0.913 	\\

        \hline
        \hline
        \multirow{2}{*}{Algorithm}&\multicolumn{4}{c|}{$``sum"$ on w8a} &  \multicolumn{4}{c|}{$``cost"$ on w8a}\cr\cline{2-9}
        &Sum(\%)&Sensitivity(\%)&Specificity  (\%)&Time(s)   & Cost&Sensitivity(\%)&Specificity  (\%)&Time(s)\cr
        \hline \hline
        PAA         &74.121 	$\pm$ 1.196 	& 57.716 	$\pm$ 2.564 	& 90.526 	$\pm$ 0.213 	& 0.990 	 &1334.555 	$\pm$ 29.329 	& 57.512 	$\pm$ 2.309 	& 90.514 	$\pm$ 0.270 	& 0.982  	\\
        OAAL  &73.919 	$\pm$ 0.848 	& 57.850 	$\pm$ 1.798 	& 89.987 	$\pm$ 0.163 	& 1.022 	&1327.965 	$\pm$ 16.675 	& 56.945 	$\pm$ 1.033 	& 90.776 	$\pm$ 0.040 	& 1.034  	\\
        CSOAL  &78.929 	$\pm$ 0.586 	& 68.078 	$\pm$ 1.307 	& 89.780 	$\pm$ 0.159 	& 0.976  	&1193.555 	$\pm$ 11.979 	& 66.699 	$\pm$ 0.967 	& 90.214 	$\pm$ 0.110 	& 0.979  	 \\
        SOAL   &75.688 	$\pm$ 0.531 	& 60.476 	$\pm$ 1.071 	& \textbf{90.899 	$\pm$ 0.042} 	& 3.340 	 &1262.290 	$\pm$ 17.663 	& 60.212 	$\pm$ 1.031 	& \textbf{90.917 	$\pm$ 0.028}  	& 3.458  	\\
        SOAL-CS  &80.076 	$\pm$ 0.380 	& 69.912 	$\pm$ 0.781 	& 90.241 	$\pm$ 0.050 	& 3.763   	&1137.000 	$\pm$ 14.635 	& 69.268 	$\pm$ 0.944 	& 90.403 	$\pm$ 0.053  	& 3.752  	\\
        OA3   &83.313 	$\pm$ 0.951 	& \textbf{84.369 	$\pm$ 0.565} 	& 82.257 	$\pm$ 2.365 	& 14.118  	&\textbf{1042.260 	$\pm$ 7.455} 	& \textbf{79.240 	$\pm$ 0.679} 	& 89.149 	$\pm$ 0.194 	& 9.002   	\\
        OA3$_{diag}$  &\textbf{84.129 	$\pm$ 0.260} 	& 83.101 	$\pm$ 0.435 	& 85.157 	$\pm$ 0.742 	& 6.691  	&1048.075 	$\pm$ 7.826 	& 79.020 	$\pm$ 0.477 	& 89.117 	$\pm$ 0.126 	& 5.398   	\\
        SOA3  &81.759 	$\pm$ 0.455 	& 80.864 	$\pm$ 0.760 	& 82.655 	$\pm$ 1.303 	& 4.556  	&1150.620 	$\pm$ 10.609 	& 76.234 	$\pm$ 0.974 	& 88.256 	$\pm$ 0.266 	& 4.364   	\\
        SSOA3  &81.803 	$\pm$ 0.265 	& 80.365 	$\pm$ 0.788 	& 83.241 	$\pm$ 0.816 	& 3.398  	&1149.620 	$\pm$ 8.379 	& 76.079 	$\pm$ 0.946 	& 88.315 	$\pm$ 0.273 	& 3.332  	\\

        \hline
        \hline
        \multirow{2}{*}{Algorithm}&\multicolumn{4}{c|}{$``sum"$ on KDDCUP08} &  \multicolumn{4}{c|}{$``cost"$ on KDDCUP08}\cr\cline{2-9}
        &Sum(\%)&Sensitivity(\%)&Specificity  (\%)&Time(s)   & Cost&Sensitivity(\%)&Specificity  (\%)&Time(s)\cr
        \hline \hline
        PAA         &53.433 	$\pm$ 4.439 	& 48.475 	$\pm$ 8.820 	& 58.391 	$\pm$ 2.286 	& 1.065  	 &1863.705 	$\pm$ 155.865 	& 23.900 	$\pm$ 2.803 	& 85.866 	$\pm$ 1.624 	& 1.082    	\\
        OAAL  &56.054 	$\pm$ 1.589 	& 27.006 	$\pm$ 2.978 	& 85.101 	$\pm$ 1.273 	& 1.034  	&1567.705 	$\pm$ 72.265 	& 21.701 	$\pm$ 3.897 	& 88.899 	$\pm$ 0.845 	& 1.045  	\\
        CSOAL  &55.891 	$\pm$ 3.842 	& 25.257 	$\pm$ 8.550 	& \textbf{86.526 	$\pm$ 1.625} 	& 0.938  	&1206.975 	$\pm$ 27.689 	& 13.756 	$\pm$ 1.855 	& 92.885 	$\pm$ 0.275 	& 0.974   	 \\
        SOAL  &53.768 	$\pm$ 2.346 	& 26.942 	$\pm$ 5.182 	& 80.594 	$\pm$ 1.574 	& 3.292   	 &1126.580 	$\pm$ 37.343 	& 9.222 	$\pm$ 0.525 	& 93.926 	$\pm$ 0.390 	& 5.879  	\\
        SOAL-CS  &54.775 	$\pm$ 2.515 	& 28.363 	$\pm$ 5.433 	& 81.188 	$\pm$ 1.693 	& 3.468   	&1129.770 	$\pm$ 36.692 	& 9.342 	$\pm$ 0.543 	& 93.888 	$\pm$ 0.383 	& 5.935   	\\
        OA3   &73.189 	$\pm$ 2.510 	& 90.859 	$\pm$ 2.354 	& 55.520 	$\pm$ 3.729 	& 7.066    	&\textbf{931.645 	$\pm$ 60.664} 	& 35.947 	$\pm$ 1.946 	& \textbf{94.369 	$\pm$ 0.507} 	& 4.467    	\\
        OA3$_{diag}$  &73.598 	$\pm$ 2.144 	& 87.844 	$\pm$ 1.286 	& 59.353 	$\pm$ 3.817 	& 2.699   	&977.920 	$\pm$ 13.775 	& \textbf{38.644 	$\pm$ 1.276} 	& 93.765 	$\pm$ 0.119 	& 2.708    	\\
        SOA3  &74.200 	$\pm$ 1.088 	& 88.965 	$\pm$ 1.810 	& 59.434 	$\pm$ 3.124 	& 3.779  	&1063.260 	$\pm$ 13.663 	& 25.939 	$\pm$ 3.965 	& 93.627 	$\pm$ 0.285 	& 3.928    	\\
        SSOA3  &\textbf{75.642 	$\pm$ 1.560} 	& \textbf{90.971 	$\pm$ 1.329} 	& 60.313 	$\pm$ 3.234 	& 3.376 &1056.040 	$\pm$ 16.917 	& 25.778 	$\pm$ 2.852 	& 93.706 	$\pm$ 0.296 	& 3.970   	\\
        \hline
	\end{tabular}}
    \end{scriptsize}
    \end{center}
    \vspace{-0.2in}
\end{table*}

\yifan{We first} evaluate all algorithms under fixed budgets.
Figs.~\ref{sum_fix} and \ref{cost_fix} show the development of \emph{sum} and \emph{cost} performance, respectively and Table \ref{performance} summarizes more details.%

\yifan{First, OAAL (asymmetric query) and CSOAL (asymmetric update) outperform PAA (symmetric rule) in most cases. This comparison} shows the importance of asymmetric strategies for imbalanced data in online active learning.

Second, as expected, all second order based algorithms (SOAL and OA3) outperform the first-order algorithms (PAA, OAAL and CSOAL) in most cases, which confirms the effectiveness of second-order information.

Third, the proposed OA3 algorithms outperform all baselines with smaller standard deviations, which demonstrates the effectiveness and stability of our methods.

\magenta{According to comparisons in terms of \emph{sensitivity} and \emph{specificity}, our proposed algorithms achieve the best \emph{sensitivity} on all datasets and produce fairly good \emph{specificity} on most datasets. This indicates that our algorithms pay more attention to minority samples, which are usually more important in practical tasks}.

Last, we compare the efficiency of the proposed methods. From Table \ref{performance}, OA3$_{diag}$, SOA3 and SSOA3 are much efficient than the original OA3 with quite slight performance degradation. Specifically, when datasets are quite high-dimensional, SOA3 and SSOA3 are further faster than OA3$_{diag}$. \yifan{These observations imply that SOA3 and SSOA3 are better choices for balancing performance and efficiency.}

\subsection{Evaluation on Varying Query Budgets}
\begin{figure}
\vspace{-0.1in}
    \begin{minipage}{0.485\linewidth}
      \centerline{\includegraphics[width=4.7cm]{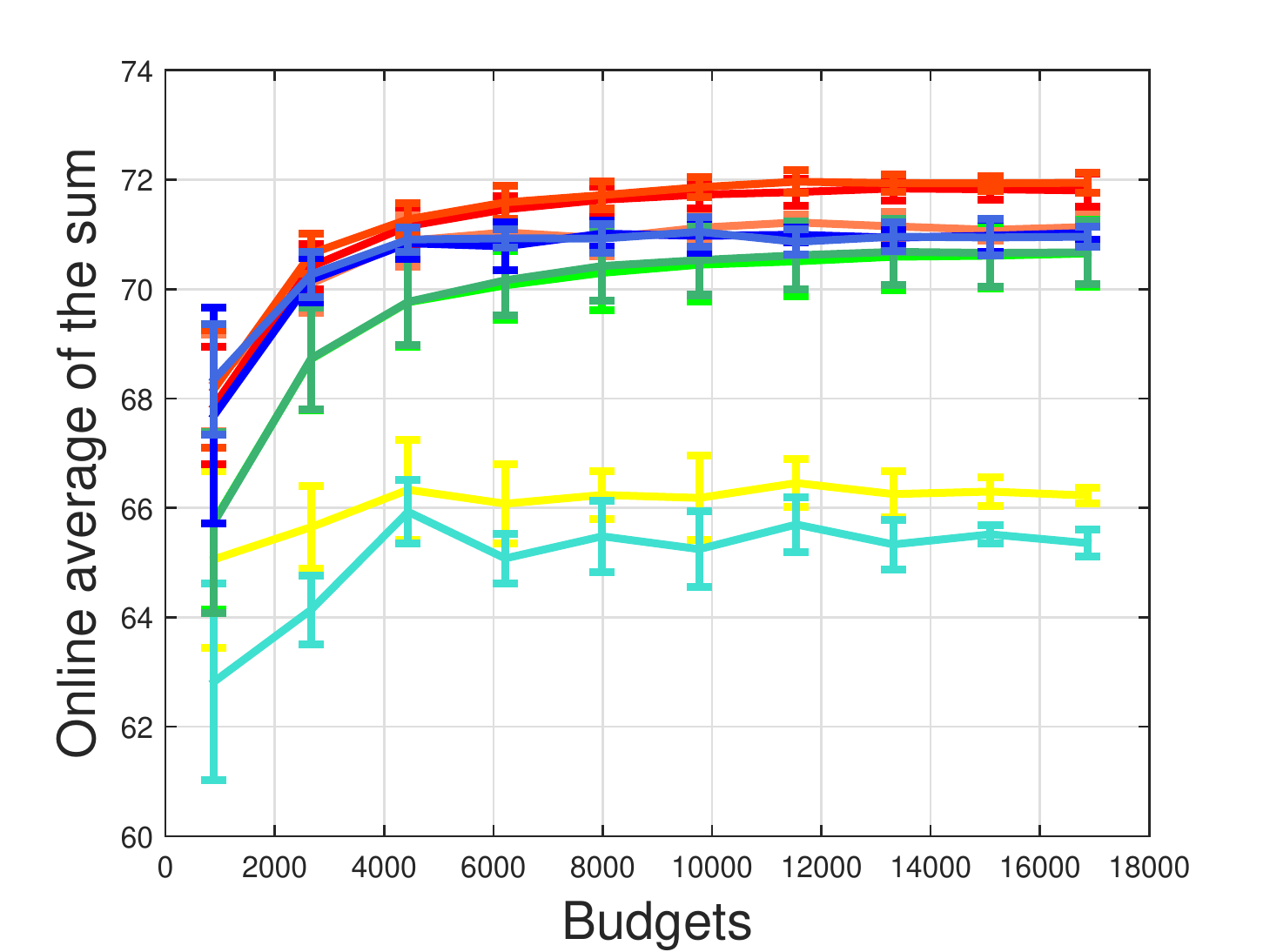}}
      \centerline{(a) protein}
    \end{minipage}
    \hfill
    \begin{minipage}{0.485\linewidth}
      \centerline{\includegraphics[width=4.7cm]{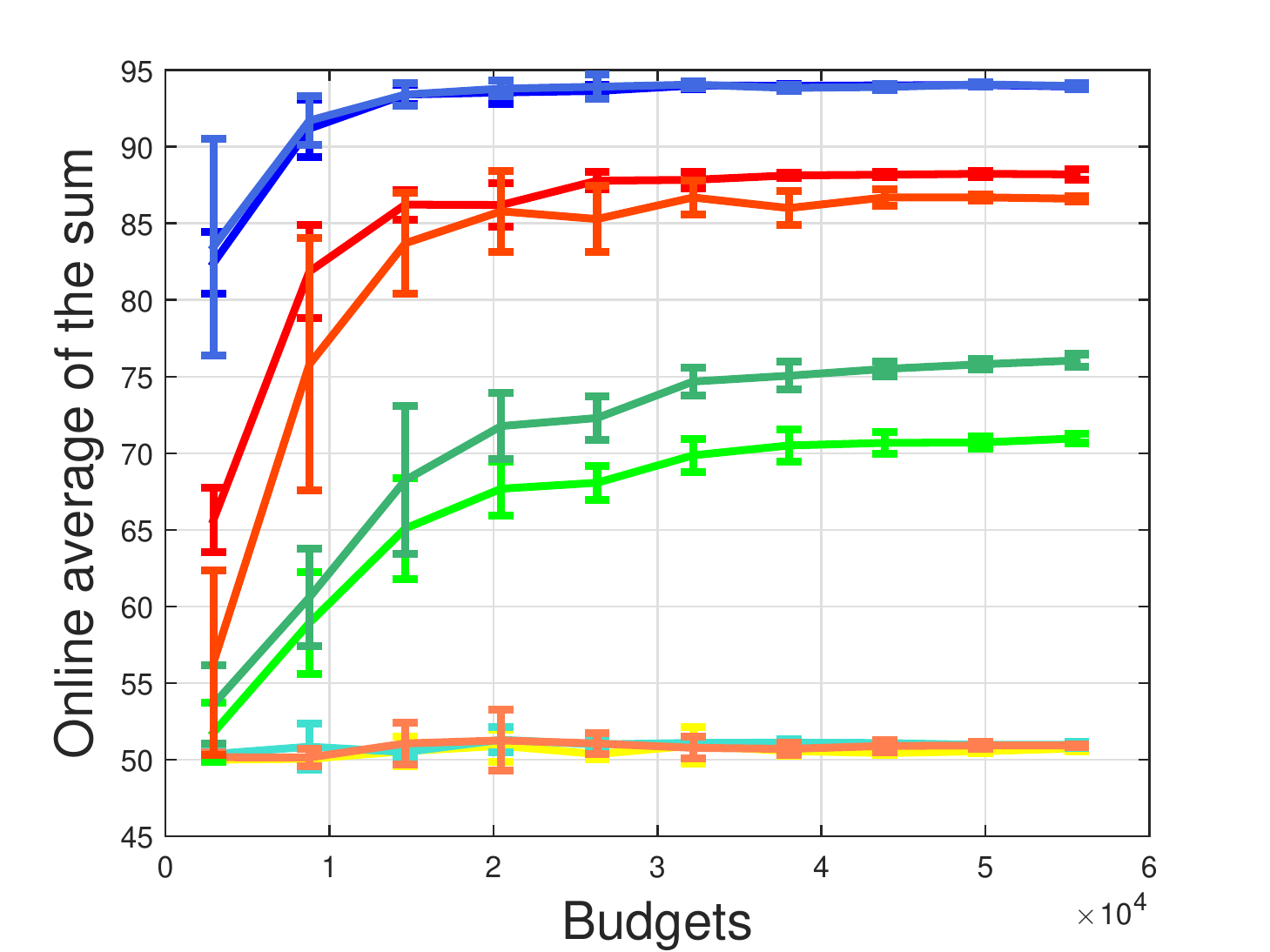}}
      \centerline{(b) Sensorless}
    \end{minipage}
    \vfill
    \begin{minipage}{0.485\linewidth}
      \centerline{\includegraphics[width=4.7cm]{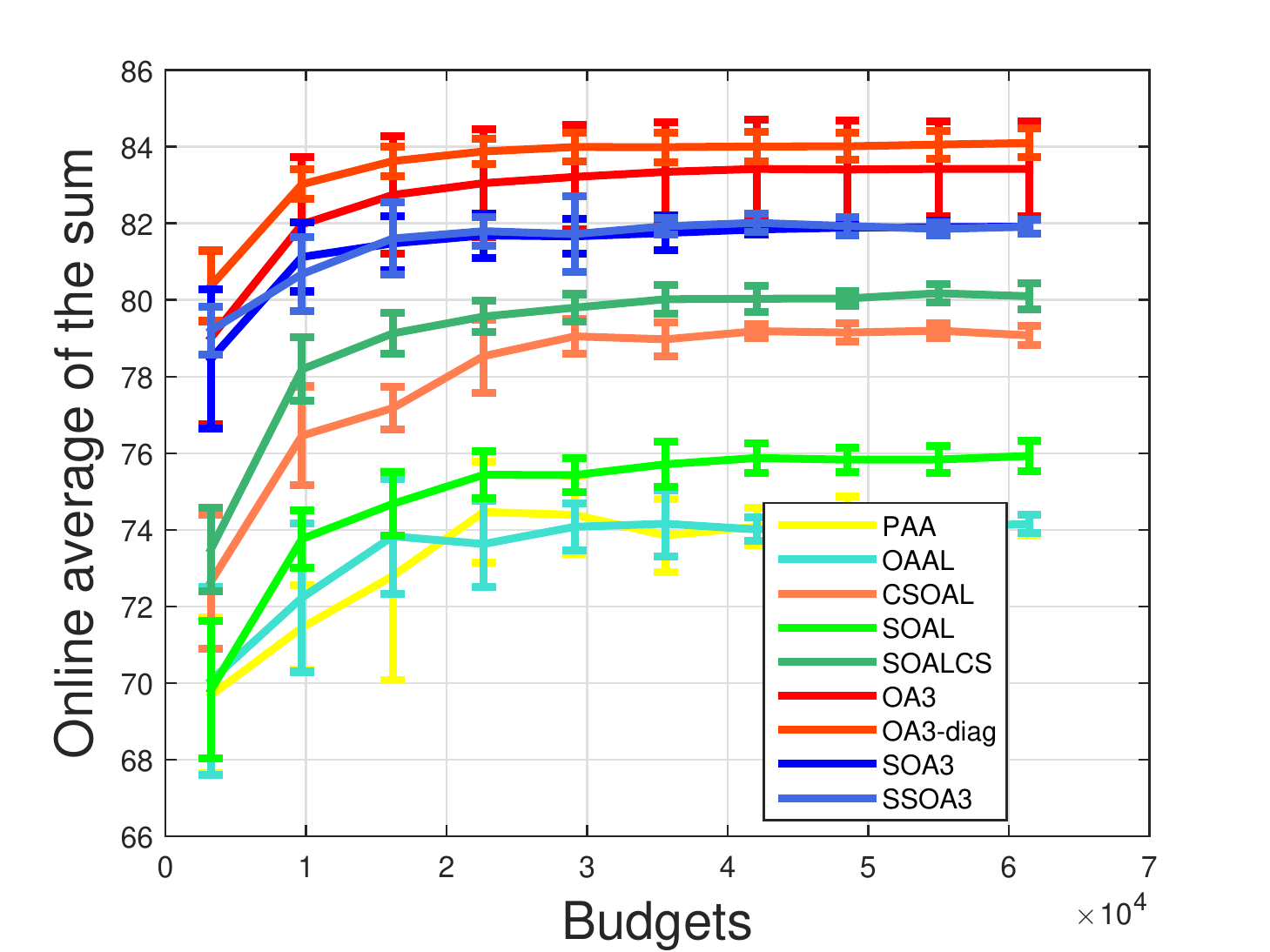}}
      \centerline{(c) w8a}
    \end{minipage}
    \hfill
    \begin{minipage}{0.485\linewidth}
      \centerline{\includegraphics[width=4.7cm]{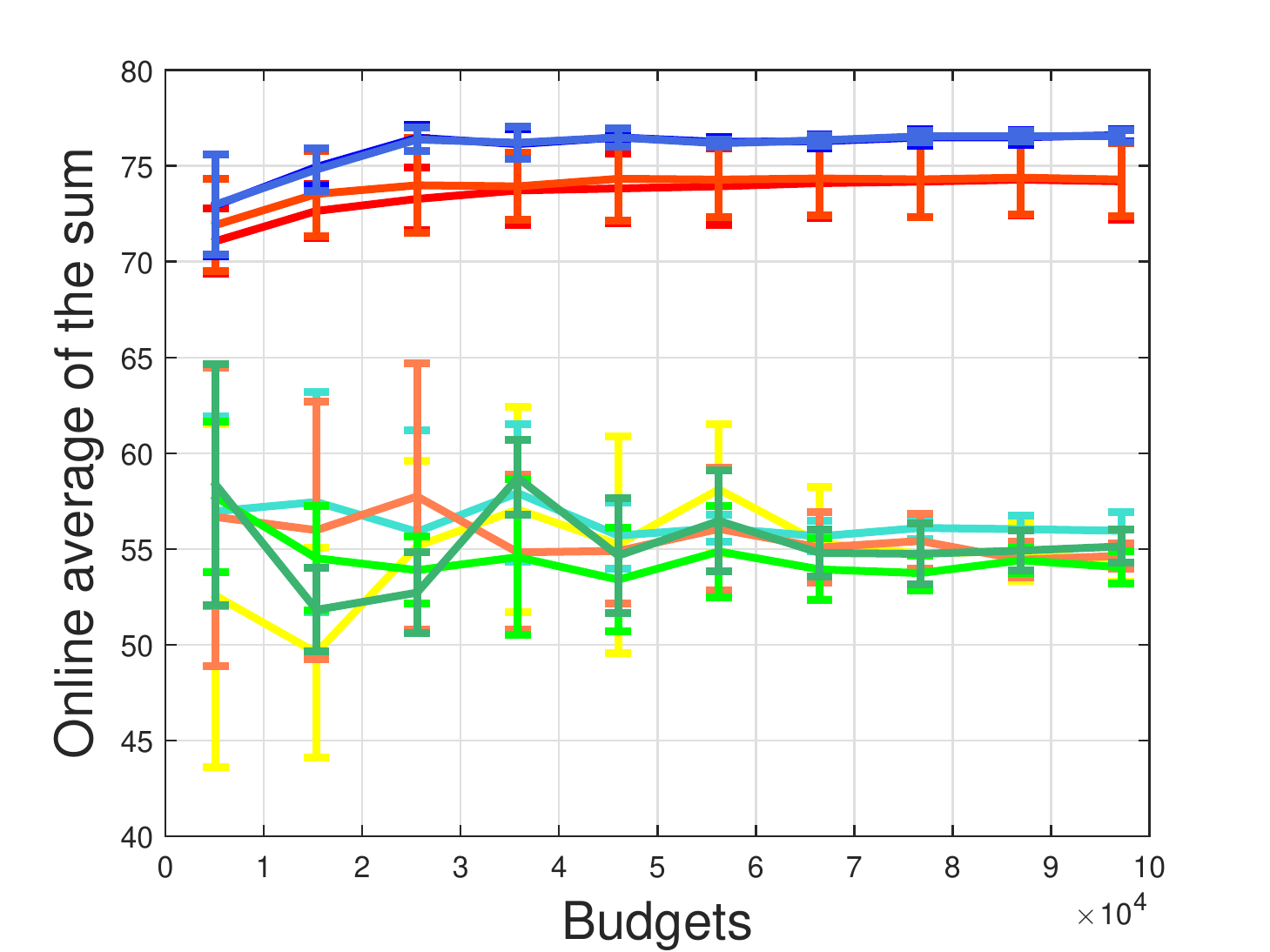}}
      \centerline{(d) KDDCUP08}
    \end{minipage}
    \vspace{-0.1in}
    \caption{Evaluation of sum with varying budgets.}\label{sum_varying}
    \vspace{-0.15in}
\end{figure}

\begin{figure}
    \begin{minipage}{0.485\linewidth}
      \centerline{\includegraphics[width=4.7cm]{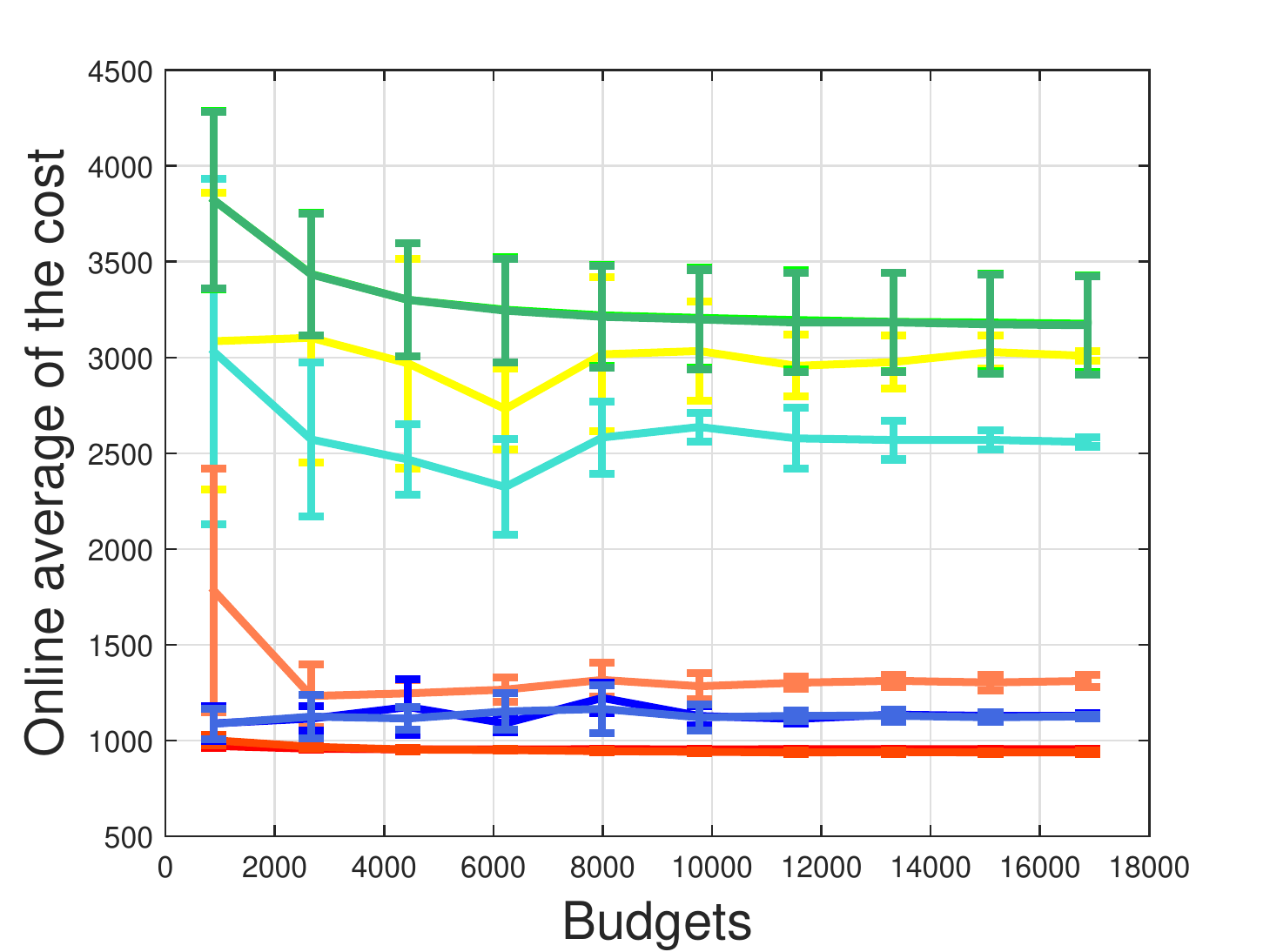}}
      \centerline{(a) protein}
    \end{minipage}
    \hfill
    \begin{minipage}{0.485\linewidth}
      \centerline{\includegraphics[width=4.7cm]{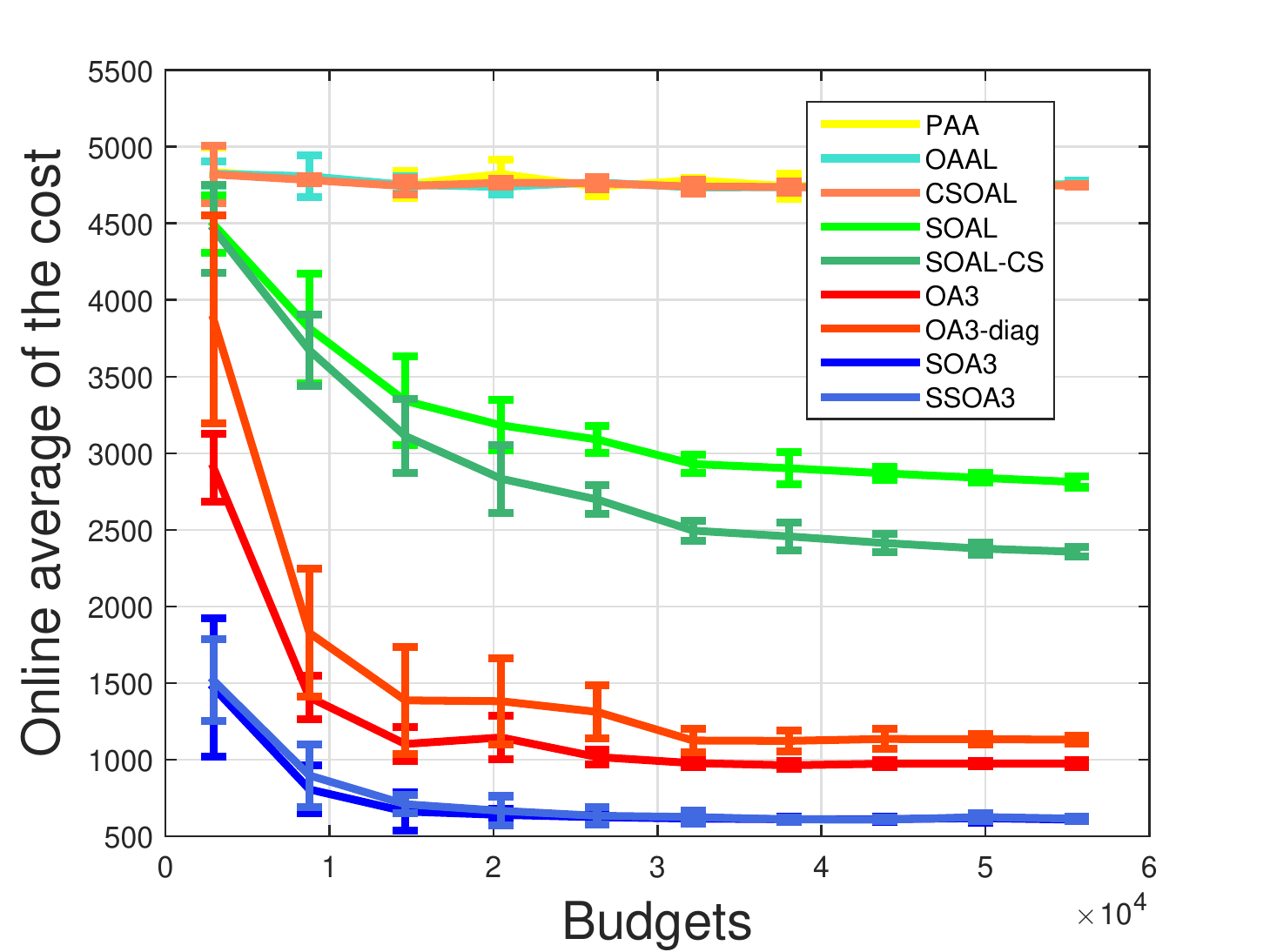}}
      \centerline{(b) Sensorless}
    \end{minipage}
    \vfill
    \begin{minipage}{0.485\linewidth}
      \centerline{\includegraphics[width=4.7cm]{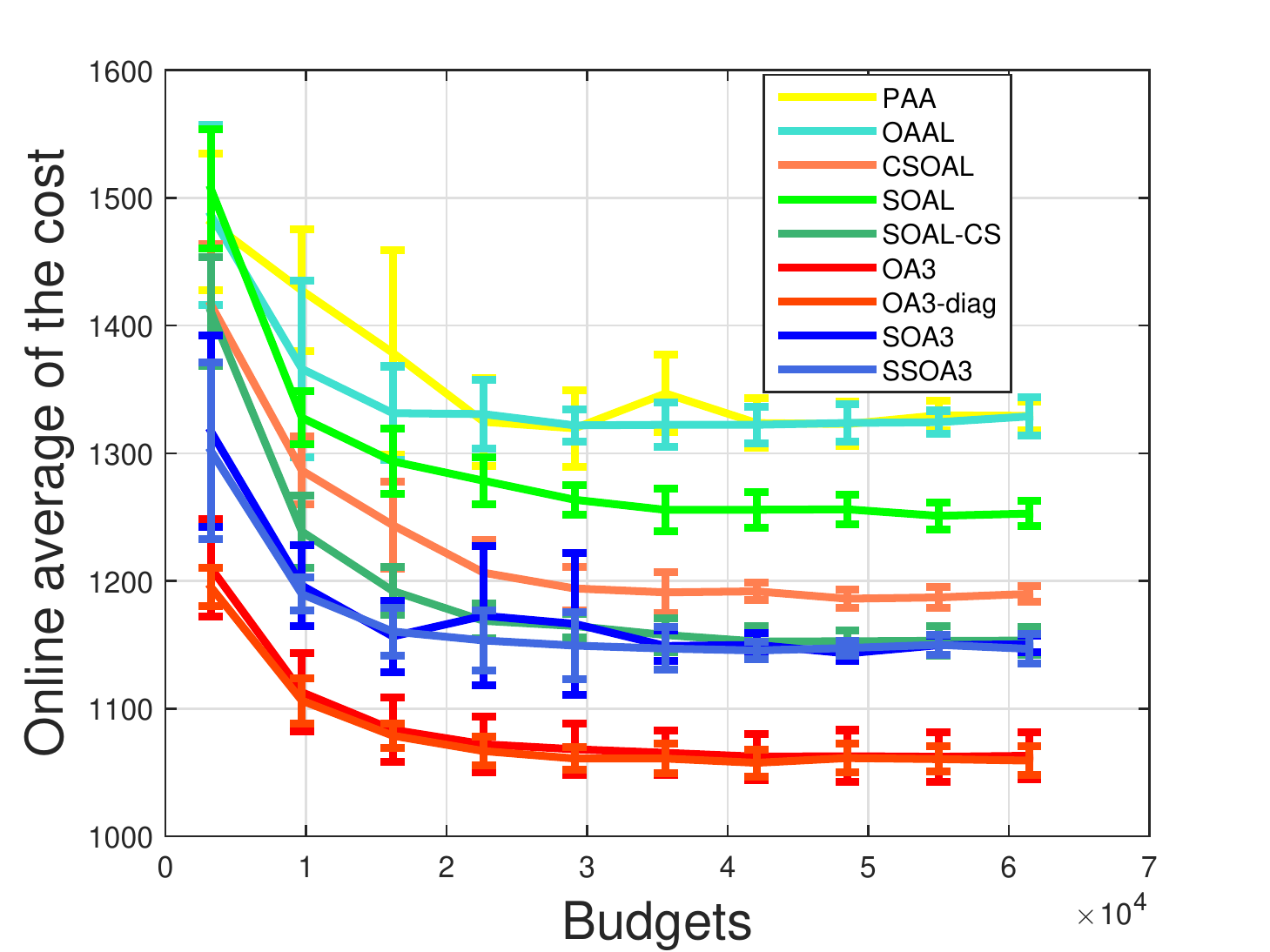}}
      \centerline{(c) w8a}
    \end{minipage}
    \hfill
    \begin{minipage}{0.485\linewidth}
      \centerline{\includegraphics[width=4.7cm]{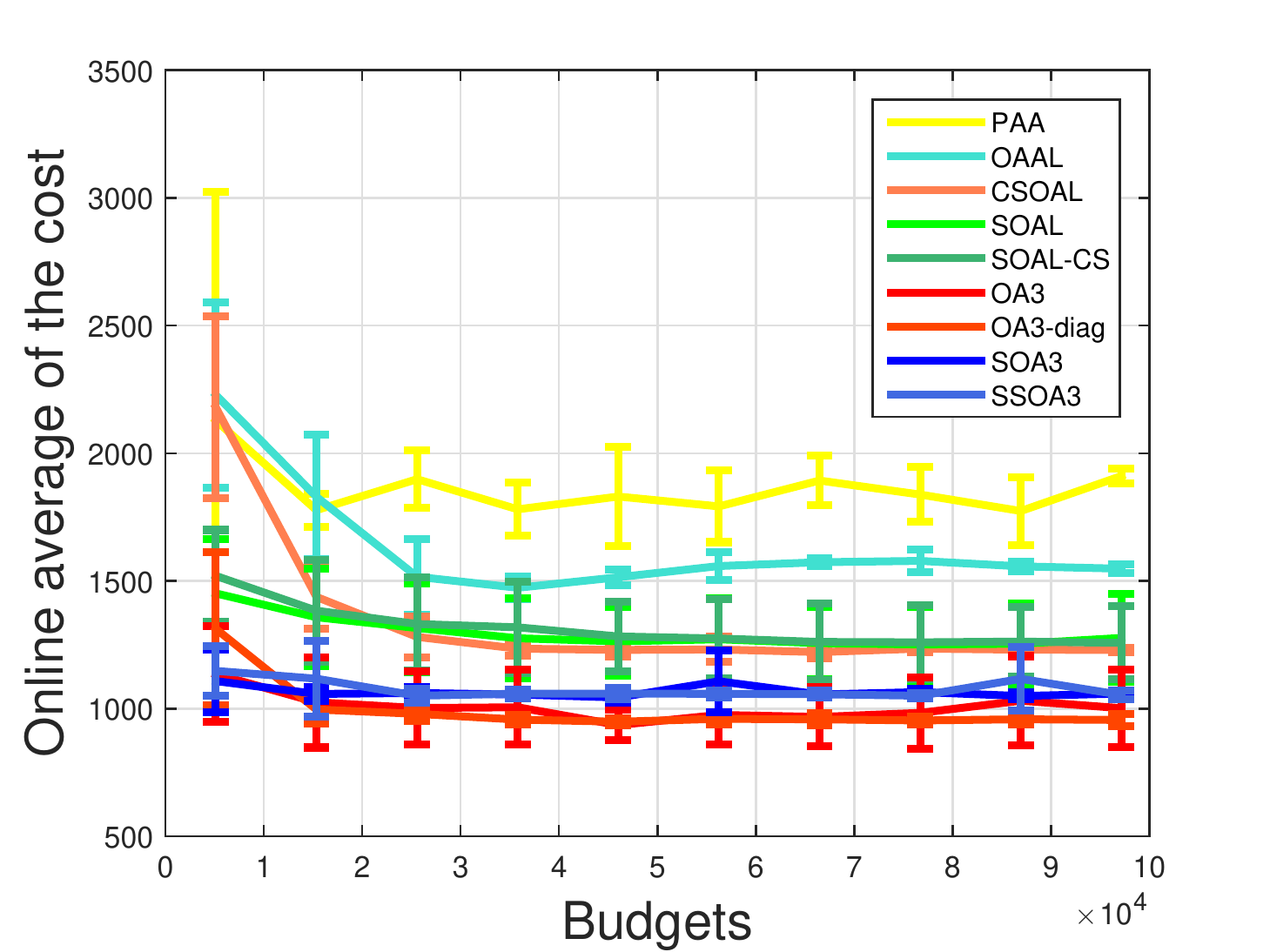}}
      \centerline{(d) KDDCUP08}
    \end{minipage}
    \vspace{-0.1in}
    \caption{Evaluation of cost with varying budgets.}\label{cost_varying}
    \vspace{-0.15in}
\end{figure}

\shuai{In this subsection, we compare the performance of all algorithms with varying query budgets. Figs.~\ref{sum_varying} and \ref{cost_varying} show the results in terms of \emph{sum} and \emph{cost} respectively.}

In detail, our algorithms achieve good performance over a wide range of budgets in terms of both metrics. \yifan{This observation further demonstrates the effectiveness and stability of our algorithms. Moreover, it suggests that our algorithms can help real-world companies with different labeling budgets.}

\shuai{In addition, when the query budget falls, the standard deviation of each algorithm \revise{increases}. This observation} implies that the randomness of samples plays an important role in performance, especially when the budget is limited, which validates the importance of query strategies.

\vspace{-0.1in}
\subsection{Evaluation of Algorithm Properties}
We have evaluated performance of the proposed algorithms in previous experiments, where promising results confirm their superiority. Next, we further examine their unique properties, including the influence of query biases, cost weights and learning rates. The examinations contribute to better understanding and applications of the proposed methods. Due to the page limitation, we only report the results of $sum$ metric, while more results based on $cost$ metric are put in Supplementary C.
Each experiment focuses on only one variable, while all other variable settings are fixed and similar with previous experiments.

\subsubsection{Evaluation Between Query Biases}

\yifan{We first} examine the influences of query biases on \yifan{OA3 under a limited budget}. All query biases ($\delta_{+}$, $\delta_{-}$) are selected from $[10^{-5},10^{-4},...,10^{4},10^{5}]$.

\begin{figure}
    \begin{minipage}{0.485\linewidth}
      \centerline{\includegraphics[width=4.7cm]{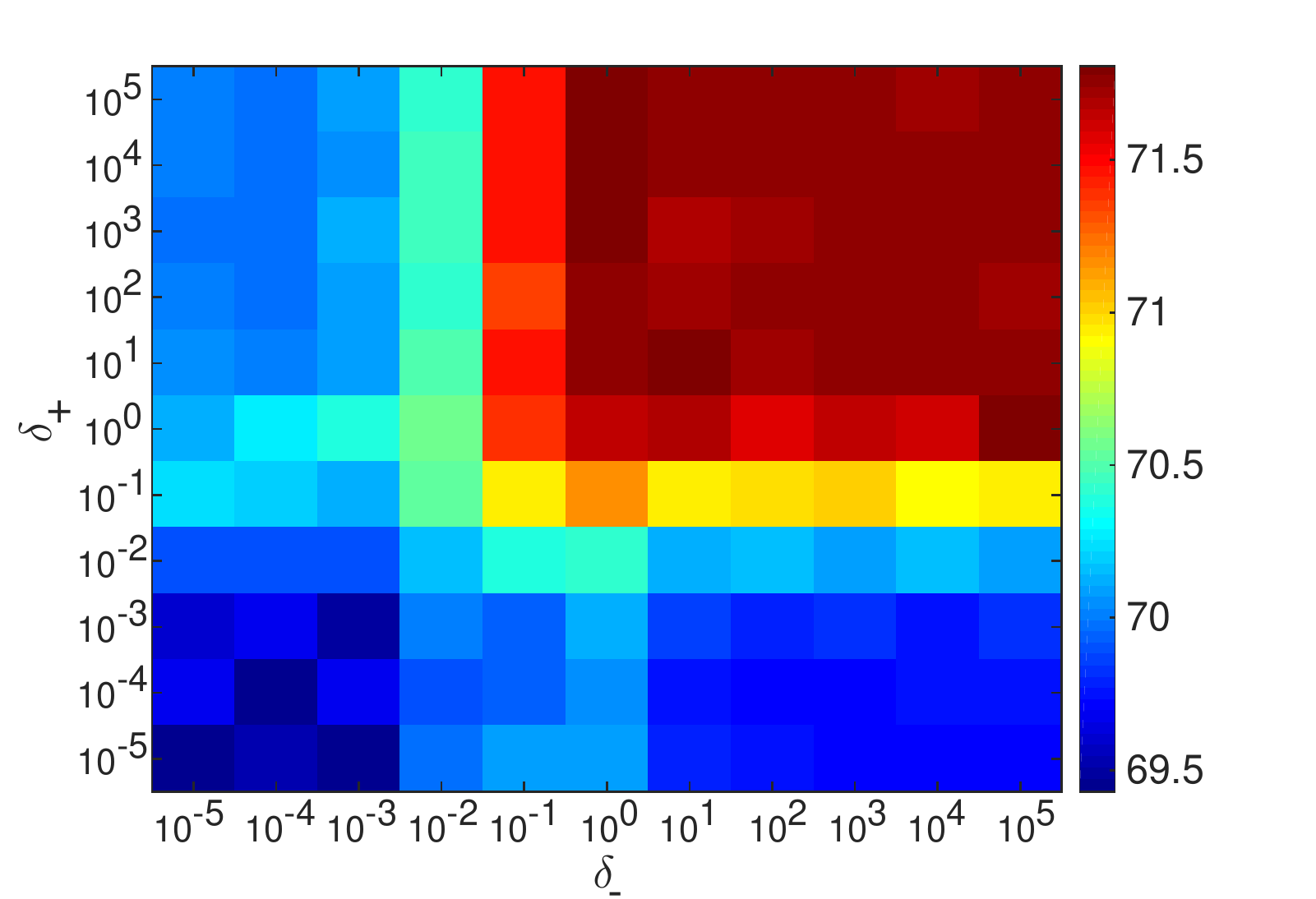}}
      \centerline{(a) protein, B=8500}
    \end{minipage}
    \hfill
    \begin{minipage}{0.485\linewidth}
      \centerline{\includegraphics[width=4.7cm]{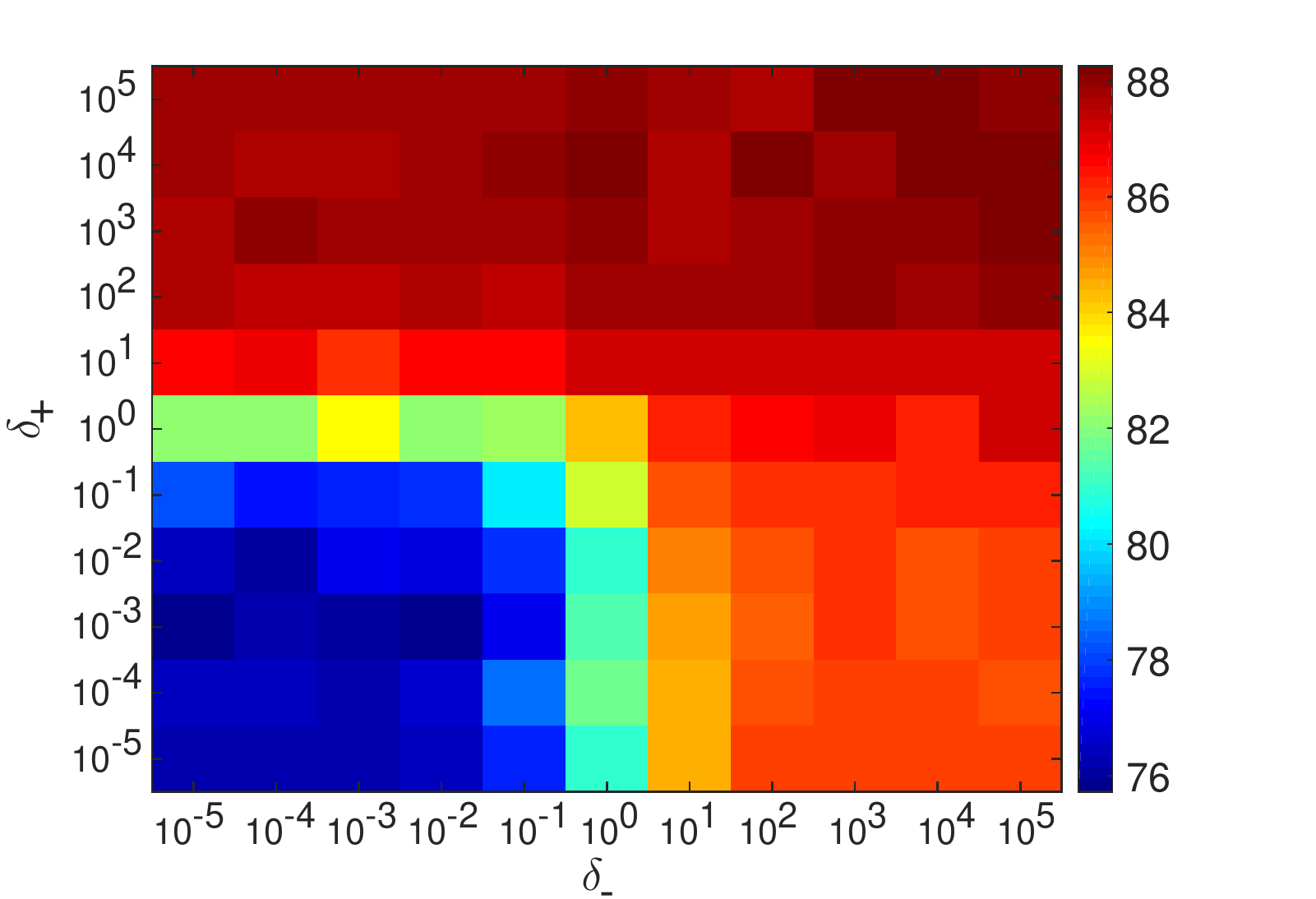}}
      \centerline{(b) Sensorless, B=29000}
    \end{minipage}
    \vfill
    \begin{minipage}{0.485\linewidth}
      \centerline{\includegraphics[width=4.7cm]{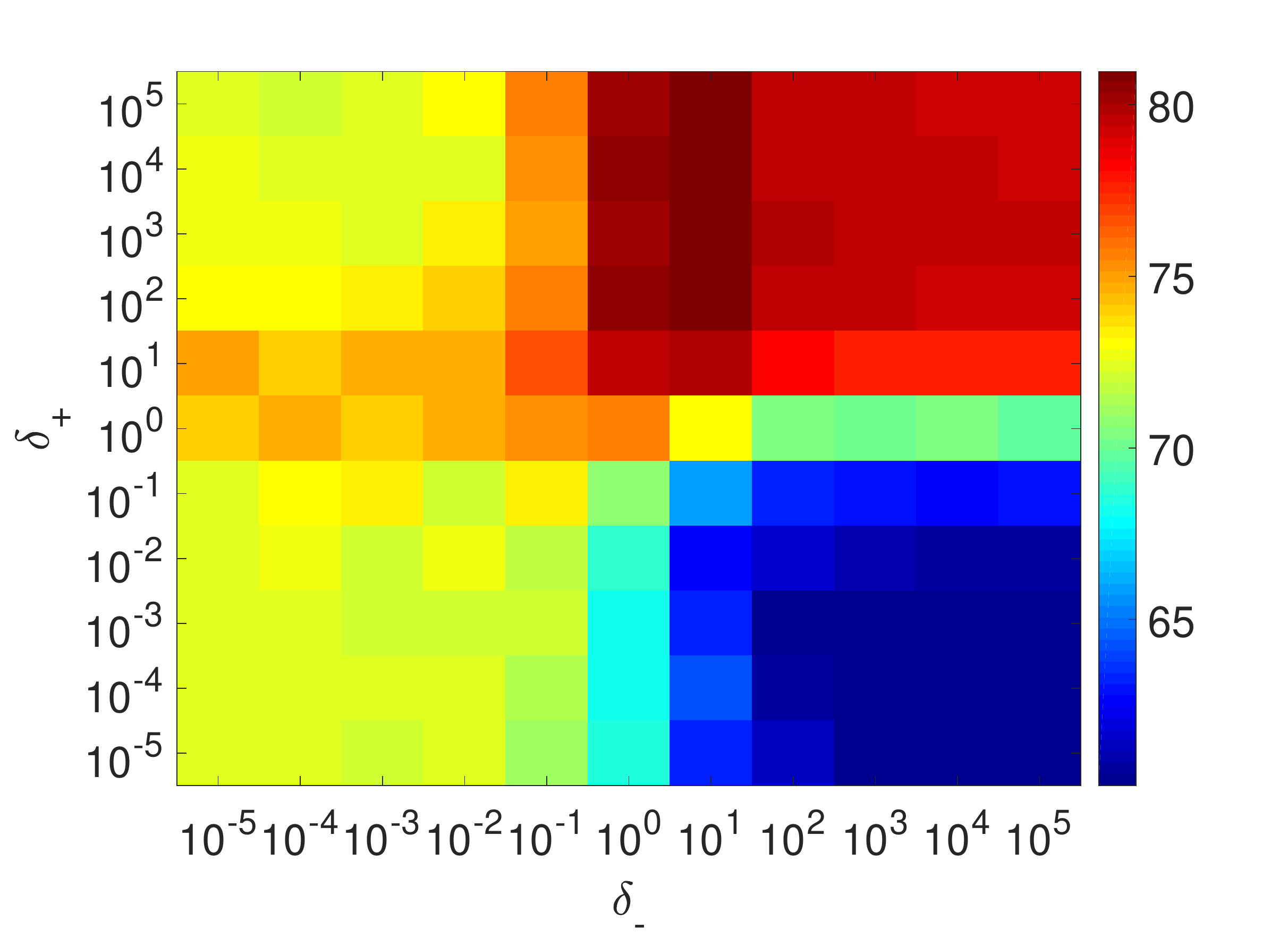}}
      \centerline{(c) w8a, B=32000}
    \end{minipage}
    \hfill
    \begin{minipage}{0.485\linewidth}
      \centerline{\includegraphics[width=4.7cm]{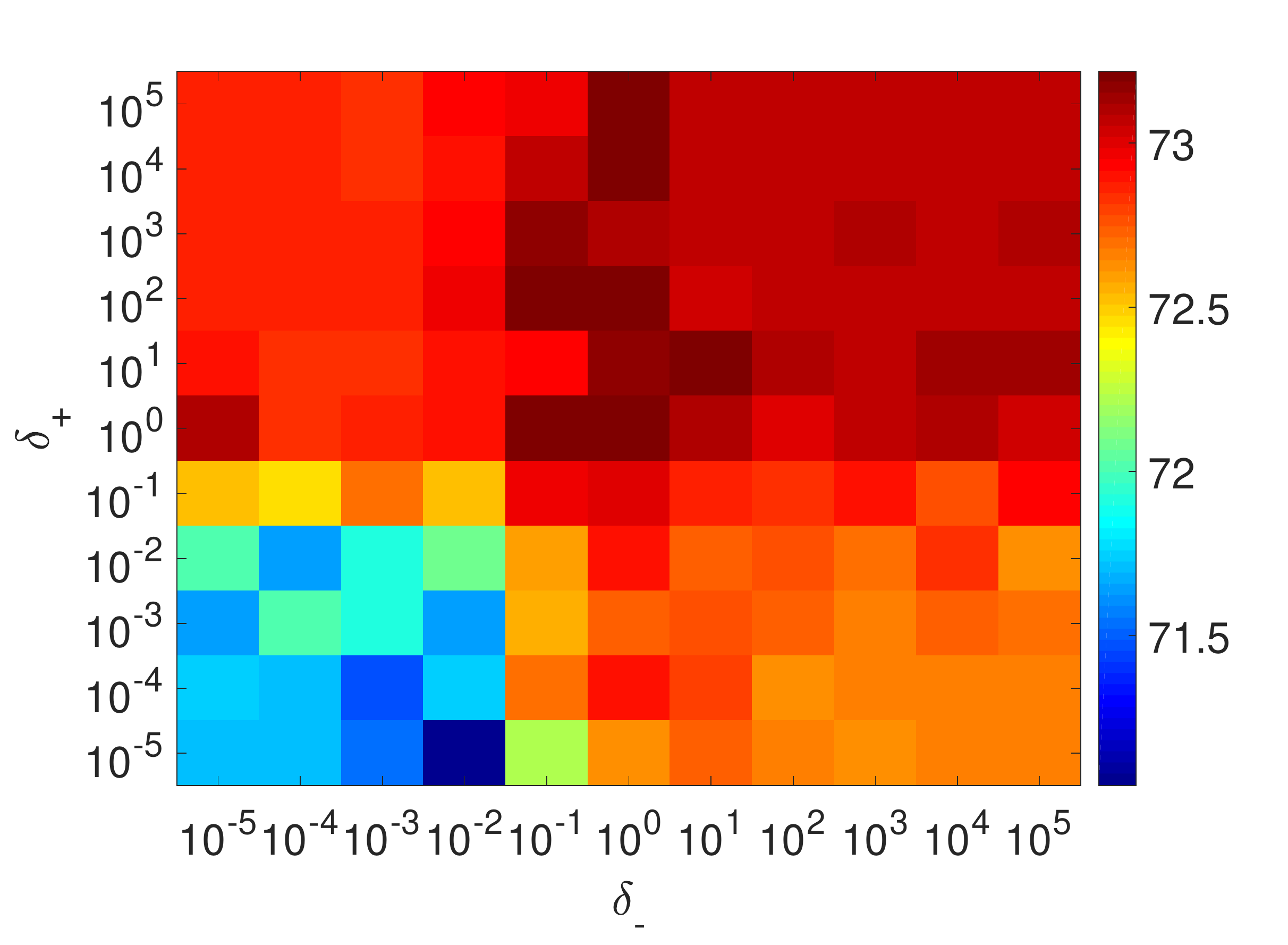}}
      \centerline{(d) KDDCUP08, B=100000}
    \end{minipage}
    \vspace{-0.1in}
    \caption{Performance of sum with varying query parameters.}\label{sum_query_biases}
    \vspace{-0.15in}
\end{figure}

The best results (\emph{\ie,} deep red color in Fig.~\ref{sum_query_biases}) are often obtained when $\delta_{+} \small{\in}\{10^2,10^3,10^4,10^5\}$ and $\delta_{-} \small{\in} \{1,10\}$. This suggests, when querying more samples with the positive prediction (\emph{\ie,} $\delta_{+} \geq \delta_{-}$), OA3 can achieve \yifan{better results}. In other words, when paying more attention to positive predictions, the model will query more positive samples which are more informative in imbalanced tasks.

This observation is different from the discussions in OAAL \cite{Zhang2016Online}, where the authors argued that $\delta_{-}$ should be larger than $\delta_{+}$. The main difference is that our algorithm  considers asymmetric strategies \yifan{in both optimization and queries; while OAAL considers only the asymmetric query. As a result, our method can query more positive samples (\ie, minority) due to the algorithm characteristics.}

In addition, when both $\delta_{+}$ and $\delta_{-}$ are large (\emph{\ie,} the upper right corners in Fig.~\ref{sum_query_biases}), our algorithms achieve fairly good performance. In this setting, the algorithm tends to query each observed sample and degrades to the ``First come first served" strategy. This means that our algorithms with weak query strategy can also perform well. Moreover, when both $\delta_{+}$ and $\delta_{-}$ are small (\emph{\ie,} the bottom left corner), the model tends to ignore the samples, so the algorithm performance \revise{decreases} significantly.

\subsubsection{Evaluation of Cost Weights}

\begin{figure}
    \begin{minipage}{0.485\linewidth}
      \centerline{\includegraphics[width=4.7cm]{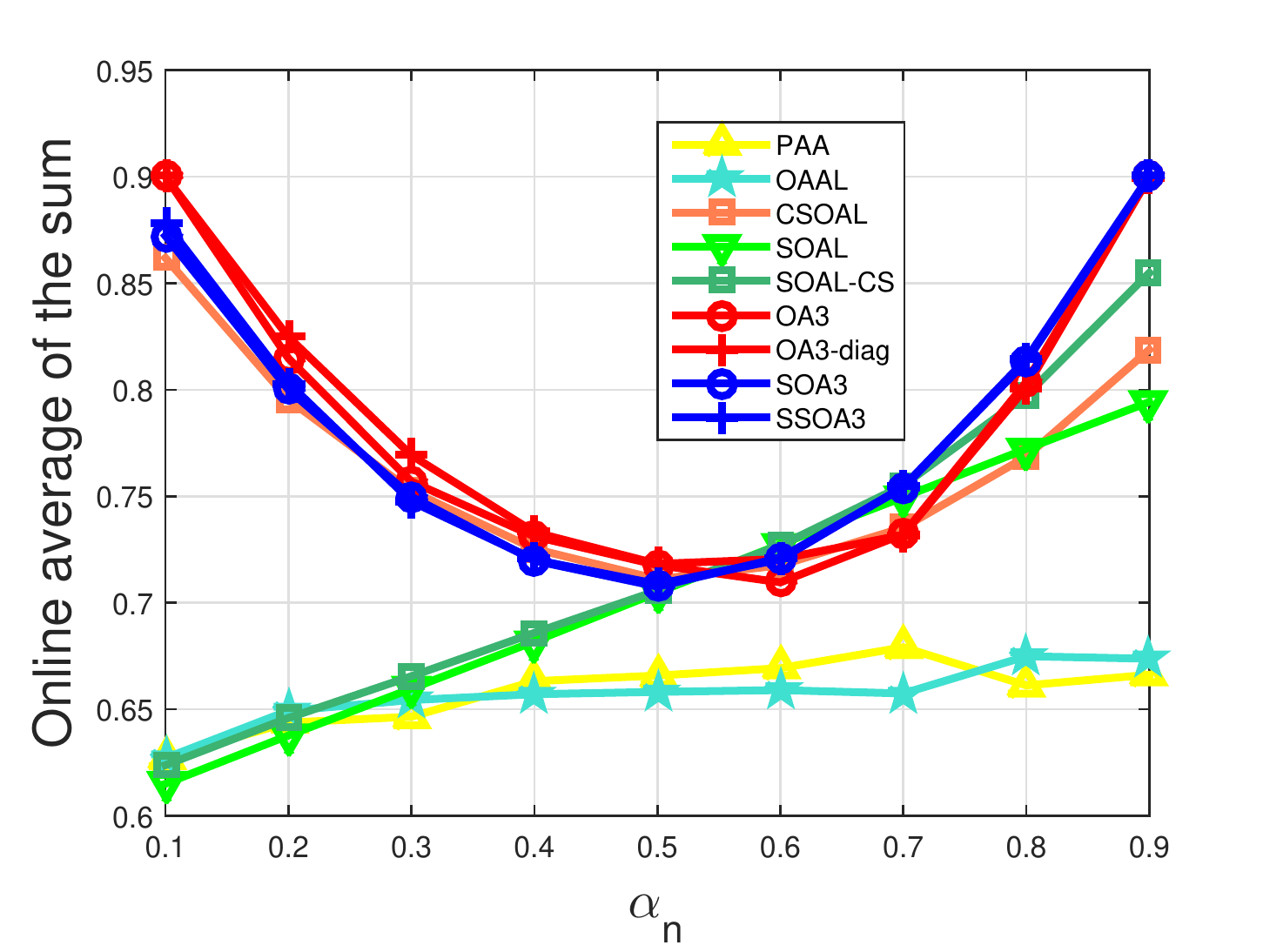}}
      \centerline{(a) protein, B=8500}
    \end{minipage}
    \hfill
    \begin{minipage}{0.485\linewidth}
      \centerline{\includegraphics[width=4.7cm]{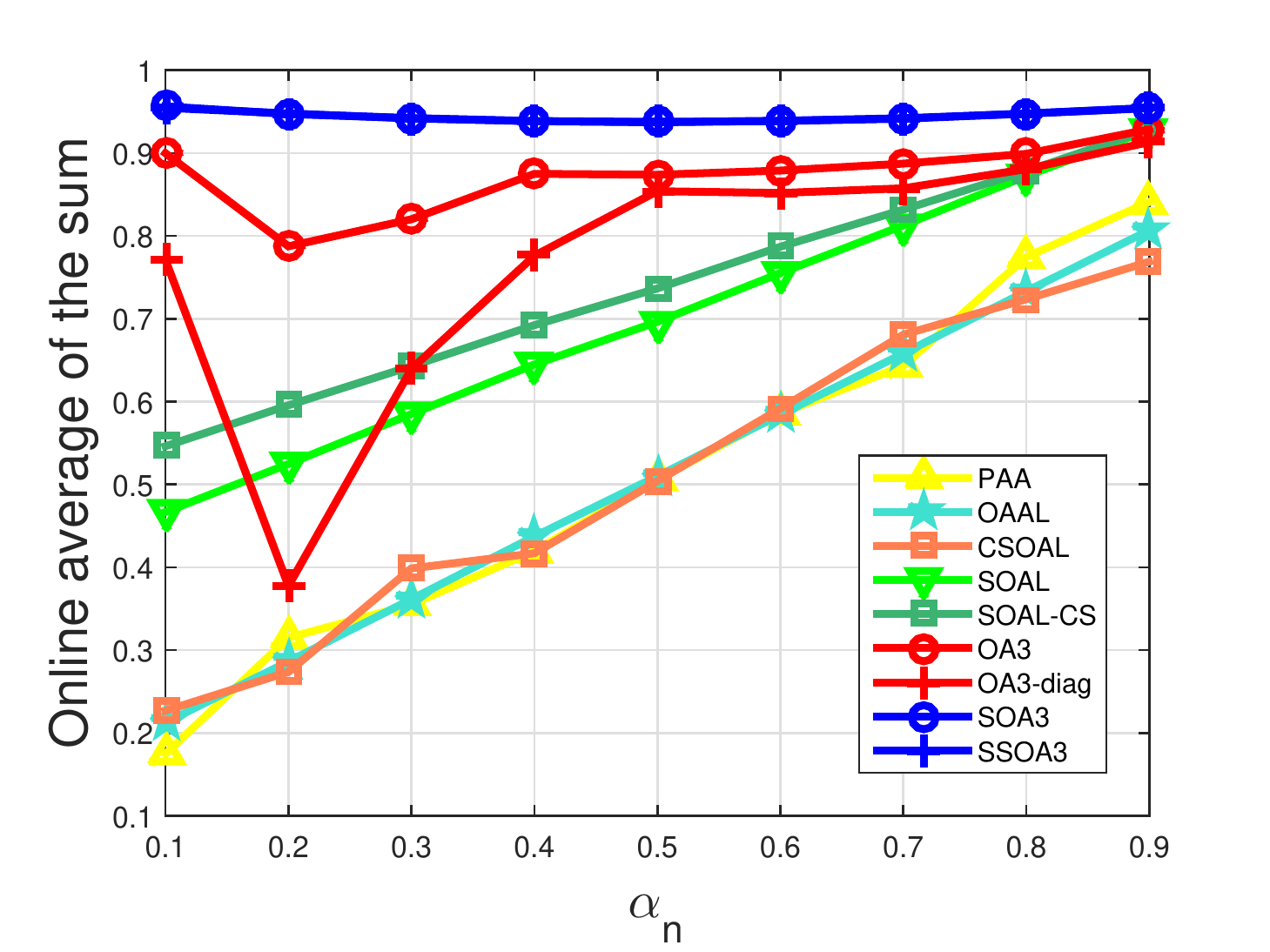}}
      \centerline{(b) Sensorless, B=29000}
    \end{minipage}
    \vfill
    \begin{minipage}{0.485\linewidth}
      \centerline{\includegraphics[width=4.7cm]{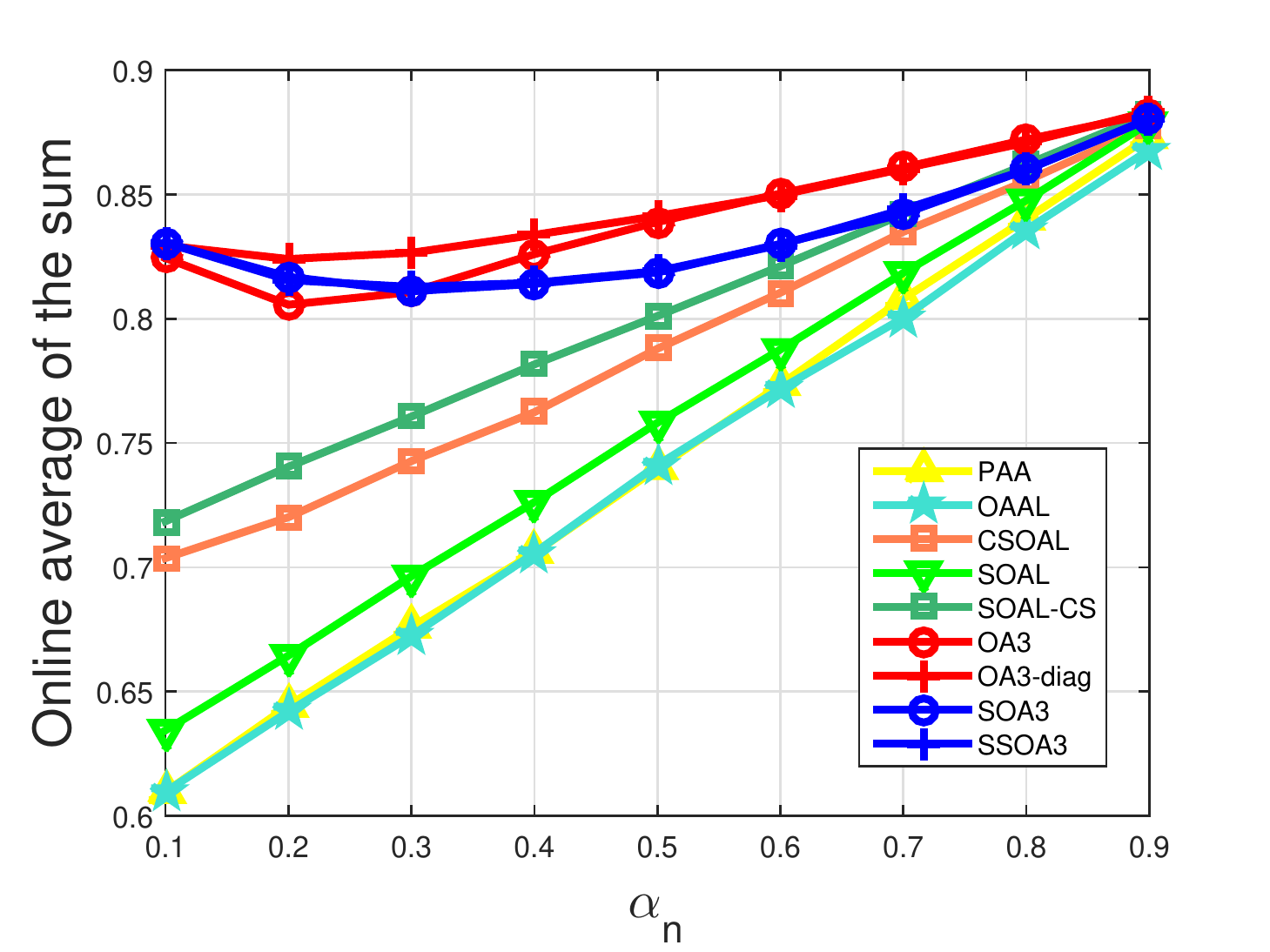}}
      \centerline{(c) w8a, B=32000}
    \end{minipage}
    \hfill
    \begin{minipage}{0.485\linewidth}
      \centerline{\includegraphics[width=4.7cm]{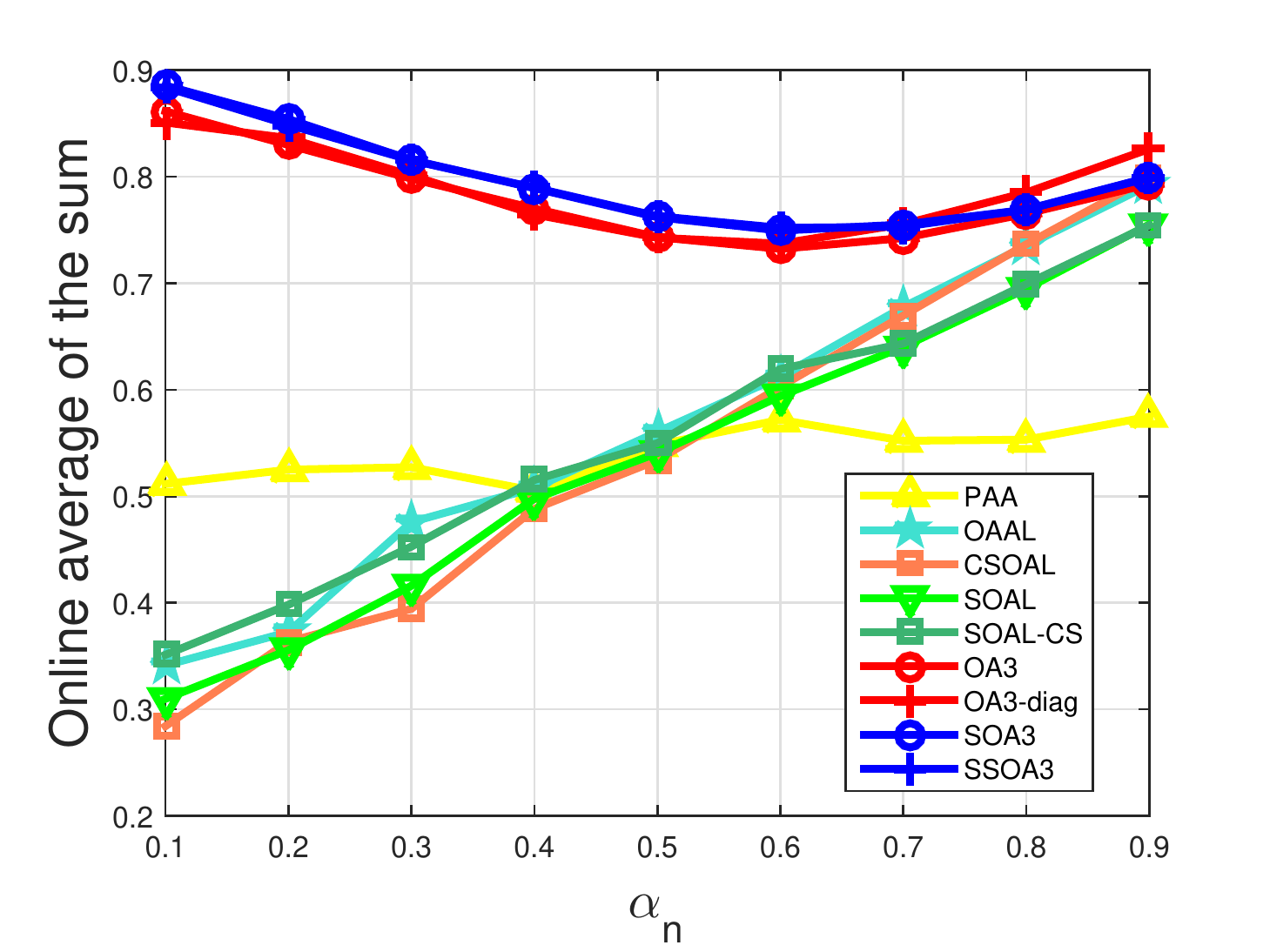}}
      \centerline{(d) KDDCUP08, B=50000}
    \end{minipage}
    \vspace{-0.1in}
    \caption{Performance of sum \yifan{with} varying cost weights.}\label{sum_alphas}
    \vspace{-0.15in}
\end{figure}

In this subsection, we evaluate the influence of different cost weights, \ie, $\alpha_n$, where $\alpha_p \small{=} 1\small{-} \alpha_n$. Fig.~\ref{sum_alphas} summarizes the results of $sum$ metric under a fixed budget.
We find that our proposed algorithms consistently outperform all other algorithms with different weights. This observation shows that OA3 based algorithms have a wide selection range of cost weights, which further validates the effectiveness of the proposed methods.

\vspace{-0.1in}
\subsubsection{Evaluation of Learning Rates}

\yifan{We next} evaluate the influence of \shuai{different learning rates to the proposed methods}, where the learning rate $\eta$ is selected from $[10^{-4}, 10^{-3}, ..., 10^{3},10^{4}]$.

\begin{figure}
    \begin{minipage}{0.485\linewidth}
      \centerline{\includegraphics[width=4.7cm]{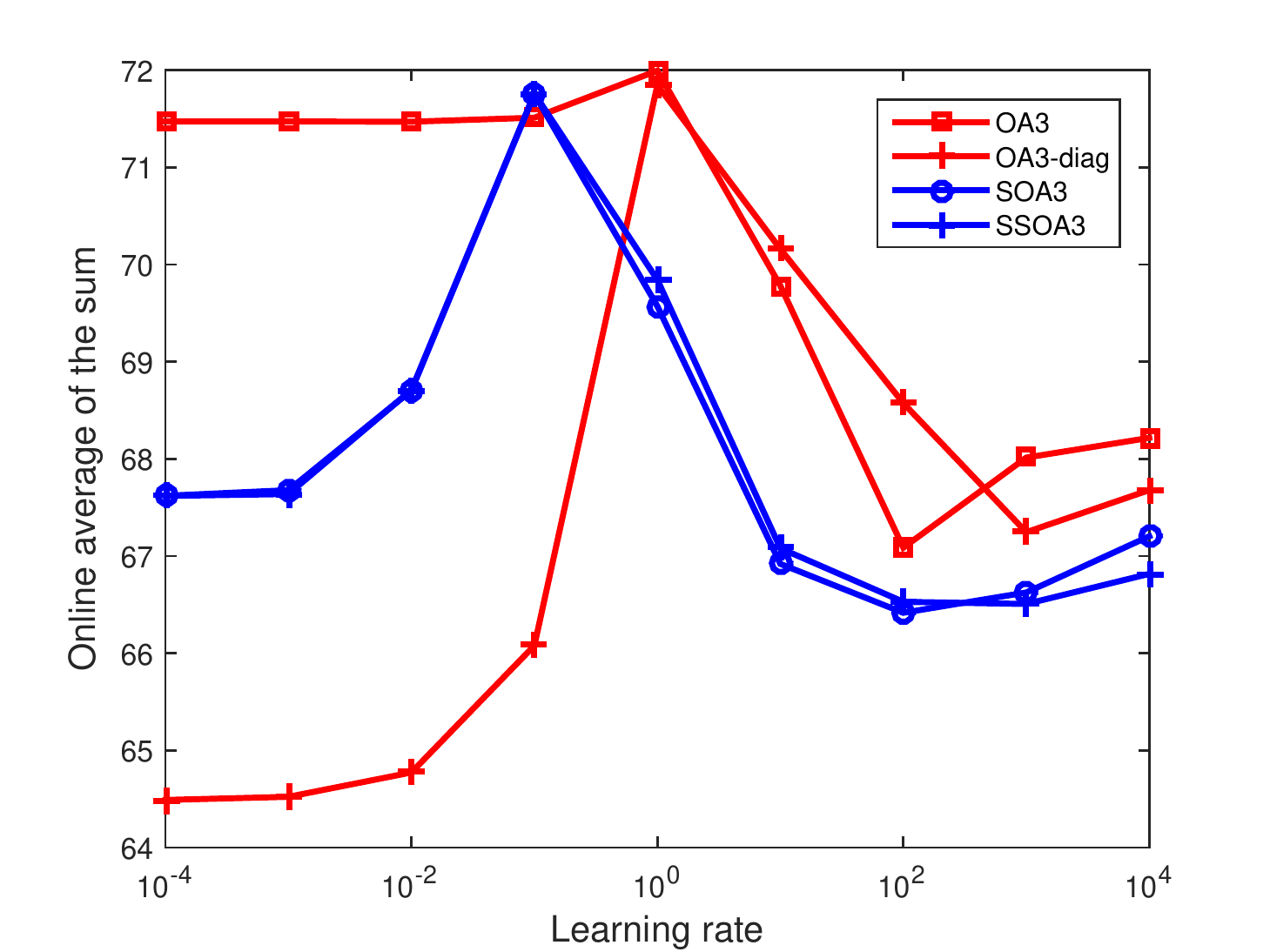}}
      \centerline{(a) protein, B=8500}
    \end{minipage}
    \hfill
    \begin{minipage}{0.485\linewidth}
      \centerline{\includegraphics[width=4.7cm]{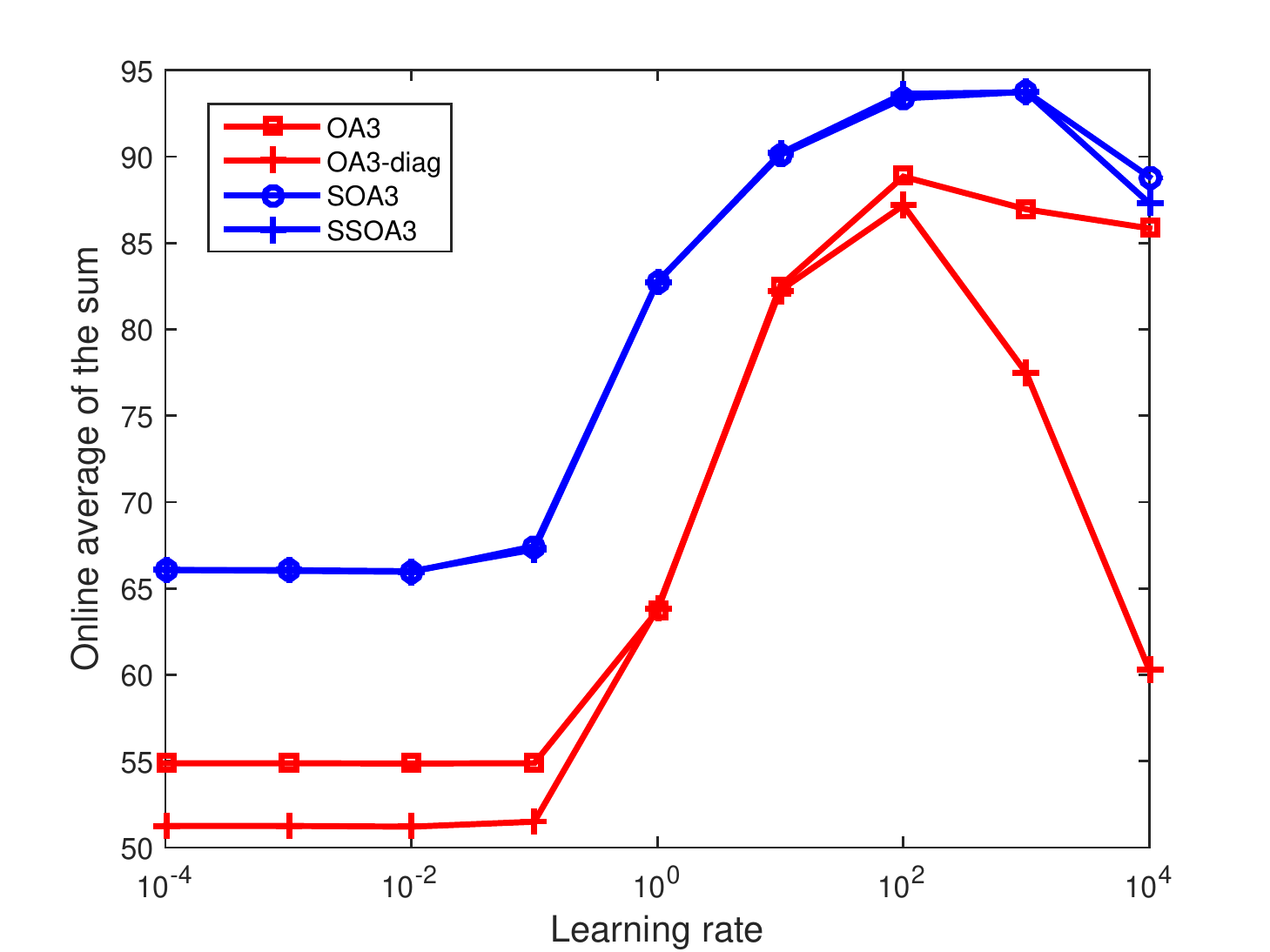}}
      \centerline{(b) Sensorless, B=29000}
    \end{minipage}
    \vfill
    \begin{minipage}{0.485\linewidth}
      \centerline{\includegraphics[width=4.7cm]{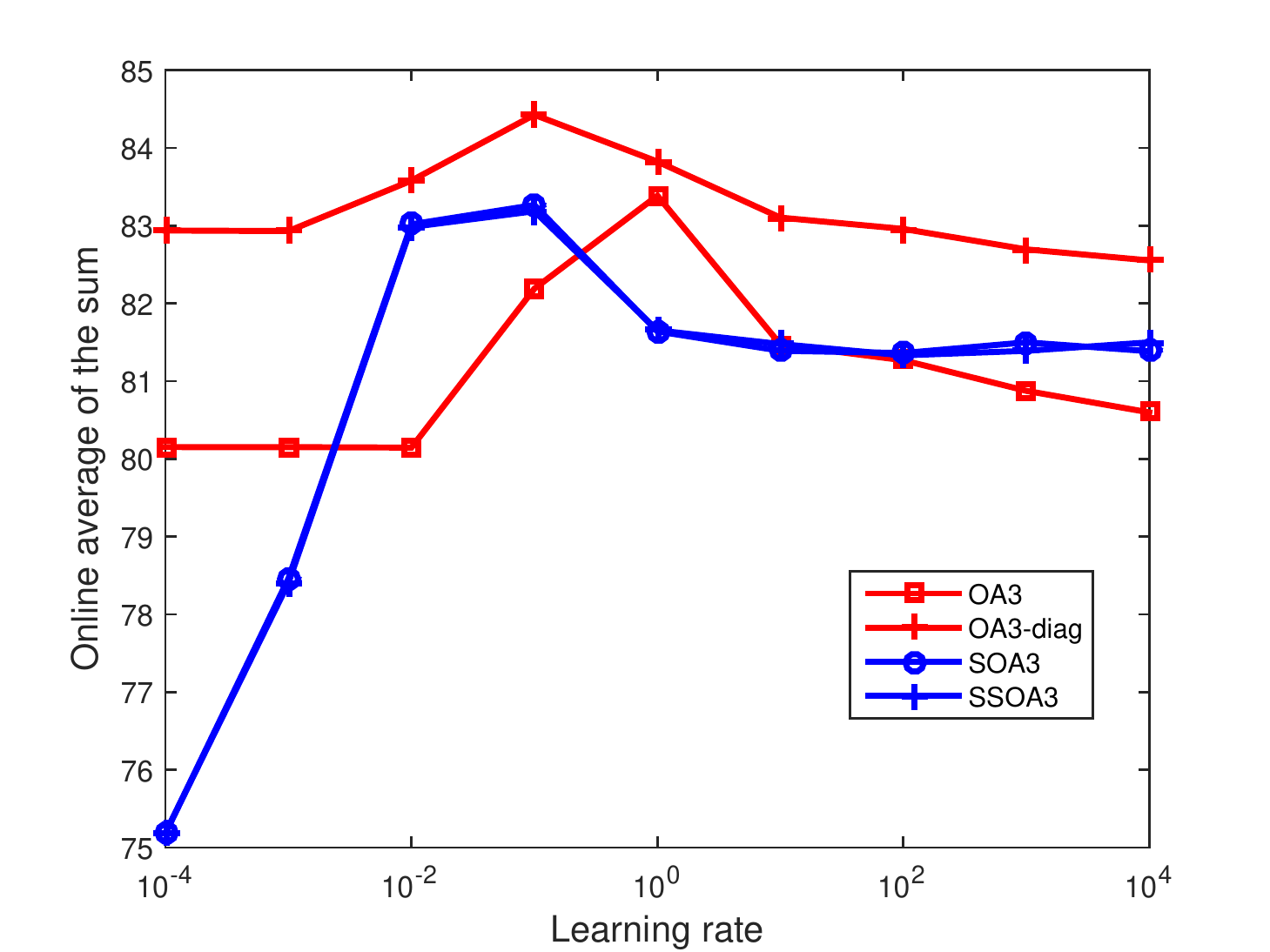}}
      \centerline{(c) w8a, B=32000}
    \end{minipage}
    \hfill
    \begin{minipage}{0.485\linewidth}
      \centerline{\includegraphics[width=4.7cm]{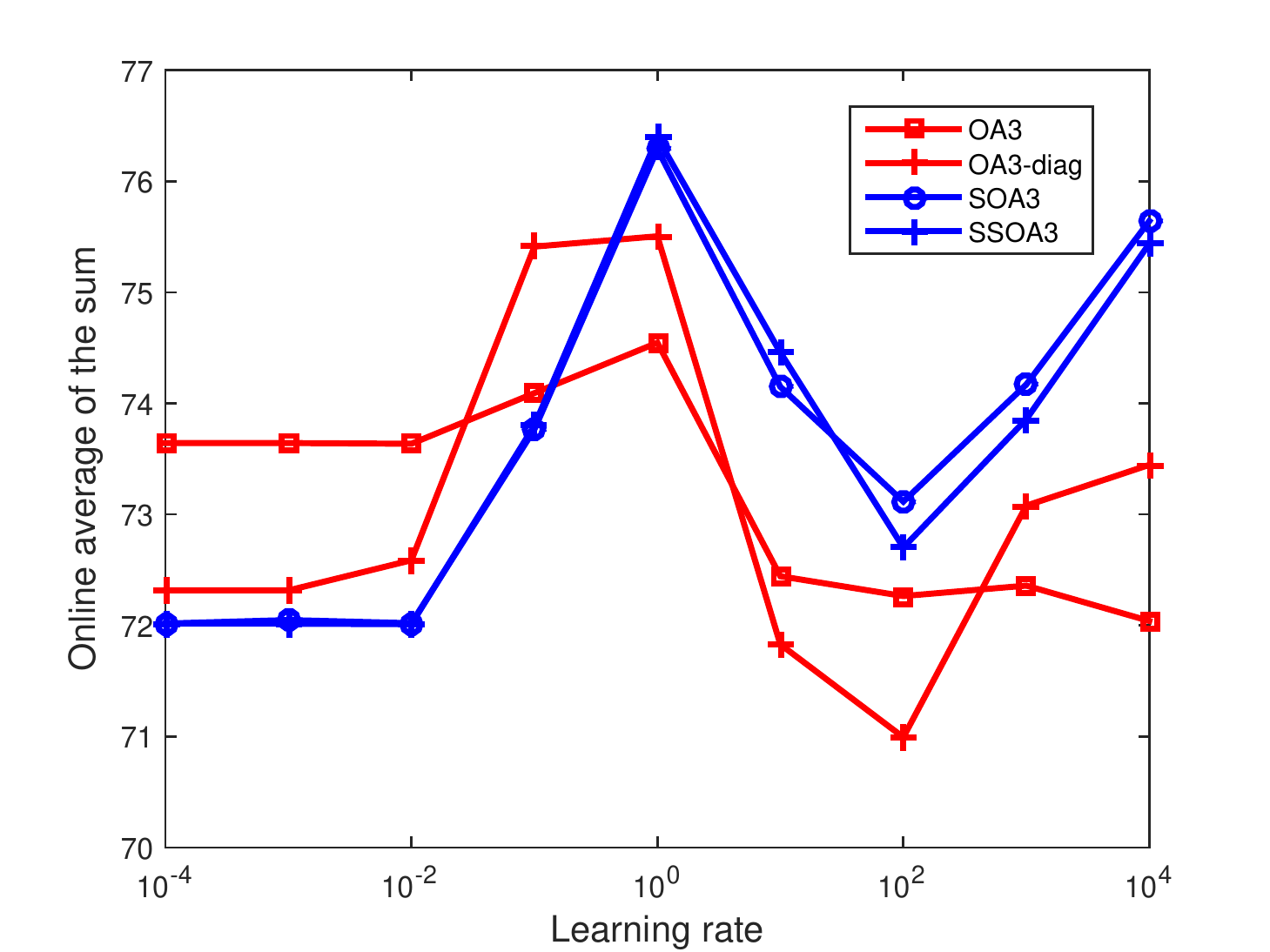}}
      \centerline{(d) KDDCUP08, B=50000}
    \end{minipage}
    \vspace{-0.1in}
    \caption{Performance of sum with varying learning rates.}\label{sum_learning_rates}
    \vspace{-0.15in}
\end{figure}

Fig.~\ref{sum_learning_rates} shows a suitable range of learning rates for different datasets, which provides a candidate choice of the learning rate for algorithm engineers.
To be specific, OA3 algorithms achieve the best result on most datasets. Moreover, SOA3 and SSOA3 perform well on most datasets and sometimes even better than OA3. Considering that SOA3 and SSOA3 are more efficient than OA3, we conclude that the sketched versions of OA3 are favorable choices to balance performance and efficiency.

\vspace{-0.1in}
\subsection{Evaluation on High Dimensional Datasets}\label{evaluate_high_dimensional}
\revise{In this subsection, we further evaluate our  methods on two higher dimensional datasets (CIFAR-10\footnote{https://www.cs.toronto.edu/\textasciitilde kriz/cifar.html} and Internet Advertisements\footnote{https://archive.ics.uci.edu/ml/datasets.php} (IAD)).  The dimensionality of CIFAR-10 is 3072 and that of IAD is 1558. To be specific, we construct the CIFAR-10 dataset by randomly sampling $20\%$ samples from the CIFAR-10 training set, and randomly take one class as the positive class and set other classes as the negative class. All images ($\mathbb{R}^{32\times32\times 3}$) are squeezed to the vectors ($\mathbb{R}^{3072}$). In addition, we clean IAD by removing the samples that have null attributes. Other settings are the same as the previous experiments. Due to the page limits, we only report the results of \textit{sum} metric, while more results based on \textit{cost} metric and additional analysis are put in Supplementary C.}

\revise{As shown in Table~\ref{high_d_performance}, OA3-based algorithms achieve the best performance on both datasets. However, the running time of OA3 is much longer than first-order methods. In contrast, the proposed sketched variants (i.e., SOA3 and SSOA3) are much faster than OA3. Meanwhile, they also perform very well and sometimes even better than OA3. In this sense, SOA3 and SSOA3 are favorable choices when handling real-world  high-dimensional datasets.}
\begin{table*}[t]
	\caption{\revise{Sum evaluation on high-dimensional datasets}}\label{high_d_performance}
    \begin{center}
    \begin{scriptsize}
    \scalebox{0.92}{
    \renewcommand*\arraystretch{1.2}
	\begin{tabular}{|l|c|c|c|c|c|c|c|c|}\hline

        \multirow{2}{*}{Algorithm}&\multicolumn{4}{c|}{$``sum``$ on CIFAR-10} &  \multicolumn{4}{c|}{$``sum``$ on IAD}\cr\cline{2-9}
        &Sum(\%)&Sensitivity(\%)&Specificity  (\%)&Time(s)   & Sum(\%)&Sensitivity(\%)&Specificity  (\%)&Time(s)\cr
        \hline \hline
        PAA         &54.271 	$\pm$ 2.255 	& 18.060 	$\pm$ 15.303 	& 90.482 	$\pm$ 15.269 	& 0.513 	
        &52.570 	$\pm$ 3.171 	& 14.216 	$\pm$ 12.900 	& 90.924 	$\pm$ 6.777 	& 0.064  \\
        OAAL  &63.510 	$\pm$ 0.525 	& 60.149 	$\pm$ 1.977 	& 66.872 	$\pm$ 1.363 	& 0.475  	
        &58.965 	$\pm$ 1.420 	& 55.123 	$\pm$ 1.865 	& 62.807 	$\pm$ 2.379 	& 0.061  	\\
        CSOAL  &56.331 	$\pm$ 3.485 	& 21.552 	$\pm$ 14.413 	& 91.110 	$\pm$ 8.913 	& 0.466  	
        &55.783 	$\pm$ 5.938 	& 19.338 	$\pm$ 13.048 	& 92.228 	$\pm$ 1.235 	& 0.059  \\
        SOAL  &56.161 	$\pm$ 0.765 	& 18.229 	$\pm$ 2.197 	& \textbf{94.092 	$\pm$ 0.863} 	& 529.586  	
        &74.496 	$\pm$ 5.005 	& 52.892 	$\pm$ 10.550 	& \textbf{96.099 	$\pm$ 0.984} 	& 21.655  \\
        SOAL-CS  &60.133 	$\pm$ 0.839 	& 27.582 	$\pm$ 1.993 	& 92.685 	$\pm$ 0.532 	& 584.077
        &78.259 	$\pm$ 0.830 	& 63.235 	$\pm$ 2.127 	& 93.282 	$\pm$ 1.162 	& 28.633 \\
        OA3  &\textbf{72.858 	$\pm$ 0.561} 	& \textbf{79.393 	$\pm$ 3.627} 	& 66.322 	$\pm$ 3.994 	& 1948.292   &80.398 	$\pm$ 0.624 	& 76.201 	$\pm$ 3.238 	& 84.595 	$\pm$ 2.594 	& 44.674   \\
        OA3$_{diag}$  &64.992 	$\pm$ 1.830 	& 64.448 	$\pm$ 8.104 	& 65.535 	$\pm$ 6.215 	& 152.623   	&81.359 	$\pm$ 0.514 	& 79.926 	$\pm$ 2.227 	& 82.792 	$\pm$ 2.307 	& 10.355   \\
        SOA3  &67.959 	$\pm$ 2.945 	& 69.463 	$\pm$ 8.768 	& 66.456 	$\pm$ 10.743 	& 66.739   	
        &81.484 	$\pm$ 2.133 	& 85.515 	$\pm$ 6.608 	& 77.453 	$\pm$ 7.641 	& 4.084   \\
        SSOA3  &68.678 	$\pm$ 2.049 	& 70.308 	$\pm$ 6.538 	& 67.047 	$\pm$ 8.636 	& 52.126
        &\textbf{82.101 	$\pm$ 7.677} 	& \textbf{91.005 	$\pm$ 5.876} 	& 73.197 	$\pm$ 19.614 	& 3.215   \\

        \hline
	\end{tabular}}
    \end{scriptsize}
    \end{center}
\end{table*}

\section{Conclusion}

In this paper, we have proposed a novel online adaptive asymmetric active learning algorithm to handle imbalanced and unlabeled datastream under limited query budgets. Relying on samples' second-order information, we develop a new asymmetric strategy, which integrates both the asymmetric losses and \revise{query} strategies. We theoretically analyze the mistake and cost-sensitive metric bounds of the proposed algorithm, for the cases within budgets and over budgets.

To overcome the time-consuming problem of second-order methods, we further propose a sketch variant of our method, which can be developed as a sparse sketch approach.
We empirically evaluate the proposed algorithms in real-world datasets.
\revise{Promising results confirm the effectiveness, efficiency and stability of the proposed methods.
In the future, we will extend the linear classifier to a nonlinear one with kernel methods.}

\section{Acknowledgement}
This work was partially supported by National Natural Science Foundation of China (NSFC) (61602185, 61502177 and 61876208), key project of NSFC (No. 61836003), Program for Guangdong Introducing Innovative and Enterpreneurial Teams 2017ZT07X183, Guangdong Provincial Scientific and Technological Funds (2018B010107001, 2017B090901008, 2017A010101011, 2017B090910005), Pearl River S$\&$T Nova Program of Guangzhou 201806010081, Tencent AI Lab Rhino-Bird Focused Research Program (No. JR201902), CCF-Tencent Open Research Fund RAGR20170105, Program for Guangdong Introducing Innovative and Enterpreneurial Teams 2017ZT07X183.

%
%

%

\vspace{-11ex}
\begin{IEEEbiography}[{\includegraphics[width=1in,height=1.25in,clip,keepaspectratio]{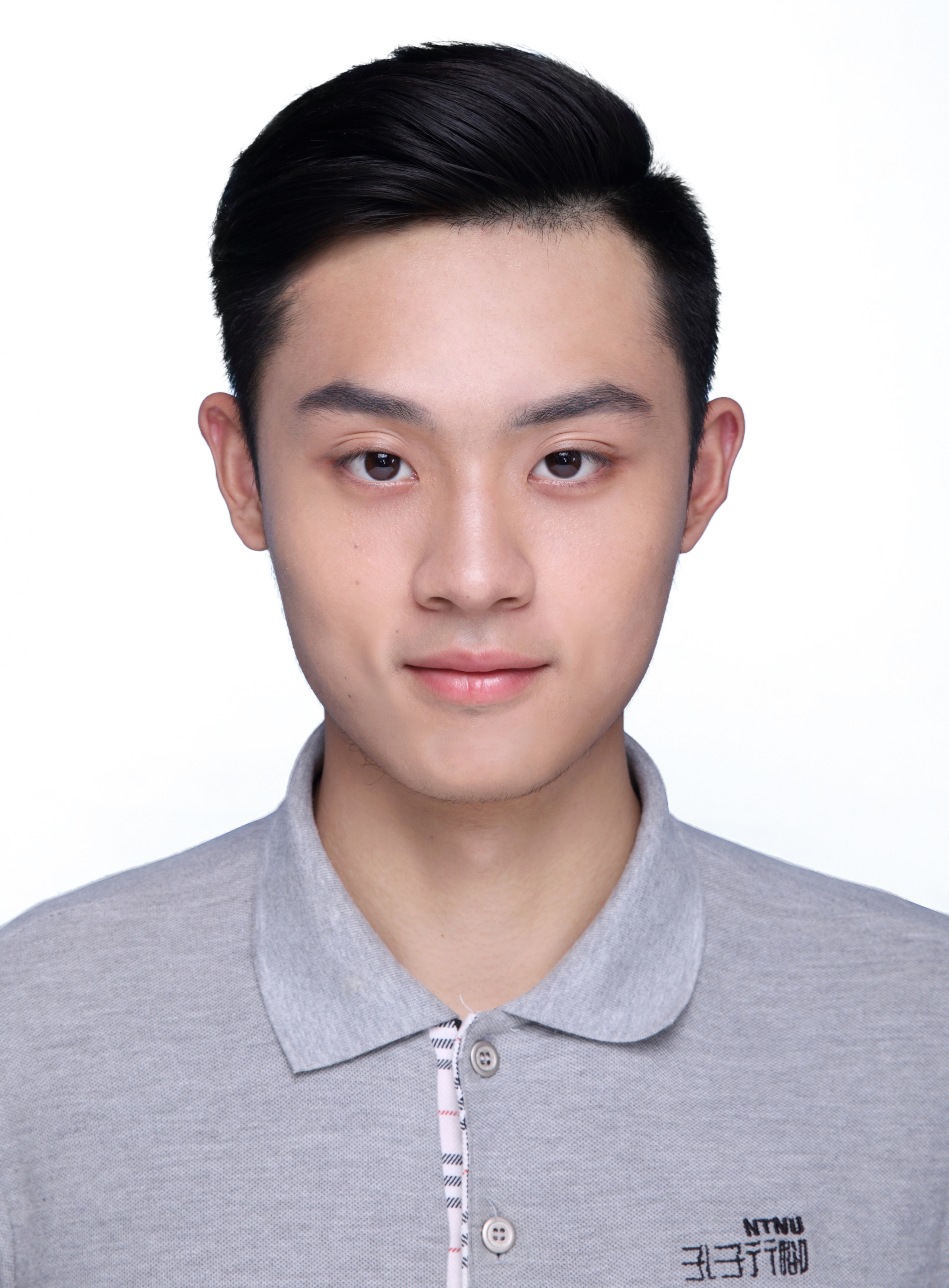}}]{Yifan Zhang}
is working toward the M.E. degree in the School of Software Engineering, South China University of Technology, China. He received the B.E. degree from the Southwest University, China, in 2017. His research interests are broadly in machine learning, with high self-motivation to design explainable and robust algorithms for data-limited problems and decision-making tasks.
\end{IEEEbiography}
\vspace{-10ex}
\begin{IEEEbiography}[{\includegraphics[width=1in,height=1.25in,clip,keepaspectratio]{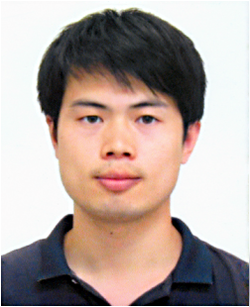}}]{Peilin Zhao}
is currently a Principal Researcher at Tencent AI Lab, China. Previously, he has worked at Rutgers University, Institute for Infocomm Research (I2R), Ant Financial Services Group. His research interests include: Online Learning, Deep Learning, Recommendation System, Automatic Machine Learning, etc. He has published over 90 papers in top venues, including JMLR, ICML, KDD, etc. He has been invited as a PC member, reviewer or editor for many international conferences and journals, such as ICML, JMLR, etc. He received his bachelor degree from Zhejiang University, and his Ph.D. degree from Nanyang Technological University.
\end{IEEEbiography}
\vspace{-10ex}
\begin{IEEEbiography}[{\includegraphics[width=1in,height=1.25in,clip,keepaspectratio]{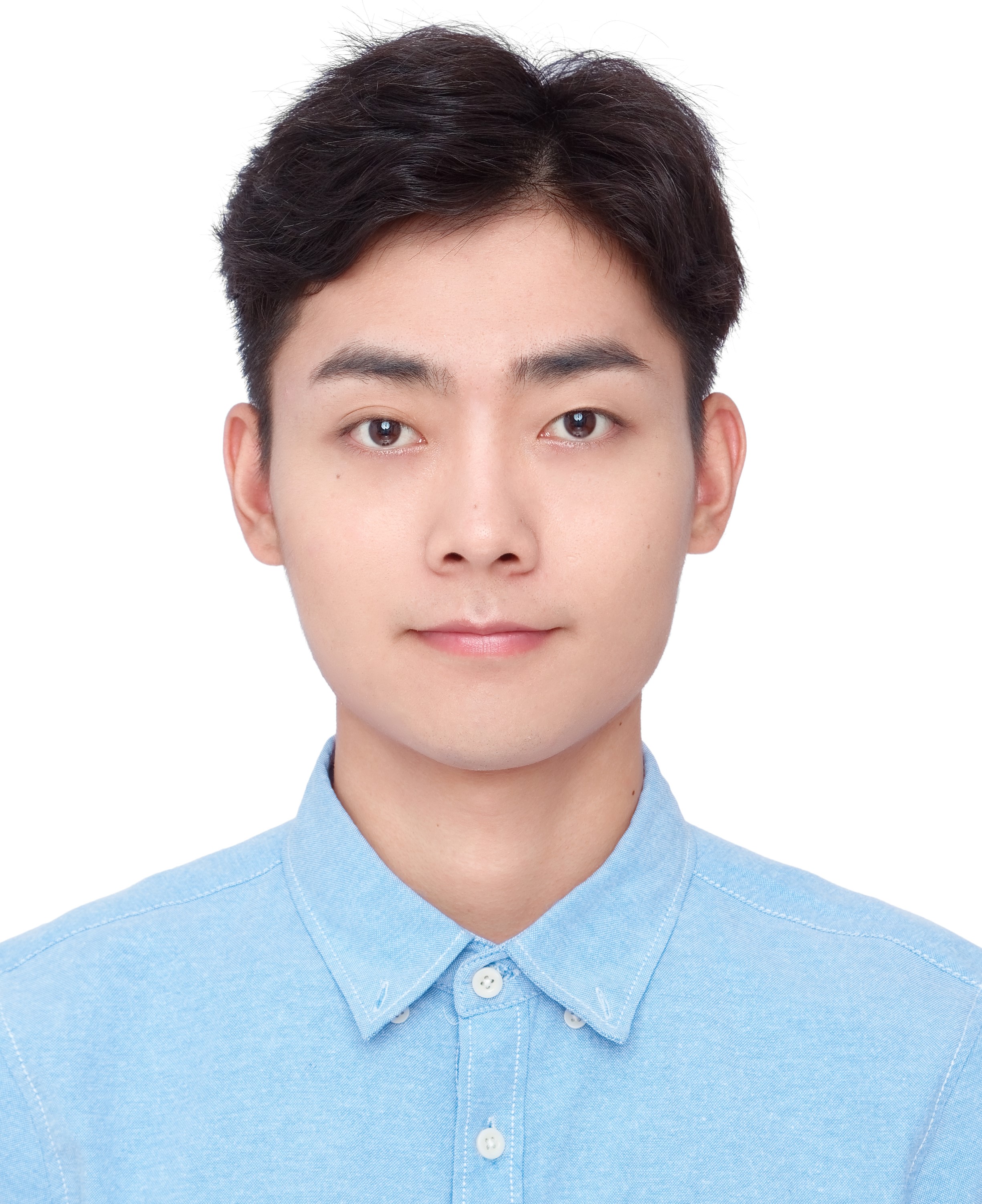}}]{Shuaicheng Niu}
is currently a Ph.D candidate in South China University of Technology (SCUT), China, School of Software Engineering. He received the Bachelor degree in mathematics from Southwest Jiaotong University (SWJTU), China, in 2018. His research interests include Machine Learning, Reinforcement Learning and Neural Network Architecture Search.
\end{IEEEbiography}
\vspace{-12ex}
\begin{IEEEbiography}[{\includegraphics[width=1in,height=1.2in,clip,keepaspectratio]{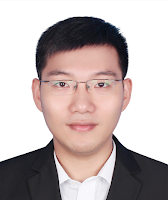}}]{Qingyao Wu}
received the Ph.D. degree in computer
science from the Harbin Institute of Technology,
Harbin, China, in 2013. He was a Post-
Doctoral Research Fellow with the School of
Computer Engineering, Nanyang Technological
University, Singapore, from 2014 to 2015. He is
currently a Professor with the School
of Software Engineering, South China University
of Technology, Guangzhou, China. His current
research interests include machine learning,
data mining, big data research.
\end{IEEEbiography}
\vspace{-12ex}
\begin{IEEEbiography}[{\includegraphics[width=1in,height=1.25in,clip,keepaspectratio]{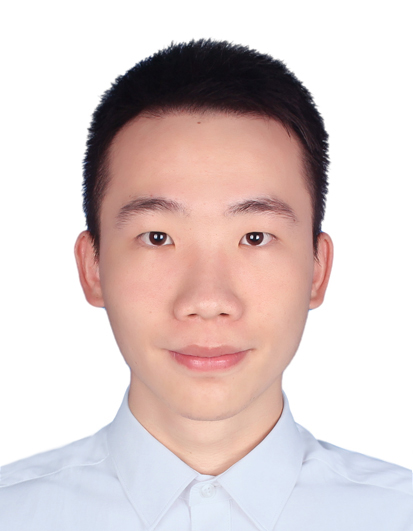}}]{Jiezhang Cao}
is a master of the School of
Software Engineering, South China University of
Technology, China. He received the B.S. degree
in statistics from the Guangdong University of
Technology, China, 2017. His current research
interests include machine learning, generative
model.
\end{IEEEbiography}
\vspace{-12ex}
\begin{IEEEbiography}[{\includegraphics[width=1in,height=1.25in,clip,keepaspectratio]{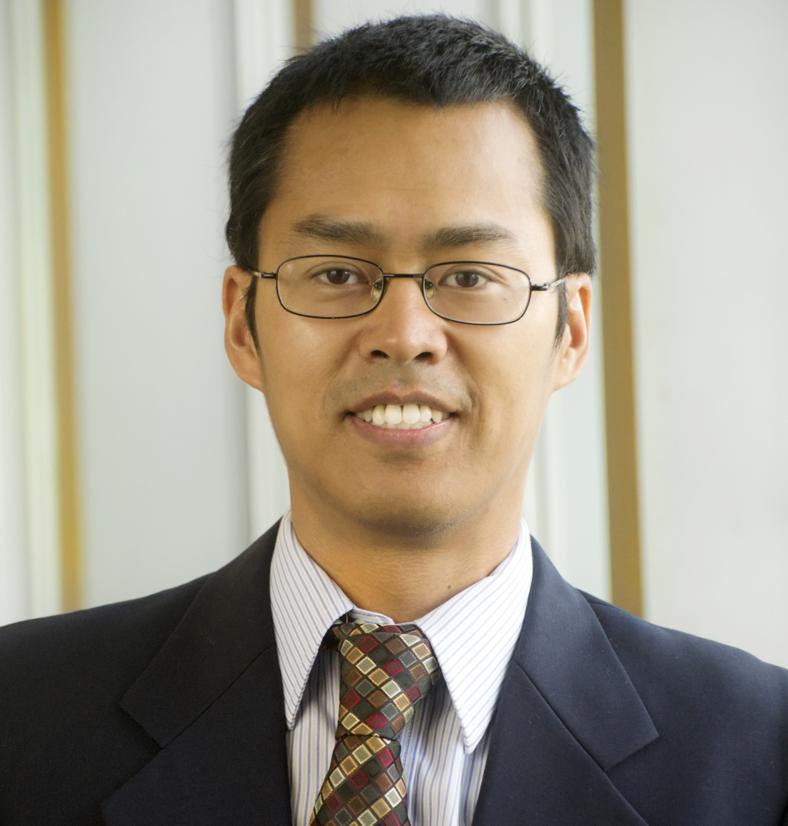}}]{Junzhou Huang}
is an Associate Professor in the Computer Science and Engineering department at the University of Texas at Arlington. He received the B.E. degree from Huazhong University of Science and Technology, China, the M.S. degree from Chinese Academy of Sciences, China, and the Ph.D. degree in Rutgers university. His major research interests include machine learning, computer vision and imaging informatics. He was selected as one of the 10 emerging leaders in multimedia and signal processing by the IBM T.J. Watson Research Center in 2010. He received the NSF CAREER Award in 2016.
\end{IEEEbiography}
\vspace{-12ex}
\begin{IEEEbiography}[{\includegraphics[width=1in,height=1.25in,clip,keepaspectratio]{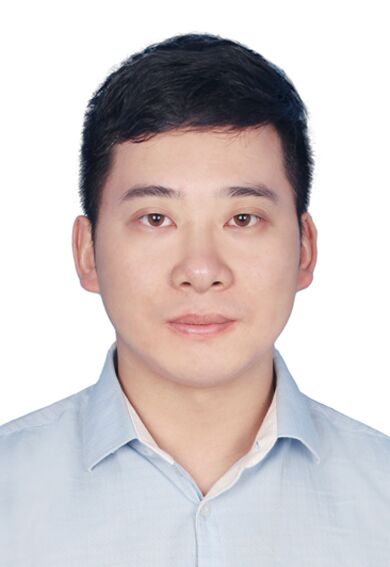}}]{Mingkui Tan}
received his Bachelor Degree in Environmental Science and Engineering in 2006 and Master degree in Control Science and Engineering in 2009, both from Hunan University in Changsha, China. He received the Ph.D. degree in Computer Science from Nanyang Technological University, Singapore, in 2014.  From 2014-2016, he  worked as a Senior Research Associate on computer vision in the School of Computer Science, University of Adelaide, Australia.  Since 2016, he has been with the School of Software Engineering, South China University of Technology, China, where he is currently a Professor.   His research interests include  machine learning,  sparse analysis, deep learning and large-scale optimization.
\end{IEEEbiography}


%
%




\end{document}


\title{Online Adaptive Asymmetric Active Learning with Limited Budgets}

%
\author{Yifan Zhang, Peilin Zhao, Shuaicheng Niu, Qingyao Wu, Jiezhang Cao, Junzhou Huang, Mingkui Tan
\IEEEcompsocitemizethanks{
\IEEEcompsocthanksitem Y. Zhang, S. Niu, J. Cao, Q. Wu and M. Tan are with South China University of Technology, China. E-mail: $\{$sezyifan, sensc, secaojiezhang$\}$@mail.scut.edu.cn; $\{$qyw, mingkuitan$\}$@scut.edu.cn.
\IEEEcompsocthanksitem P. Zhao and J. Huang are with Tencent AI Lab, China. Email: peilinzhao@hotmail.com; joehhuang@tencent.com.
\IEEEcompsocthanksitem P. Zhao, S. Niu and Q. Wu are co-first authors;  Corresponding to M. Tan.}
}

\markboth{IEEE TRANSACTIONS ON KNOWLEDGE AND DATA ENGINEERING}%
{Zhang \MakeLowercase{\textit{et al.}}}

\IEEEtitleabstractindextext{
\begin{abstract}
This supplemental file provides the proofs of theorems, some algorithms, additional experiments and the related work in our paper of "Online Adaptive Asymmetric Active Learning with Limited Budgets"~\cite{zhang2019online}.
\end{abstract}

\begin{IEEEkeywords}
Active Learning; Online Learning; Class Imbalance; Budgeted Query; Sketching Learning.
\end{IEEEkeywords}}

\maketitle

\IEEEdisplaynontitleabstractindextext

\IEEEpeerreviewmaketitle

\appendices
\IEEEraisesectionheading{\section{Proofs of Theorems}\label{theorems}}
This section presents the proofs for all the theorems. For convenience, we introduce the following notations:
\begin{align}
   &M_t  = \mathbb{I}_{(\hat{y}_t\neq y_t)}, \ \rho = \frac{\alpha_p T_n}{\alpha_n T_p} \ {\rm or} \ \frac{c_p}{c_n}, \nonumber  \\
    \rho_t \small{=} \rho \mathbb{I}_{(y_t =+1)}\small{+}&\mathbb{I}_{(y_t =-1)} , \ \rho_{max}\small{=}\max\{1,\rho\},  \ \rho_{min}\small{=}\min\{1,\rho\}.  \nonumber
\end{align}

\subsection{Proof of Lemma 1}
\textbf{Lemma 1.} \ \emph{Let $(x_1,y_1),...,(x_T,y_T)$ be a sequence of input samples, where $x_t \in \mathbb{R}^{d}$ and $y_t \in \{-1,+1\}$ for all t. Let $T_B$ be the round that runs out of the budgets, i.e.,  $B_{T_{B}+1}=B$. For any $\mu \in \mathbb{R}^{d}$ and any $\delta>0$, OA3 algorithm satisfies:
\begin{align}
   \sum_{t=1}^{T_B} M_t  Z_t(\delta \small{+} q_t) \small{\leq} \frac{\delta}{\rho_{min}}\sum_{t=1}^{T_B} \ell_t(\mu)\small{+} & \frac{1}{\eta \rho_{min}}\emph{Tr}(\Sigma_{T_{B}+1}^{-1})\times    \nonumber \\
   & \big{[}M(\mu) +(1-\delta)^2 ||\mu||^2 \big{]}, \nonumber
\end{align}
where $M(\mu)= \max_t||\mu_t \small{-}\mu||^2$.}

\vspace{1ex}
\begin{proof}
    Consider that OA3 queries a label but makes a mistake at the round $t$, so that $Z_t=1$ and $M_t = 1$. Then, based on the adaptive asymmetric update strategy, we have:
    \begin{align}
       \mu_{t+1} ={\rm arg \min}_\mu f_t(\mu,\Sigma)= {\rm arg \min}_\mu h_t(\mu), \nonumber
    \end{align}
    where $h_t(\mu)= \frac{1}{2}||\mu_t-\mu||_{\Sigma^{-1}_{t+1}}^2 + \eta g_t^{\top} \mu$.

    Since $h_t$ is convex and continuous, one can easily obtain the following inequality:
    \begin{align}
      \partial & h_t(\mu_{t+1})^\top (\mu-\mu_{t+1}) \nonumber \\
      =& \big{[}(\mu_{t+1}-\mu_t)^\top \Sigma_{t+1}^{-1} + \eta g_t^\top\big{]} (\mu-\mu_{t+1})\geq 0, \forall \mu.  \nonumber
    \end{align}

    Rearranging the inequality will give:
    \begin{align}\label{equation a1}
      (\eta g_t)^\top (\mu_{t+1}-\mu) \leq & (\mu_{t+1}- \mu_t)^\top \Sigma_{T+1}^{-1}(\mu-\mu_{t+1})  \nonumber \\
      = & \frac{1}{2}\Big{[}||\mu_{t}-\mu||_{\Sigma^{-1}_{t+1}}^2 -||\mu_{t+1}-\mu||_{\Sigma^{-1}_{t+1}}^2 \nonumber \\
      &\ -||\mu_{t}-\mu_{t+1}||_{\Sigma^{-1}_{t+1}}^2\Big{]}.
    \end{align}

    Now, we provide a lower bound for $g_t^\top (\mu_{t+1}-\mu)$. Since $\ell_t(\mu)=\rho_t \max(0,1-y_tx_t^{\top}\mu_t)$ is a convex function, and based on $g_t^\top = M_t(-\rho_ty_t x_t^\top)$ and
    \begin{align}\label{equation a2}
       \partial  h_t(\mu_{t+1})=0 \Longleftrightarrow (\mu_{t+1}-\mu_t)^\top \Sigma_{t+1}^{-1} + \eta g_t^\top =0,
    \end{align}
    we have:
    \begin{align}\label{equation a3}
      &g_t^\top (\mu_{t+1}-\mu) =  g_t^\top (\mu_{t+1}-\mu+\mu_t-\mu_t)  \nonumber \\
      &\small{=}  M_t(-\rho_ty_tx_t^{\top}\mu_t) \small{+}  M_t(\rho_ty_tx_t^{\top}\mu) \small{-} \frac{1}{\eta}||\mu_{t} \small{-}\mu_{t+1}||_{\Sigma^{-1}_{t+1}}^2.
    \end{align}

    Combining the above Equation (\ref{equation a3}) with the facts:
    \begin{align}
      \rho_tM_t(-y_tx_t^{\top}\mu_t) = \rho_tM_t |y_tx_t^{\top}\mu_t| = \rho_tM_t |p_t|, \nonumber
    \end{align}
    and
    \begin{align}
      \delta \ell_t(\frac{\mu}{\delta}) \geq \delta \rho_t(1-y_tx_t^{\top}\frac{\mu}{\delta}) \ \Leftrightarrow \ y_tx_t^{\top} \mu \geq \delta - \frac{\delta}{\rho_t} \ell_t(\frac{\mu}{\delta}), \nonumber
    \end{align}
    we get the following bound for $g_t^{\top}(\mu_{t+1}-\mu_t)$, \emph{i.e.,}
    \begin{align}\label{equation a4}
      g_t^{\top}&(\mu_{t+1}\small{-}\mu_t) \nonumber \\
      &\geq \rho_tM_t |p_t|\small{+}\rho_tM_t \Big{[} \delta\small{-} \frac{\delta}{\rho_t}\ell_t(\frac{\mu}{\delta})\Big{]} \small{-}\frac{1}{\eta}||\mu_{t}\small{-}\mu_{t+1}||_{\Sigma^{-1}_{t+1}}^2  \nonumber \\
      & =\rho_tM_t(\delta\small{+}|p_t|) \small{-} M_t \delta\ell_t(\frac{\mu}{\delta})\small{-}\frac{1}{\eta}||\mu_{t}\small{-}\mu_{t+1}||_{\Sigma^{-1}_{t+1}}^2.
    \end{align}

    Combining Equations (\ref{equation a1}) and (\ref{equation a4}) will give the following important inequality:
    \begin{align}\label{equation a5}
     M_tZ_t(\delta\small{+}|p_t|) \leq & \frac{Z_t}{2\eta\rho_t} \Big{[}||\mu_{t}-\delta\mu||_{\Sigma^{-1}_{t+1}}^2 -||\mu_{t+1}-\delta\mu||_{\Sigma^{-1}_{t+1}}^2 \nonumber \\
      &\small{+}||\mu_{t}-\mu_{t+1}||_{\Sigma^{-1}_{t+1}}^2 \Big{]} \small{+}M_tZ_t \Big{[} \frac{\delta}{\rho_t}\ell_t(\mu)\Big{]} ,
    \end{align}
    where we replace $\delta\mu$ with $\mu$.

    Then, according to Equation (\ref{equation a2}), we have:
    \begin{align}
      ||\mu_{t}-\mu_{t+1}||_{\Sigma^{-1}_{t+1}}^2 &= \eta^2  g_t^\top \Sigma_{t+1}g_t \nonumber \\
      &=M_t \eta^2 \rho_t^2 x_t^\top \Sigma_{t+1}x_t \nonumber \\
      &=M_t \eta^2 \rho_t^2 \Big{(} x_t^\top \Sigma_{t}x_t - \frac{x_t^\top \Sigma_{t}x_tx_t^\top \Sigma_{t}x_t}{\gamma + x_t^\top \Sigma_{t}x_t}\Big{)} \nonumber \\
      &=M_t \eta^2 \rho_t^2 \frac{\gamma v_t}{\gamma+v_t} \nonumber \\
      &=M_t \frac{\eta^2 \rho_t^2 }{\frac{1}{\gamma}+\frac{1}{v_t}}, \nonumber
    \end{align}
    where we use the updating rule of $\Sigma$.

    Then, according to $M_t\leq 1$ and $Z_t\leq 1$, we rearrange Equation (\ref{equation a5}):
    \begin{align}\label{equation a6}
      M_tZ_t(\delta+q_t) &=M_tZ_t\big{( } \delta\small{+}|p_t| + c_t\big{)} \nonumber \\
      &=M_tZ_t\Big{(} \delta\small{+}|p_t|- \frac{1}{2}\frac{\eta \rho_{max}}{\frac{1}{\gamma}+\frac{1}{v_t}}\Big{)} \nonumber \\
      &\leq M_tZ_t\Big{(} \delta\small{+}|p_t|- \frac{1}{2}\frac{\eta\rho_t}{\frac{1}{\gamma}+\frac{1}{v_t}}\Big{)}  \nonumber \\
      &\leq \frac{Z_t}{2\eta\rho_t} \Big{[}||\mu_{t}-\delta\mu||_{\Sigma^{-1}_{t+1}}^2 -||\mu_{t+1}-\delta\mu||_{\Sigma^{-1}_{t+1}}^2 \Big{]} \nonumber \\
      & \ \ \ \ +M_tZ_t \Big{[} \frac{\delta}{\rho_t}\ell_t(\mu)\Big{]} \nonumber \\
      &\leq \frac{Z_t}{2\eta\rho_{t}}\Big{[} ||\mu_{t}-\delta\mu||_{\Sigma^{-1}_{t+1}}^2 -||\mu_{t+1}-\delta\mu||_{\Sigma^{-1}_{t+1}}^2 \Big{]}  \nonumber \\
      & \ \ \ \ +\frac{\delta}{\rho_{min}}\ell_t(\mu).
    \end{align}

    We highlight the analysis here provides the theoretical guarantees for the definition of query confidence $c_t$, which fascinates the theoretical studies of the proposed algorithm. Next, summing the first right term of above inequality over $t=1,...,T_B$, we have:
    \begin{align}\label{equation a7}
      & \sum_{t=1}^{T_B} \frac{Z_t}{\rho_{t}}\Big{[} ||\mu_{t}-\delta\mu||_{\Sigma^{-1}_{t+1}}^2 -||\mu_{t+1}-\delta\mu||_{\Sigma^{-1}_{t+1}}^2\Big{]}   \nonumber \\
       & \leq  \frac{1}{\rho_{min}}\Big{\{} ||\mu_{1}\small{-}\delta\mu||_{\Sigma^{-1}_{2}}^2 \small{+} \sum_{t=2}^{T_B}\big{[} ||\mu_{t}\small{-}\delta\mu||_{\Sigma^{-1}_{t+1}}^2 \small{-}||\mu_{t} \small{-} \delta\mu||_{\Sigma^{-1}_{t}}^2\big{]} \Big{\}}   \nonumber \\
       & = \frac{1}{\rho_{min}}\Big{[} ||\mu_{1}-\delta\mu||_{\Sigma^{-1}_{2}}^2 + \sum_{t=2}^{T_B}||\mu_{t}-\delta\mu||_{(\Sigma^{-1}_{t+1}-\Sigma^{-1}_{t})}^2\Big{]} \nonumber \\
       & \small{\leq} \frac{1}{\rho_{min}}\small{[||}\mu_{1}\small{-}\delta\mu\small{||}^2 \lambda_{max}\small{(}\Sigma_2^{\small{-1}}\small{)} \small{+} \small{\sum_{t=2}^{T_B}}\small{||}\mu_{t}\small{-}\delta\mu\small{||}^2 \lambda_{max}\small{(}\Sigma^{\small{-1}}_{t\small{+}1}\small{-}\Sigma^{\small{-1}}_{t}\small{)]} \nonumber \\
       & \small{\leq} \frac{1}{\rho_{min}} \Big{[}||\mu_{1}\small{-}\delta\mu||^2 {\rm Tr}(\Sigma_2^{-1}) \small{+} \sum_{t=2}^{T_B} ||\mu_{t} \small{-} \delta\mu||^2  {\rm Tr}(\Sigma^{-1}_{t+1}\small{-}\Sigma^{-1}_{t})\Big{]} \nonumber \\
       & \leq  \frac{1}{\rho_{min}}\max_{t\leq T_B}||\mu_{t}\small{-}\delta\mu||^2{\rm Tr}(\Sigma_{T_{B}+1}^{-1})\nonumber \\
       & \leq \frac{2}{\rho_{min}}\big{[} M(\mu)+ (1-\delta)^2||\mu||^2\big{]}  {\rm Tr}(\Sigma_{T_{B}+1}^{-1}),
    \end{align}
    where $M(\mu) = {\rm max}_t||\mu_t-\mu||^2$ and $\lambda_{max}(\Sigma)$ is the largest eigenvalue of $\Sigma$.

    Now, combining Inequalities (\ref{equation a6}) and (\ref{equation a7}), we can easily obtain:
    \begin{align}
       \sum_{t=1}^{T_B} M_t  Z_t(\delta + q_t) \small{\leq} \frac{\delta}{\rho_{min}}\sum_{t=1}^{T_B} \ell_t(\mu)\small{+} & \frac{1}{\eta \rho_{min}}{\rm Tr}(\Sigma_{T_{B}+1}^{-1})\times    \nonumber \\
       & \big{[}M(\mu) +(1-\delta)^2 ||\mu||^2 \big{]}. \nonumber
    \end{align}

    Consider another situation that $M_tZ_t=0$, and we can find above inequality still holds. As results, we conclude the proofs of Lemma 1.
\end{proof}

\subsection{Proof of Theorem 1}
\textbf{Theorem 1.} \ \emph{Let $(x_1,y_1),...,(x_T,y_T)$ be a sequence of input samples, where $x_t \in \mathbb{R}^{d}$ and $y_t \in \{-1,+1\}$ for all t. Let $T_B$ be the round that runs out of the budgets, i.e.,  $B_{T_{B}+1}=B$. For any $\mu \in \mathbb{R}^d$, the expected mistake number of OA3 within budgets is bounded by:
\begin{align}
    \mathbb{E}\bigg{[}\sum_{t=1}^{T_B} M_t\bigg{]} &= \mathbb{E}\bigg{[}\sum_{\substack{t=1\\ y_t = +1}}^{T_B}M_t+\sum_{\substack{t=1\\ y_t = -1}}^{T_B}M_t\bigg{]}    \nonumber \\
    & \leq \frac{1}{\rho_{min}}\bigg{[}\sum_{t=1}^{T_B}\ell_t(\mu) + \frac{1}{\eta} D(\mu)  \emph{Tr}(\Sigma_{T_{B}+1}^{-1})\bigg{]},  \nonumber
\end{align}
where $D(\mu)\small{=} \max \Big{\{}\frac{M(\mu)\small{+} (1\small{-}\delta_{+})^2 ||\mu||^2}{\delta_{+}}, \frac{M(\mu)\small{+} (1\small{-}\delta_{-})^2 ||\mu||^2}{ \delta_{-}} \Big{\}}$.}

\begin{proof}
    Considering that OA3 queries a label but makes a mistake at the round $t$, so that $Z_t=1$ and $M_t = 1$, there are two scenarios. That is, $p_t \small{\geq} 0$ with $M_tZ_t\small{=}1$ represents our estimated class of sample $x_t$ is positive, but true label is negative; while $p_t \small{<}0$ with $M_tZ_t\small{=}1$ represents our estimated class of sample $x_t$ is negative, but true label is positive.

    First, if $p_t \geq 0$, based on Lemma 1, for any $\mu \in \mathbb{R}^d$ and any $\delta_{+}>0$, we have :
    \begin{align}
       \sum_{\substack{t=1\\ y_t = -1}}^{T_B} M_t  Z_t(\delta_{+} + q_t) \small{\leq} &\frac{\delta_{+}}{\rho_{min}}\sum_{\substack{t=1\\ y_t = -1}}^{T_B}\ell_t(\mu) + \frac{\mathbb{I}_{(y_t =-1)} }{\eta\rho_{min}}\times  \nonumber \\
       & \big{[}M(\mu) +(1-\delta_{+})^2 ||\mu||^2 \big{]} {\rm Tr}(\Sigma_{T_{B}+1}^{-1}) . \nonumber
    \end{align}

    One can easily prove that this inequality still holds for $M_tZ_t=0$.

    Now, we would like to remove the random variable $Z_t$. First, when the query parameter $q_t \small{>} 0$, taking the expectation over random variables $\mathbb{E}(Z_t) \small{=} \frac{\delta_{+}}{\delta_{+}+q_t}$ for $p_t\small{\geq} 0$, we have:
    \begin{align}
       \mathbb{E}\bigg{[}\sum_{\substack{t=1\\ y_t = -1}}^{T_B}\delta_{+} M_t\bigg{]} \small{\leq} \frac{\delta_{+}}{\rho_{min}}&\sum_{\substack{t=1\\ y_t = -1}}^{T_B}\ell_t(\mu) + \frac{\mathbb{I}_{(y_t =-1)} }{\eta\rho_{min}}\times     \nonumber \\
       &  \big{[}M(\mu) +(1-\delta_{+})^2 ||\mu||^2 \big{]} {\rm Tr}(\Sigma_{T_{B}+1}^{-1}). \nonumber
    \end{align}

    On the other hand, when the query parameter $q_t \leq 0$, we set $q_t = 0 $, and the random variables satisfy $\mathbb{E}(Z_t) = 1$. Then, we find the above inequality still holds. In addition, one can easily prove this inequality holds for $M_t=0$.

    Now, we obtain:
    \begin{align}\label{equation a8}
       \mathbb{E}\bigg{[}\sum_{\substack{t=1\\ y_t = -1}}^{T_B}M_t\bigg{]} \leq &\sum_{\substack{t=1\\ y_t = -1}}^{T_B}\frac{\ell_t(\mu)}{\rho_{min}} + \frac{\mathbb{I}_{(y_t =-1)}}{\eta \rho_{min} \delta_{+}}\times\nonumber \\
       & \big{[}M(\mu) +(1-\delta_{+})^2 ||\mu||^2 \big{]} {\rm Tr}(\Sigma_{T_{B}+1}^{-1}) .
    \end{align}

    Similarly, when $p_t <0$, for any $\mu \in \mathbb{R}^d$ and any $\delta_{-}>0$, we obtain:
    \begin{align}\label{equation a9}
       \mathbb{E}\bigg{[}\sum_{\substack{t=1\\ y_t = +1}}^{T_B}M_t\bigg{]} \leq & \sum_{\substack{t=1\\ y_t = +1}}^{T_B}\frac{\ell_t(\mu)}{\rho_{min}} +\frac{\mathbb{I}_{(y_t =+1)} }{\eta\rho_{min}\delta_{-}}\times   \nonumber \\
       &  \big{[}M(\mu) +(1-\delta_{-})^2 ||\mu||^2 \big{]} {\rm Tr}(\Sigma_{T_{B}+1}^{-1}) .
    \end{align}

    Summing Equations (\ref{equation a8}) and (\ref{equation a9}) will give:
    \begin{align}
        \mathbb{E}\bigg{[}\sum_{t=1}^{T_B} M_t\bigg{]} &= \mathbb{E}\bigg{[}\sum_{\substack{t=1\\ y_t = +1}}^{T_B}M_t+\sum_{\substack{t=1\\ y_t = -1}}^{T_B}M_t\bigg{]}    \nonumber \\
        & \leq \sum_{t=1}^{T_B}\frac{\ell_t(\mu)}{\rho_{min}} + \frac{1}{\eta \rho_{min}} D(\mu) {\rm Tr}(\Sigma_{T_{B}+1}^{-1})  \nonumber \\
        &= \frac{1}{\rho_{min}}\Big{[} \sum_{t=1}^{T_B}\ell_t(\mu) + \frac{1}{\eta} D(\mu)  {\rm Tr}(\Sigma_{T_{B}+1}^{-1})\Big{]} , \nonumber
    \end{align}
    where $D(\mu)\small{=} \max \big{\{}\frac{M(\mu)\small{+} (1\small{-}\delta_{+})^2 ||\mu||^2}{\delta_{+}}, \frac{M(\mu)\small{+} (1\small{-}\delta_{-})^2 ||\mu||^2}{\delta_{-}} \big{\}}$.

    Then, we conclude the proofs of Theorem 1.
\end{proof}

\subsection{Proof of Theorem 2}
\textbf{Theorem 2.} \ \emph{Under the same condition in Theorem 1, by setting $\rho = \frac{\alpha_pT_n}{\alpha_nT_p}$, the proposed OA3 within budgets satisfies for any $\mu \in \mathbb{R}^d$:
\begin{align}
   \mathbb{E}\Big{[}sum\Big{]}  \geq   1  -  \frac{\alpha_n \rho_{max}}{T_n \rho_{min}} \bigg{[}\sum_{t=1}^{T_B}\ell_t(\mu) \small{+}\frac{1}{\eta} D(\mu) \emph{Tr}(\Sigma_{T_{B}+1}^{-1})\bigg{]}.\nonumber
\end{align}}

\begin{proof}
    According to Equation (\ref{equation a9}), we have:
    \begin{align}
       \mathbb{E}\bigg{[}\sum_{\substack{t=1\\ y_t = +1}}^{T_B} \rho M_t\bigg{]} \leq & \sum_{\substack{t=1\\ y_t = +1}}^{T_B} \frac{\rho \ell_t(\mu)}{\rho_{min}} +\frac{\rho \mathbb{I}_{(y_t =+1)} }{\eta \rho_{min} \delta_{-}}\times   \nonumber \\
       &  \big{[}M(\mu) +(1-\delta_{-})^2 ||\mu||^2 \big{]} {\rm Tr}(\Sigma_{T_{B}+1}^{-1}).   \nonumber
    \end{align}

    Now, combining with Equation (\ref{equation a8}) will give:
    \begin{align}\label{equation a10}
        \mathbb{E}\bigg{[}\sum_{t=1}^{T_B} \rho_t M_t\bigg{]} &\small{=} \mathbb{E}\bigg{[}\sum_{\substack{t=1\\ y_t = +1}}^{T_B}\rho M_t+\sum_{\substack{t=1\\ y_t = -1}}^{T_B}M_t\bigg{]}    \nonumber \\
        &\small{\leq} \frac{\max\{1,\rho\}}{\rho_{min}} \Big{[}\sum_{t=1}^{T_B}\ell_t(\mu) + \frac{1}{\eta} D(\mu)  {\rm Tr}(\Sigma_{T_{B}+1}^{-1})\Big{]}  \nonumber \\
        &\small{=} \frac{\rho_{max}}{\rho_{min}} \Big{[}\sum_{t=1}^{T_B}\ell_t(\mu) \small{+} \frac{1}{\eta} D(\mu) {\rm Tr}(\Sigma_{T_{B}+1}^{-1})\Big{]}.
    \end{align}

    Now, from the definition of the weighted $sum$, we have:
    \begin{align}\label{equation a11}
       & \mathbb{E}\bigg{[}\sum_{t=1}^{T_B} \rho_t M_t\bigg{]} \small{=} \rho\mathbb{E} \big{[}M_p\big{]}+ \mathbb{E}\big{[}M_n\big{]} \small{=} \Big{(}\frac{\alpha_p T_n}{\alpha_n T_p}\Big{)}\mathbb{E}\big{[}M_p\big{]}\small{+}\mathbb{E}\big{[}M_n\big{]}   \nonumber \\
       & \small{=}\frac{T_n}{\alpha_n}\bigg{[}\alpha_p \frac{\mathbb{E}\big{[}M_p\big{]}}{T_p} \small{+}\alpha_n  \frac{\mathbb{E}\big{[}M_n\big{]}}{T_n} \bigg{]} \small{=} \frac{T_n}{\alpha_n}\bigg{(}1 \small{-} \mathbb{E}\big{[} sum\big{]}\bigg{)},
    \end{align}
    where we used $\alpha_p + \alpha_n =1$.

    Combining Equations (\ref{equation a10}) and (\ref{equation a11}), we have:
    \begin{align}
       \mathbb{E}\Big{[}  sum\Big{]} \geq  1 - \frac{\alpha_n \rho_{max}}{T_n \rho_{min}}  \bigg{[}\sum_{t=1}^{T_B}\ell_t(\mu) \small{+}\frac{1}{\eta} D(\mu) {\rm Tr}(\Sigma_{T_{B}+1}^{-1})\bigg{]}.\nonumber
    \end{align}

    Then, we conclude Theorem 2.
\end{proof}

\subsection{Proof of Theorem 3}
\textbf{Theorem 3.} \ \emph{Under the same condition in Theorem 1, by setting $\rho\small{=}\frac{c_p}{c_n}$, the proposed OA3 within budgets satisfies for any $\mu \in \mathbb{R}^d$:
\begin{align}
   \mathbb{E}\Big{[}cost\Big{]} \leq  \frac{c_n \rho_{max}}{\rho_{min}} \bigg{[}\sum_{t=1}^{T_B}\ell_t(\mu) \small{+}\frac{1}{\eta} D(\mu) \emph{Tr}(\Sigma_{T_{B}+1}^{-1})\bigg{]}. \nonumber
\end{align}}

\begin{proof}
    From the definition of $cost$ metric, we have:
    \begin{align}\label{equation a12}
       & \mathbb{E}\bigg{[}\sum_{t=1}^{T_B} \rho_t M_t\bigg{]} \small{=} \rho\mathbb{E} \big{[}M_p\big{]}\small{+} \mathbb{E}\big{[}M_n\big{]} \small{=} \frac{c_p}{c_n}\mathbb{E}\big{[}M_p\big{]}+\mathbb{E}\big{[}M_n\big{]}   \nonumber \\
       & =\frac{1}{c_n}\Big{(}c_p\mathbb{E}\big{[}M_p\big{]}\small{+} c_n\mathbb{E}\big{[}M_n\big{]}\Big{)} \small{=} \frac{1}{c_n}\mathbb{E}\Big{[}cost\Big{]}.
    \end{align}

    Combining both Equations (\ref{equation a10}) and (\ref{equation a12}) concludes Theorem 3.
\end{proof}

\subsection{Proof of Theorem 4}
\textbf{Theorem 4.} \ \emph{Let $(x_1,y_1),...,(x_T,y_T)$ be a sample stream, where $x_t \in \mathbb{R}^{d}$ and $y_t \in \{-1,+1\}$. Let $T_B$ be the round that uses up the budgets, i.e.,  $B_{T_{B}+1}\small{=}B$. For any $\mu \small{\in} \mathbb{R}^d$, the expected mistakes of OA3 over budgets is bounded by:
\begin{align}
   \mathbb{E}\bigg{[}\sum_{T_{B}+1}^{T} M_t\bigg{]} \leq \sum_{T_{B}+1}^{T}\bigg{[}\frac{\ell_t(\mu)}{\rho_{min}} + y_tx_t^{\top}\mu_{T_{B}+1}\bigg{]}, \nonumber
\end{align}
where $\mu_{T_{B}+1}$ is the predictive vector of model, trained by all the previous queried samples.}

\begin{proof}
    When running out of budget, the sample sequence is from $(x_{T_{B}+1},y_{T_{B}+1}),...,(x_T,y_T)$. Now, for any $t$ after $T_B$, the predictive vector $\mu_{t+1}=\mu_t=\mu_{T_{B}+1}$. Combining this with the fact:
    \begin{align}
      \ell_t(\mu) \geq \rho_t(1-y_tx_t^{\top}\mu) \ \Leftrightarrow \ y_tx_t^{\top} \mu \geq 1 - \frac{1}{\rho_t} \ell_t(\mu), \nonumber
    \end{align}
    we have:
    \begin{align}
       M_t &\leq M_t\Big{[}y_t x_t^\top \mu_{T_{B}+1} + \frac{\ell_t(\mu)}{\rho_t}\Big{]} \nonumber \\
       &\leq y_t x_t^\top \mu_{T_{B}+1} + \frac{\ell_t(\mu)}{\rho_{min}}, \nonumber
    \end{align}
    where we use  $M_t \leq 1$. Then, we obtain:
    \begin{align}
       \mathbb{E}\bigg{[}\sum_{T_{B}+1}^{T} M_t\bigg{]} \leq \sum_{T_{B}+1}^{T}\bigg{[}\frac{\ell_t(\mu)}{\rho_{min}} + y_tx_t^{\top}\mu_{T_{B}+1}\bigg{]}, \nonumber
    \end{align}
    which concludes Theorem 4.
\end{proof}

\subsection{Proof of Theorem 5}
\textbf{Theorem 5.} \ \emph{Under the same condition in Theorem 4, by setting $\rho = \frac{\alpha_pT_n}{\alpha_nT_p}$, the sum performance of OA3 over budgets satisfies for any $\mu \in \mathbb{R}^d$:
\begin{align}
   \mathbb{E}\Big{[} sum\Big{]}\geq 1  - \frac{\alpha_n \rho_{max}}{T_n}\sum_{T_{B}+1}^{T}\bigg{[}\frac{\ell_t(\mu)}{\rho_{min}} + y_tx_t^{\top}\mu_{T_{B}+1}\bigg{]}. \nonumber
\end{align}}

\begin{proof}
    From Theorem 4, we have:
    \begin{align}\label{equation a13}
       \mathbb{E}\bigg{[}\sum_{\substack{T_{B}+1\\ y_t = +1}}^{T} \rho M_t\bigg{]} \leq \sum_{\substack{T_{B}+1\\ y_t = +1}}^{T}\rho\Big{[}\ell_t(\mu) + y_tx_t^{\top}\mu_{T_{B}+1}\Big{]},
    \end{align}
    \begin{align}\label{equation a14}
       \mathbb{E}\bigg{[}\sum_{\substack{T_{B}+1\\ y_t = -1}}^{T}  M_t\bigg{]} \leq \sum_{\substack{T_{B}+1\\ y_t = -1}}^{T}\Big{[}\ell_t(\mu) + y_tx_t^{\top}\mu_{T_{B}+1}\Big{]}.
    \end{align}

    According to Equations (\ref{equation a11}), (\ref{equation a13}) and (\ref{equation a14}), we have:
    \begin{align}
       \mathbb{E}\Big{[} sum\Big{]} &\geq 1 \small{-} \frac{\alpha_n}{T_n}\small{\times}\max\{1,\rho\}\sum_{T_{B}+1}^{T}\bigg{[}\frac{\ell_t(\mu)}{\rho_{min}} \small{+} y_tx_t^{\top}\mu_{T_{B}+1}\bigg{]}\nonumber \\
       &\geq 1 - \frac{\alpha_n \rho_{max}}{T_n}\sum_{T_{B}+1}^{T}\bigg{[}\frac{\ell_t(\mu)}{\rho_{min}} + y_tx_t^{\top}\mu_{T_{B}+1}\bigg{]}, \nonumber
    \end{align}
    which concludes Theorem 5.
\end{proof}

\subsection{Proof of Theorem 6}
\textbf{Theorem 6.} \ \emph{Under the same condition in Theorem 4, by setting $\rho\small{=}\frac{c_p}{c_n}$, the misclassification cost of OA3 over budgets satisfies for any $\mu \in \mathbb{R}^d$:
\begin{align}
   \mathbb{E}\Big{[}cost\Big{]} \leq c_n \rho_{max}\sum_{T_{B}+1}^{T}\bigg{[}\frac{\ell_t(\mu)}{\rho_{min}} + y_tx_t^{\top}\mu_{T_{B}+1}\bigg{]}. \nonumber
\end{align}}

\begin{proof}
    Based on Equations (\ref{equation a12}), (\ref{equation a13}) and (\ref{equation a14}), we have:
    \begin{align}
       \mathbb{E}\Big{[}cost\Big{]} &\leq c_n \max\{1,\rho\}\sum_{T_{B}+1}^{T}\bigg{[}\frac{\ell_t(\mu)}{\rho_{min}} + y_tx_t^{\top}\mu_{T_{B}+1}\bigg{]} \nonumber \\
       &\leq c_n \rho_{max}\sum_{T_{B}+1}^{T}\bigg{[}\frac{\ell_t(\mu)}{\rho_{min}} + y_tx_t^{\top}\mu_{T_{B}+1}\bigg{]}, \nonumber
    \end{align}
    which concludes Theorem 6.
\end{proof}

\section{Algorithms}

\subsection{Diagonal Version of OA3}

In this subsection, we provide details for diagonal version of OA3 (named, OA3$_{diag}$).

The difference between OA3 and OA3$_{diag}$ is the update strategy. \yifan{OA3$_{diag}$ only exploits the diagonal elements of covariance when updating both the covariance matrix $\Sigma$ and predictive vector $\mu$. In other words, OA3$_{diag}$ only considers the model confidence of each feature independently and ignores the correlations among different features. As a result, it can achieve faster computational efficiency than OA3.}

\yifan{Note that the initialization of $\Sigma^{diag}_1$ is a identity matrix. All non-diagonal elements equal to 0. At round $t$, we only update diagonal elements of covariance matrix. To this end, we adjust the update rule
\begin{align}\label{eq:sigma_t+1}
  \Sigma_{t+1} = \Sigma_{t} - \frac{\Sigma_{t}x_tx_t^\top\Sigma_{t}}{\gamma+x_t^\top\Sigma_{t}x_t}
\end{align}
to
\begin{align}\label{eq:sigma_t+1_diag}
  \Sigma_{t+1}^{diag} = \Sigma_{t}^{diag} - \frac{\Sigma_{t}^{diag}\circ x_t \circ \Sigma_{t}^{diag}\circ x_t}{\gamma+x_t  \circ \Sigma_{t}^{diag}\circ {x_t}},
\end{align}
where $\circ$ is the element-wise product so that $\Sigma_{t+1}^{diag} \in \mathbb{R}^{d}$ can keep the diagonal property. Then, the diagonal update of $\mu_t$ is as follows:
\begin{align}\label{eq:mu_t+1_diag}
  \mu_{t+1} = \mu_t- \eta \Sigma_{t+1}\circ g_t.
\end{align}}

We summarize the diagonal OA3 update strategy in Algorithm \ref{al:oa3_update_diag}.

\begin{algorithm}
    \caption{Diagonal Adaptive Asymmetric Update Strategy: \textbf{\red{\textbf{DiagUpdate}}$(\mu_{t},\Sigma_{t};x_t,y_t)$}.} \label{al:oa3_update_diag}
    \begin{algorithmic}[1]
        \Require $\rho=\frac{\alpha_pT_n}{\alpha_nT_p}$ for ``$sum$" or $\rho=\frac{c_p}{c_n}$ for ``$cost$";
            \State Receive a sample $(x_t,y_t);$
            \State Compute the loss $\ell_t(\mu_t)$, based on Equation (4)$;$
            \If {$\ell_t(\mu_t) > 0$}
                \State $\Sigma_{t+1}^{diag} = \Sigma_{t}^{diag} - \frac{\Sigma_{t}^{diag}\circ x_t \circ \Sigma_{t}^{diag}\circ x_t}{\gamma+x_t^\top \circ \Sigma_{t}^{diag}{x_t}};$
                \State $\mu_{t+1} = \mu_t- \eta \Sigma_{t+1}\circ g_t,$   where $ g_t=\partial\ell_t(\mu_t).$
            \Else
                \State $\mu_{t+1} = \mu_{t}, \Sigma_{t+1}^{diag}= \Sigma_{t}^{diag}.$
            \EndIf
         \State Return \hspace{0.5ex} $\mu_{t+1}, \Sigma_{t+1}^{diag}.$
    \end{algorithmic}
\end{algorithm}

\subsection{Sparse Sketch Algorithms}
This section provides algorithms (Algorithm \ref{al:ssoa3_query}, \ref{al:ssoa3_oja} and \ref{al:ssoa3_decompose}) for Section 4.2.

\begin{algorithm}
    \caption{Sparse Sketched Asymmetric Query Strategy: \textbf{\red{\textbf{SparseSketchQuery}}$(p_t)$}.}\label{al:ssoa3_query}
    \begin{algorithmic}[1]
    \Require $\rho_{max}=\max\{1,\rho\}$; query bias $(\delta_{+}\small{,} \delta_{-})$ for positive and negative predictions.
        \State Variance $v_t \small{=}x_t^{\top}(I_d \small{-} \yifan{U}_{t-1}^{\top}F_{t-1}^{\top}(t\small{-}1)\Lambda_{t-1}H_{t-1}F_{t-1}\yifan{U}_{t-1})x_t;$
        \State Compute the query parameter $q_t=|p_t|- \frac{1}{2} \frac{\eta\rho_{max}}{\frac{1}{v_t}+\frac{1}{\gamma}};$
        \If {$ q_t \leq 0$}
            \State Set $q_t=0;$
        \EndIf

        \If {$ p_t \geq 0$}
            \State $p_t^{+} = \frac{\delta_{+}}{\delta_{+}+q_t};$
            \State Draw a Bernoulli variable $Z_t \small{\in} \{0,1\}$ with $p_t^{+}.$
        \Else
            \State $p_t^{-} = \frac{\delta_{-}}{\delta_{-}+q_t};$
            \State Draw a Bernoulli variable $Z_t \small{\in} \{0,1\}$ with $p_t^{-}.$
        \EndIf

        \State Return \hspace{0.5ex} $Z_t.$
    \end{algorithmic}
\end{algorithm}

\begin{algorithm}
    \caption{Sparse Oja's Sketch for OA3}\label{al:ssoa3_oja}
    \begin{algorithmic}
        \Require $m$, $\hat{x}$ and stepsize matrix $\Gamma_t$.
        \State \hspace{-2.7ex} \textbf{Internal State} \ $t$, $\Lambda$, $F$, $\yifan{U}$, $K$ and $H.$
        \State \hspace{-2.8ex} \textbf{SparseSketchInit}$(m)$
        \State \hspace{1ex}1: Set $t=0, F = K= H = I_m , \Lambda = 0_{m\times m}$
        \State \hspace{3ex} and $\yifan{U}$ to any $ m \times d$ matrix with orthonormal rows$;$
        \State \hspace{1ex}2: Return $(\Lambda,F,\yifan{U},H).$ \\

        \State \hspace{-2.7ex} \textbf{SparseSketchUpdate}$(\hat{x})$
        \State \hspace{1ex}1: Update $t \leftarrow t+1;$
        \State \hspace{1ex}2: $\Lambda=(I_m - \Gamma_{t})\Lambda+ \Gamma_{t} diag\{F \yifan{U}\hat{x}\}^2;$
        \State \hspace{1ex}3: Set $\delta = F^{-1}\Gamma_tF\yifan{U}        \hat{x}^{\top};$
        \State \hspace{1ex}4: $K \leftarrow K +\yifan{U} \hat{x} \delta^{\top}+ \delta\hat{x}^{\top}\yifan{U}^{\top}+ \delta \hat{x}^{\top} \hat{x} \delta^{\top};$
        \State \hspace{1ex}5: $\yifan{U} \leftarrow \yifan{U} + \delta \hat{x}^{\top};$
        \State \hspace{1ex}6: $(L,Q) \leftarrow$ Decompose$(F, K)$,
        \State \hspace{3ex} where $LQ\yifan{U}=F\yifan{U}$ and $Q\yifan{U}$ is orthogonal$;$
        \State \hspace{1ex}7: Set $F=Q;$
        \State \hspace{1ex}8: Set $H = diag\{\frac{1}{1+t\Lambda_{1,1}},...,\frac{1}{1+t\Lambda_{m,m}}\};$
        \State \hspace{1ex}9: Return $(\Lambda, F, \yifan{U}, H,\delta).$
    \end{algorithmic}
\end{algorithm}

\begin{algorithm}
    \caption{Decompose$(F, K)$}\label{al:ssoa3_decompose}
    \begin{algorithmic}[1]
        \Require $F \in \mathbb{R}^{m\times m}$ and Gram matrix $K=\yifan{U}\yifan{U}^{\top} \in \mathbb{R}^{m\times m};$
        \Ensure $L= 0_{m \times m}$ and $Q = 0_{m \times m};$
        \For{$i = 1 \to m$}
            \State Let $f^{\top}$ be the $i$-th row of $F;$
            \State Compute $\alpha = QKf$, $\beta = f - Q^{\top}\alpha$ and $c=\sqrt{\beta^{\top}K\beta};$
            \If {$c \neq 0$}
                \State Insert $\frac{1}{c}\beta^{\top}$ to the $i$-th row of $Q;$
            \EndIf
            \State Set the $i$-th entry of $\alpha$ to be $c;$
            \State Insert $\alpha$ to the $i$-th row of $L;$
            \EndFor
            \State Delete the all-zero columns of $L$ and all-zero rows of $Q;$
            \State Return $(L,Q).$

    \end{algorithmic}
\end{algorithm}

\section{Additional Experiments}
This section provides the additional experimental results.

\subsection{Cost Evaluation of Query Biases}

\begin{figure}
    \begin{minipage}{0.47\linewidth}
      \centerline{\includegraphics[width=4.7cm]{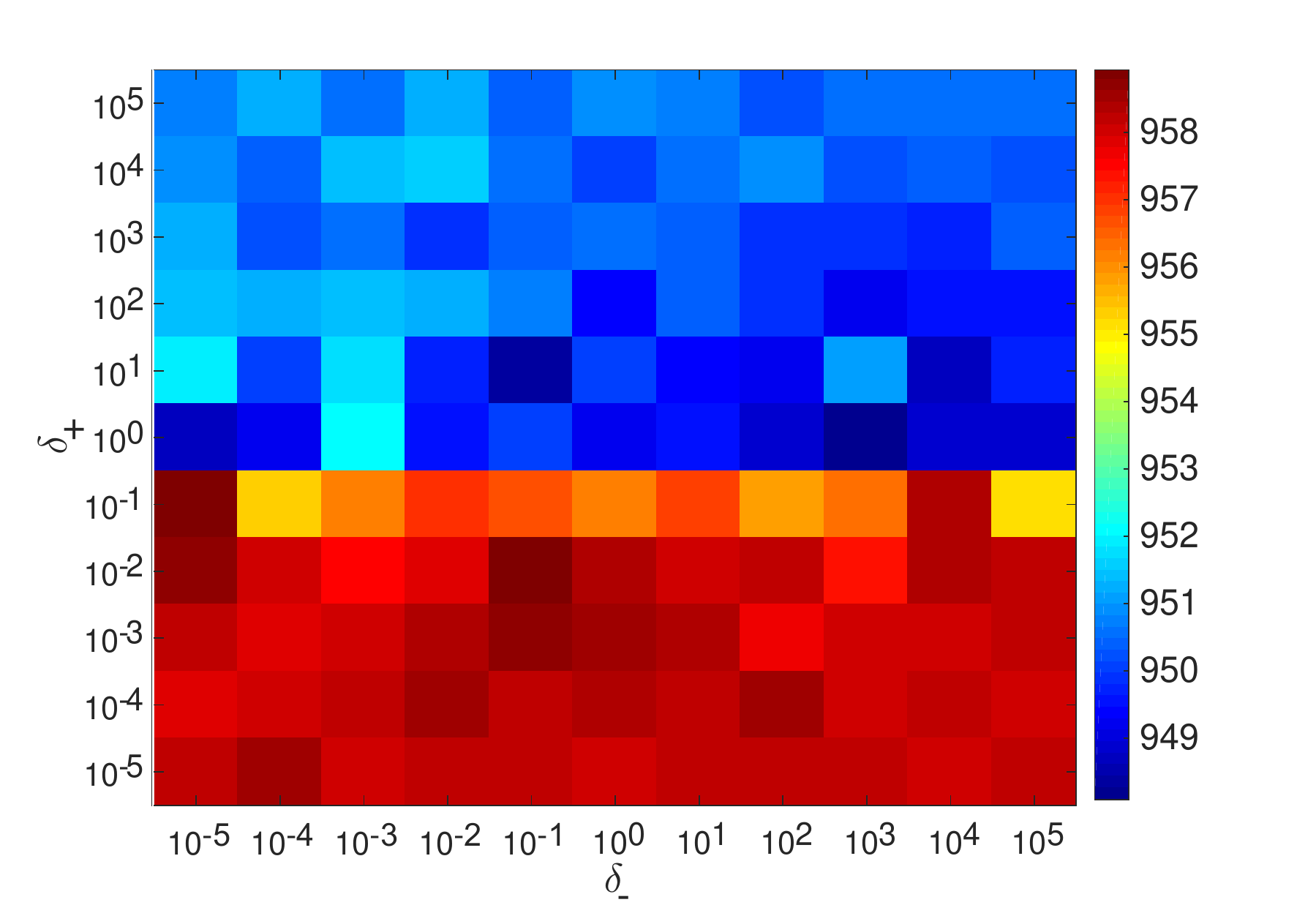}}
      \centerline{(a) protein, B=8500}
    \end{minipage}
    \hfill
    \begin{minipage}{0.47\linewidth}
      \centerline{\includegraphics[width=4.7cm]{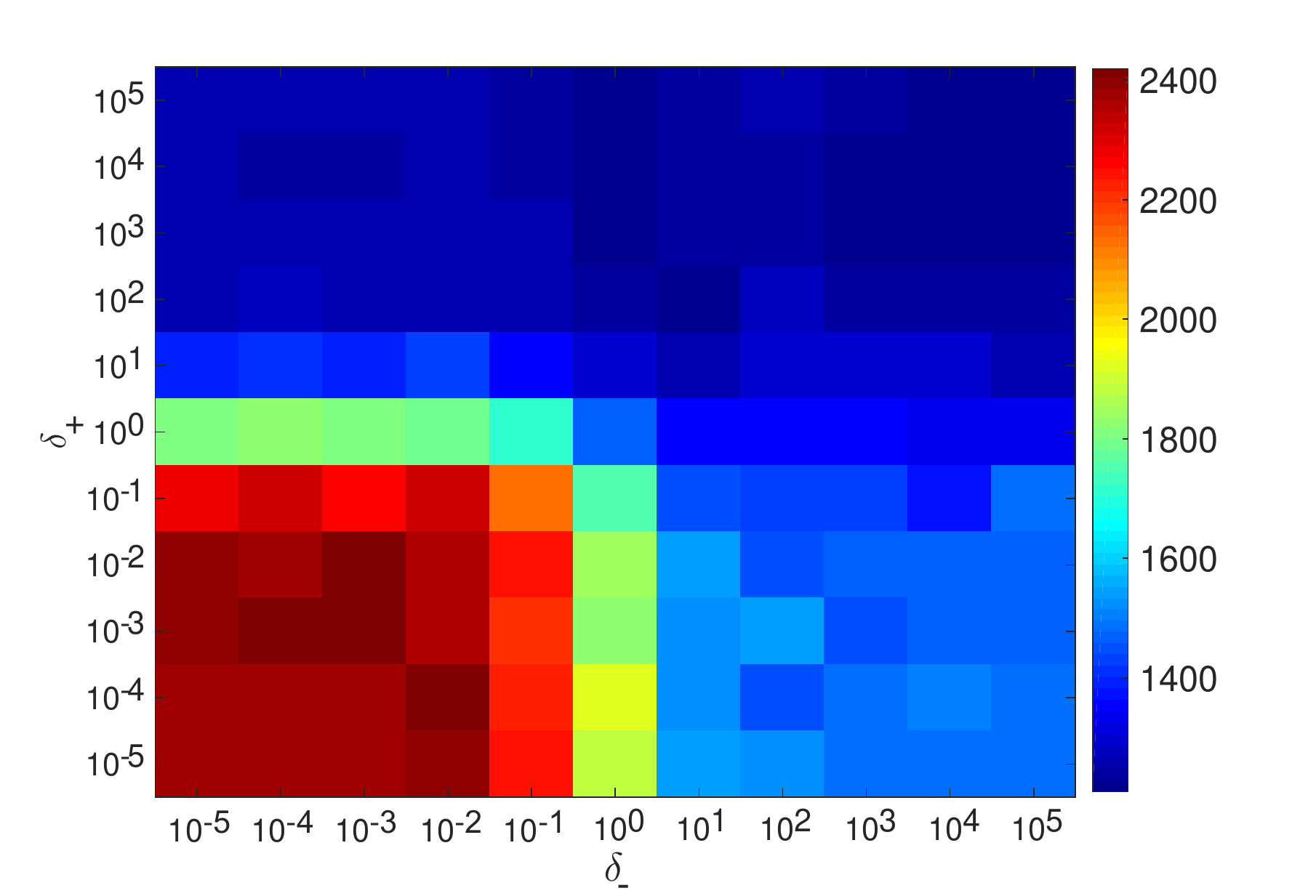}}
      \centerline{(b) Sensorless, B=29000}
    \end{minipage}
    \vfill
    \begin{minipage}{0.47\linewidth}
      \centerline{\includegraphics[width=4.6cm]{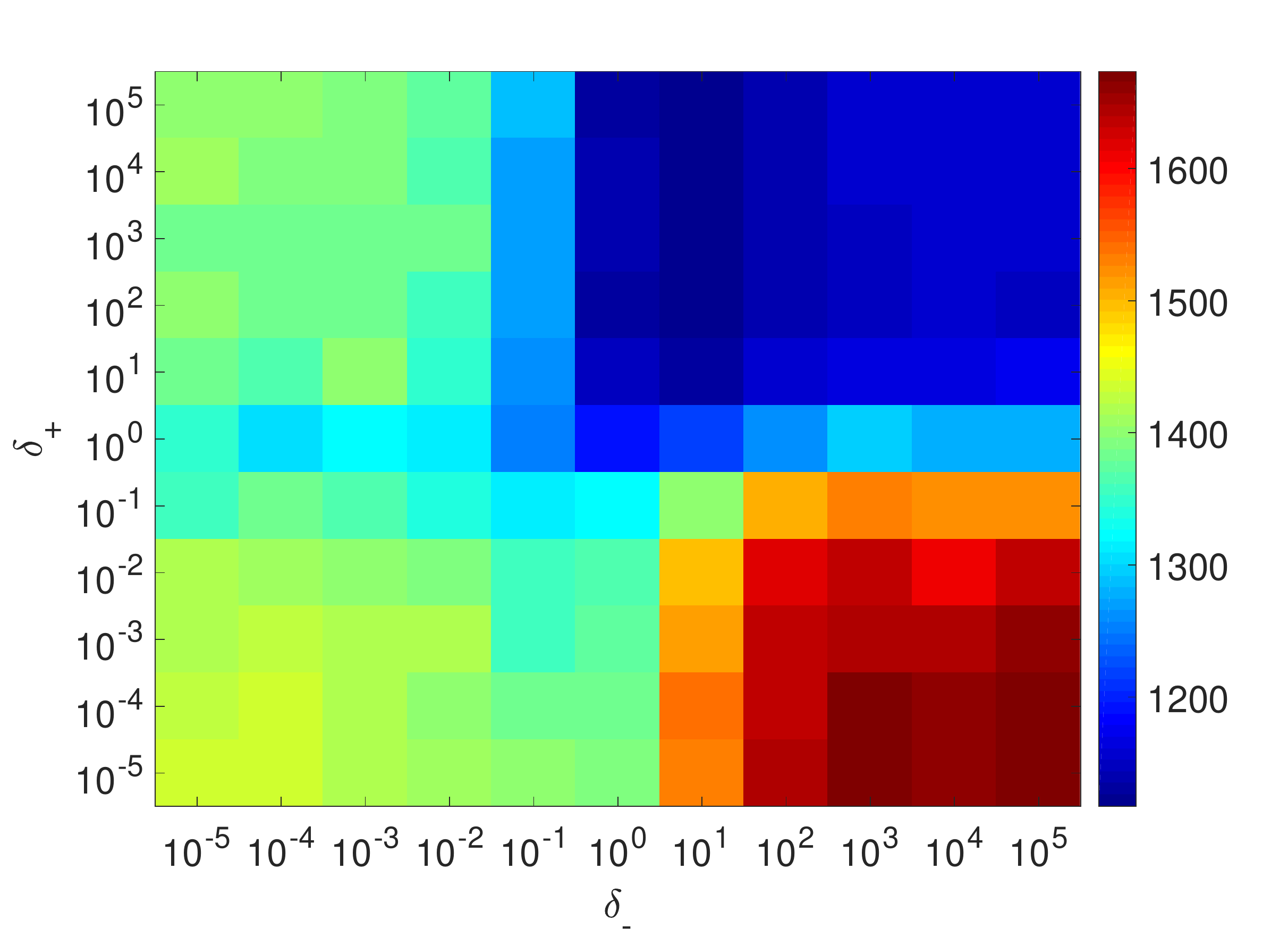}}
      \centerline{(c) w8a, B=32000}
    \end{minipage}
    \hfill
    \begin{minipage}{0.47\linewidth}
      \centerline{\includegraphics[width=4.6cm]{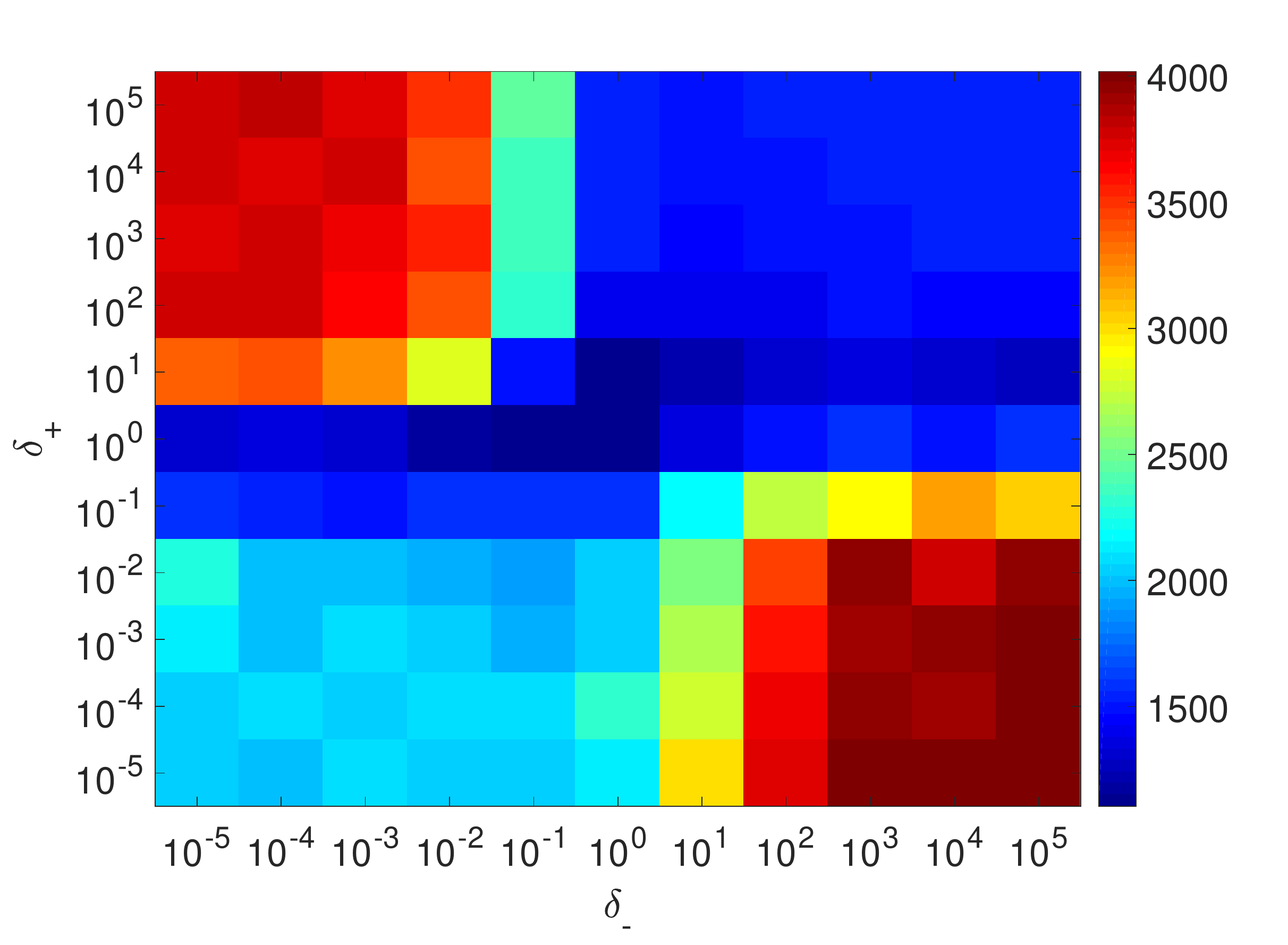}}
      \centerline{(d) KDDCUP08, B=50000}
    \end{minipage}
    \vspace{-0.1in}
    \caption{Cost Evaluation of query biases for OA3.}\label{cost_query_biases}
    \vspace{-0.15in}
\end{figure}

This subsection evaluates the influence of the query biases on $cost$ results under fixed budgets, where both query biases ($\delta_{+}$ and $\delta_{-}$) are selected from $[10^{-5},...,10^{5}]$, and other parameters are fixed. From Fig.~\ref{cost_query_biases}, we draw several observations.

First, the best results (\emph{i.e.,} deep blue) are usually achieved when $\delta_{+} \in \{10, 10^2,10^3,10^4\}$ and $\delta_{-} \in \{1, 10\}$. This observation suggests the potential settings of query biases.

Second, when both $\delta_{+}$ and $\delta_{-}$ are large (\emph{i.e.,} the upper right corner), OA3 obtains relatively good performance; while when both $\delta_{+}$ and $\delta_{-}$ are small (\emph{i.e.,} the bottom left corner), OA3 performs relatively bad. This observation further validates the findings in $sum$ results.

Finally, OA3 with large $\delta_{+}$ and small $\delta_{-}$ (the upper left corner) outperforms the performance with large $\delta_{-}$ and small $\delta_{+}$ (the bottom right corner). This means that OA3 performs better when querying more samples with the positive prediction and training itself by more positive samples. The main reason is that the positive samples are often more important in real-world tasks. Thus, our algorithms would be more effective in practical tasks due to the good algorithm characteristics, compared with the algorithms that treat all data equally, or tend to query more negative data.

\subsection{Cost Evaluation of Cost Weights}

\begin{figure}
    \begin{minipage}{0.47\linewidth}
      \centerline{\includegraphics[width=4.6cm]{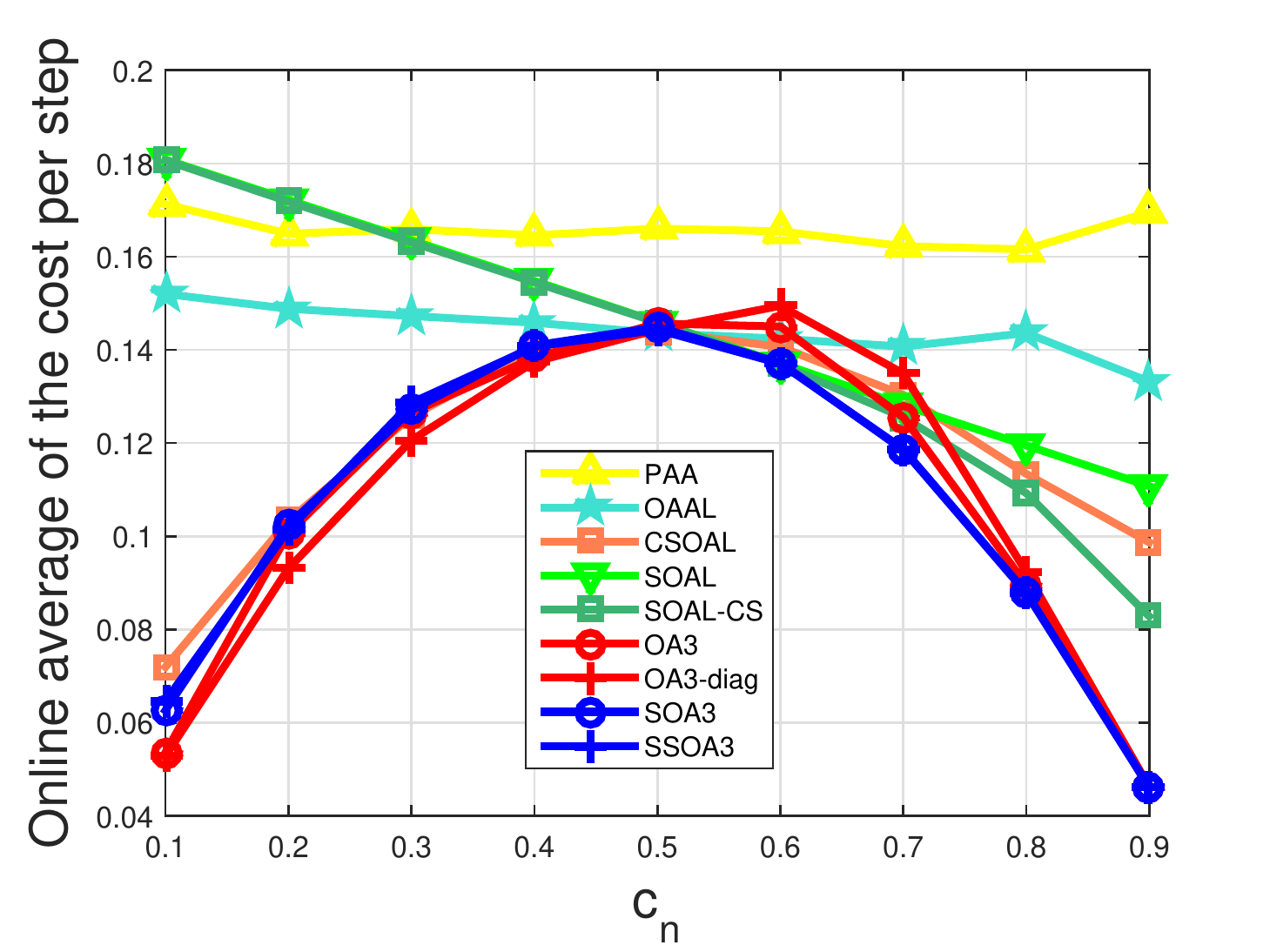}}
      \centerline{(a) protein, B=8500}
    \end{minipage}
    \hfill
    \begin{minipage}{0.47\linewidth}
      \centerline{\includegraphics[width=4.6cm]{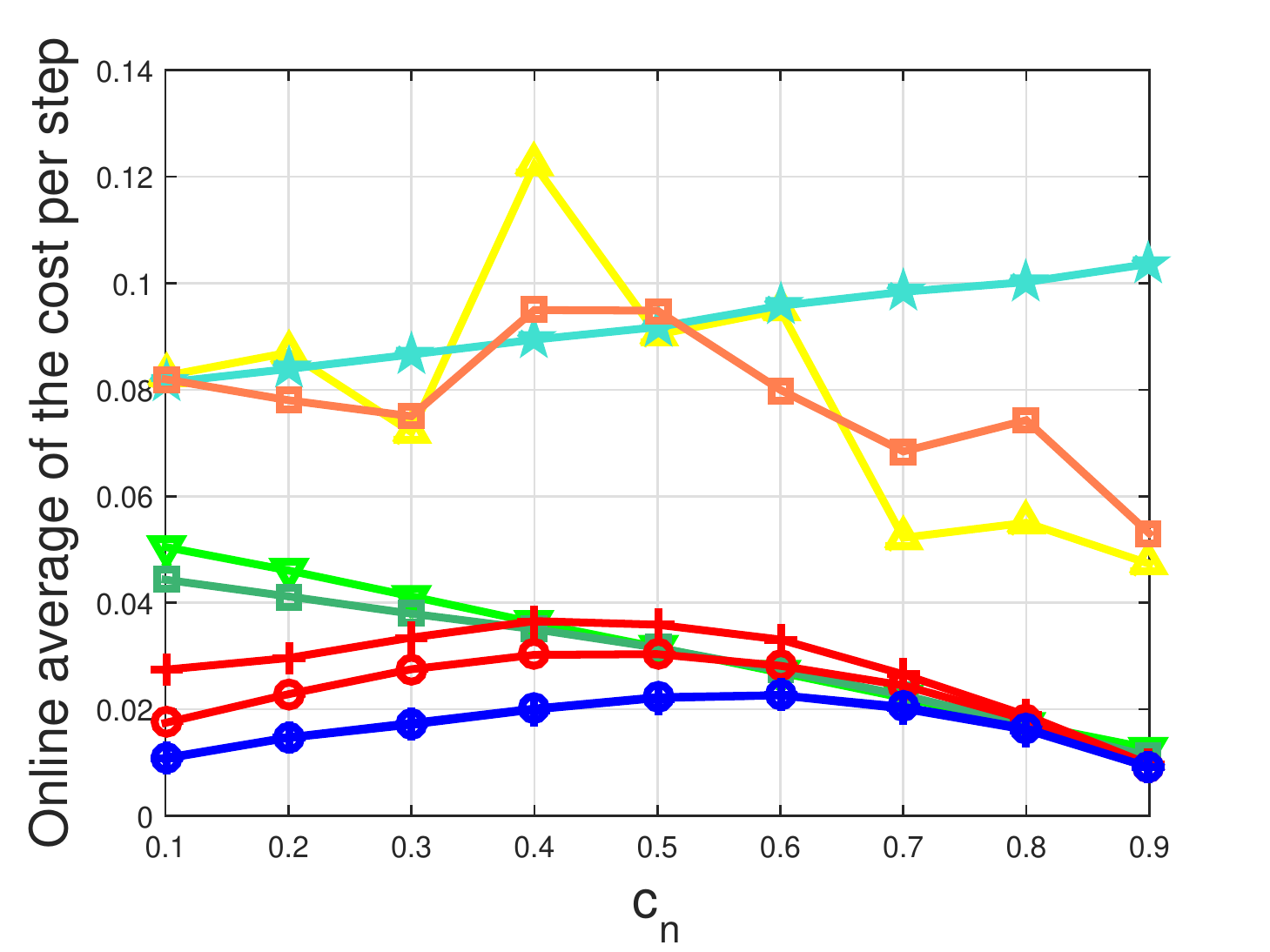}}
      \centerline{(b) Sensorless, B=29000}
    \end{minipage}
    \vfill
    \begin{minipage}{0.47\linewidth}
      \centerline{\includegraphics[width=4.6cm]{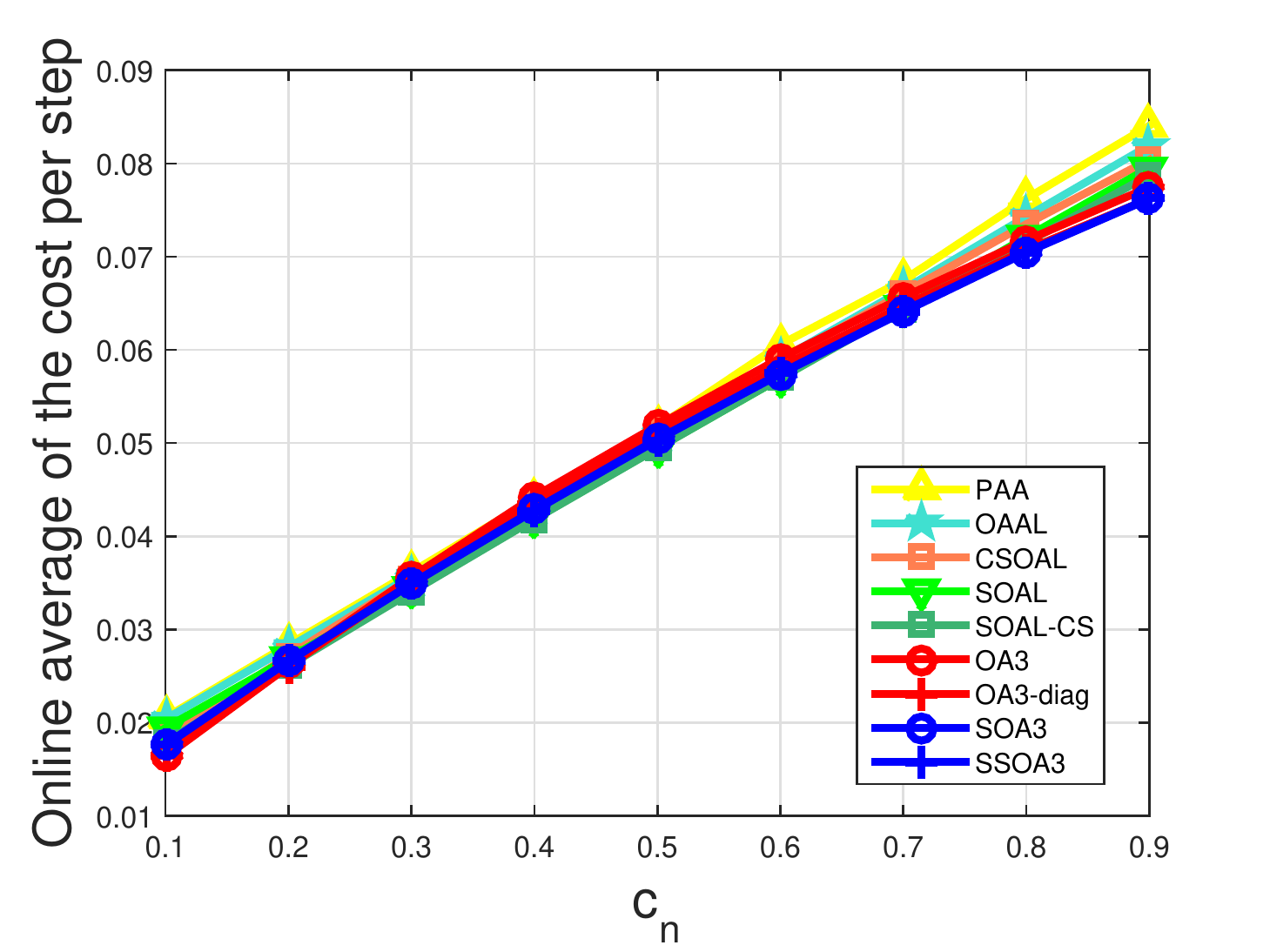}}
      \centerline{(c) w8a, B=32000}
    \end{minipage}
    \hfill
    \begin{minipage}{0.47\linewidth}
      \centerline{\includegraphics[width=4.6cm]{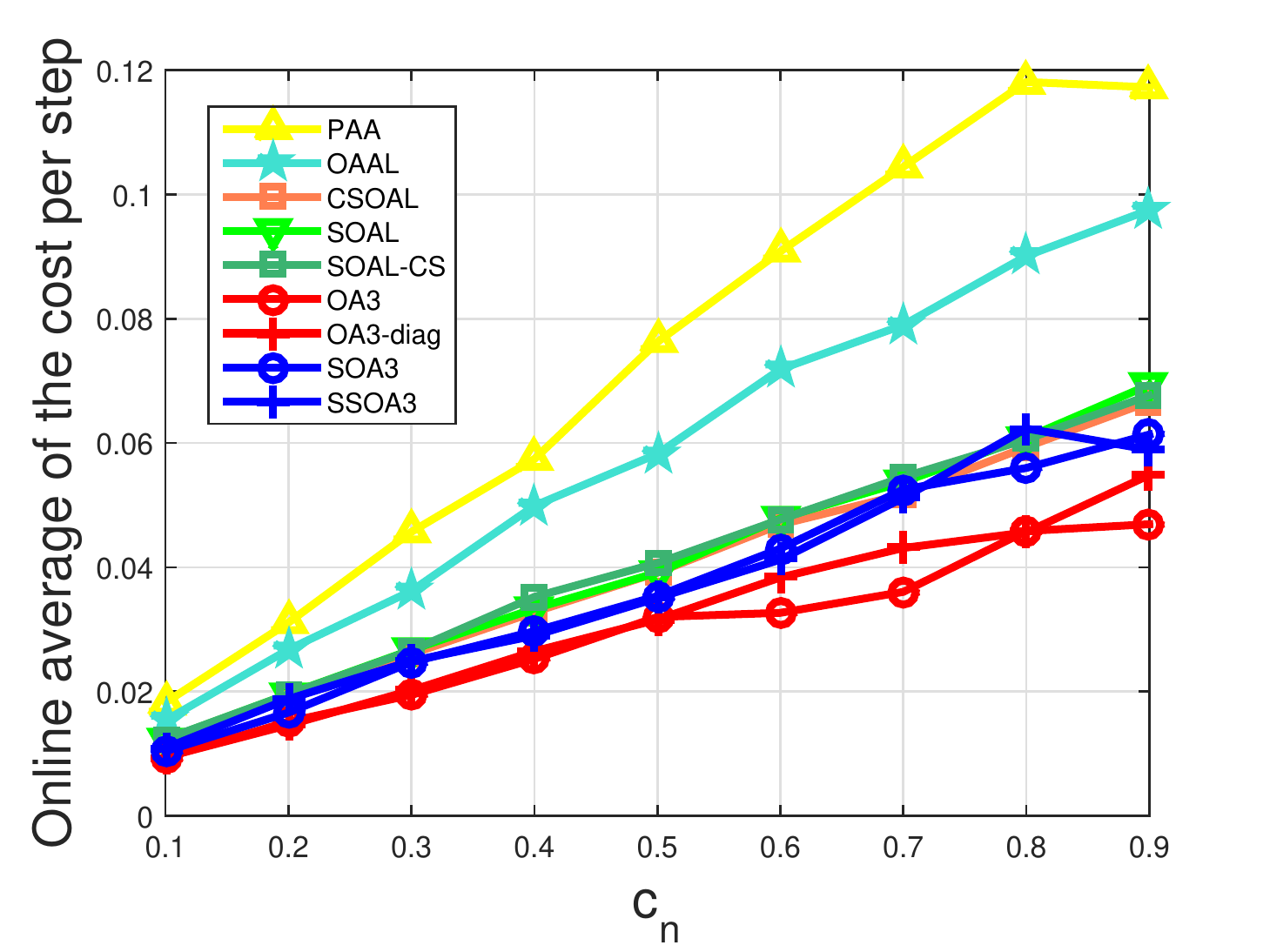}}
      \centerline{(d) KDDCUP08, B=50000}
    \end{minipage}
    \vspace{-0.1in}
    \caption{Performance of cost with varying cost weights.}\label{cost_cns}
    \vspace{-0.15in}
\end{figure}

In this subsection, we evaluate the influence of different cost weights on the $cost$ metric, i.e., $c_n$, where $c_p = 1 - c_n$. Fig.~\ref{cost_cns} summarize the results of $cost$ metric under fixed budgets.

From the results, we find our proposed algorithms consistently outperform all other algorithms with different weights. This observation shows that OA3 based algorithms have a wide selection range of cost weights, which further validates the effectiveness of our proposed methods again.

\subsection{Cost Evaluation of Learning Rate}

In this subsection, we evaluate the influence of learning rate in terms of the $cost$ performances. We examine proposed methods with different learning rates $\eta$ from $[10^{-4}, 10^{-3}, ..., 10^{3},10^{4}]$.

\begin{figure}
    \begin{minipage}{0.47\linewidth}
      \centerline{\includegraphics[width=4.6cm]{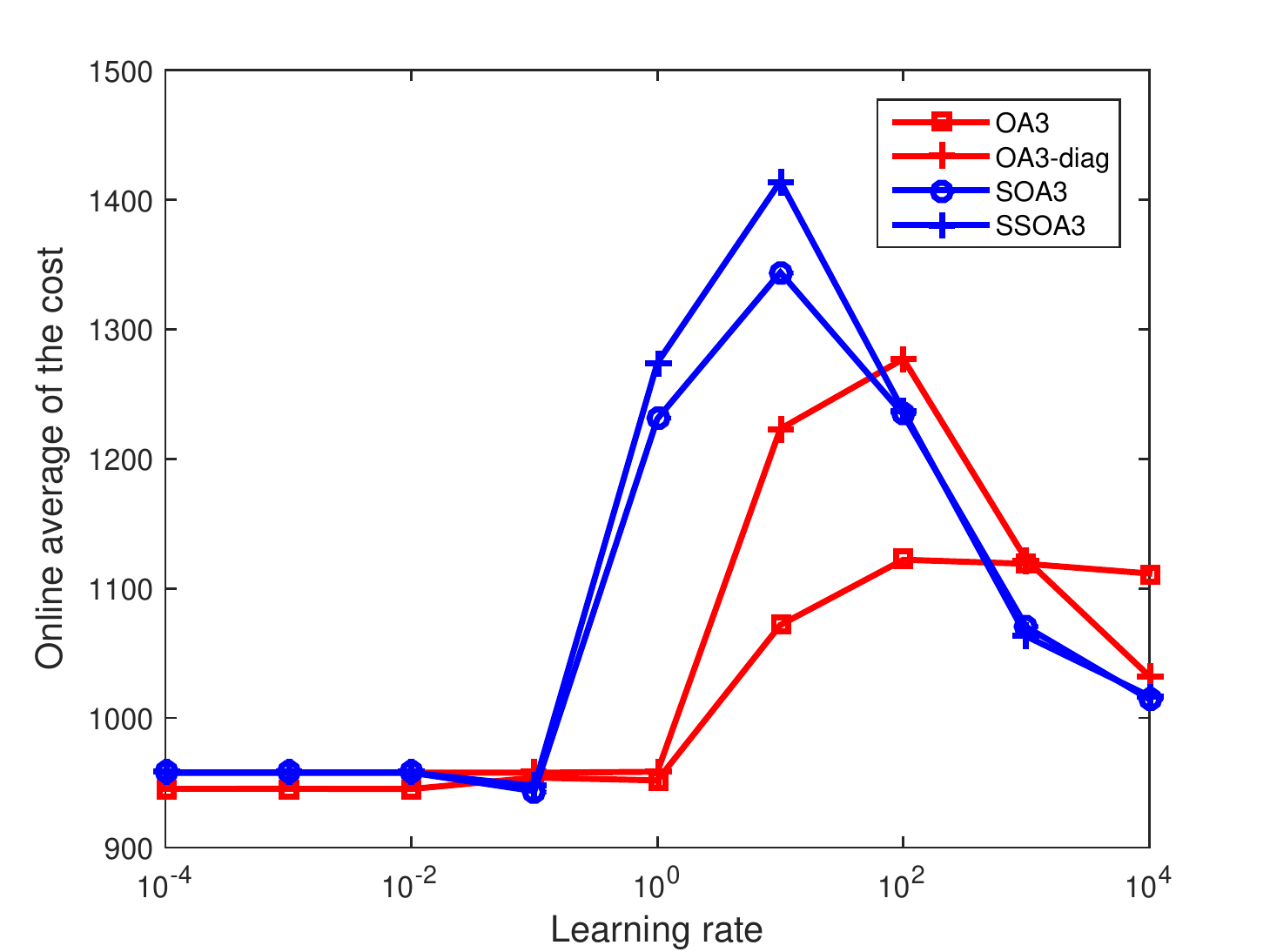}}
      \centerline{(a) protein, B=8500}
    \end{minipage}
    \hfill
    \begin{minipage}{0.47\linewidth}
      \centerline{\includegraphics[width=4.6cm]{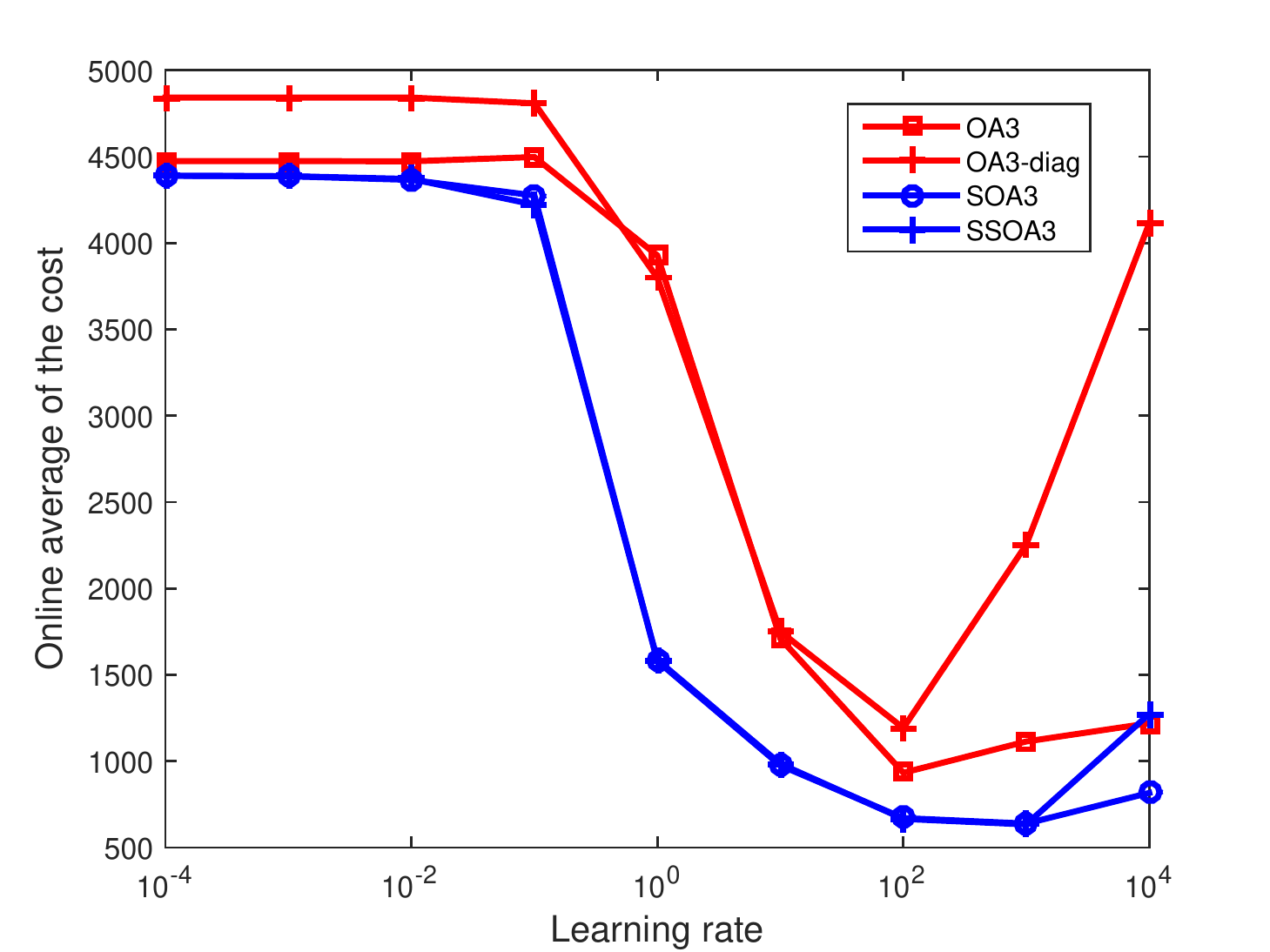}}
      \centerline{(b) Sensorless, B=29000}
    \end{minipage}
    \vfill
    \begin{minipage}{0.47\linewidth}
      \centerline{\includegraphics[width=4.6cm]{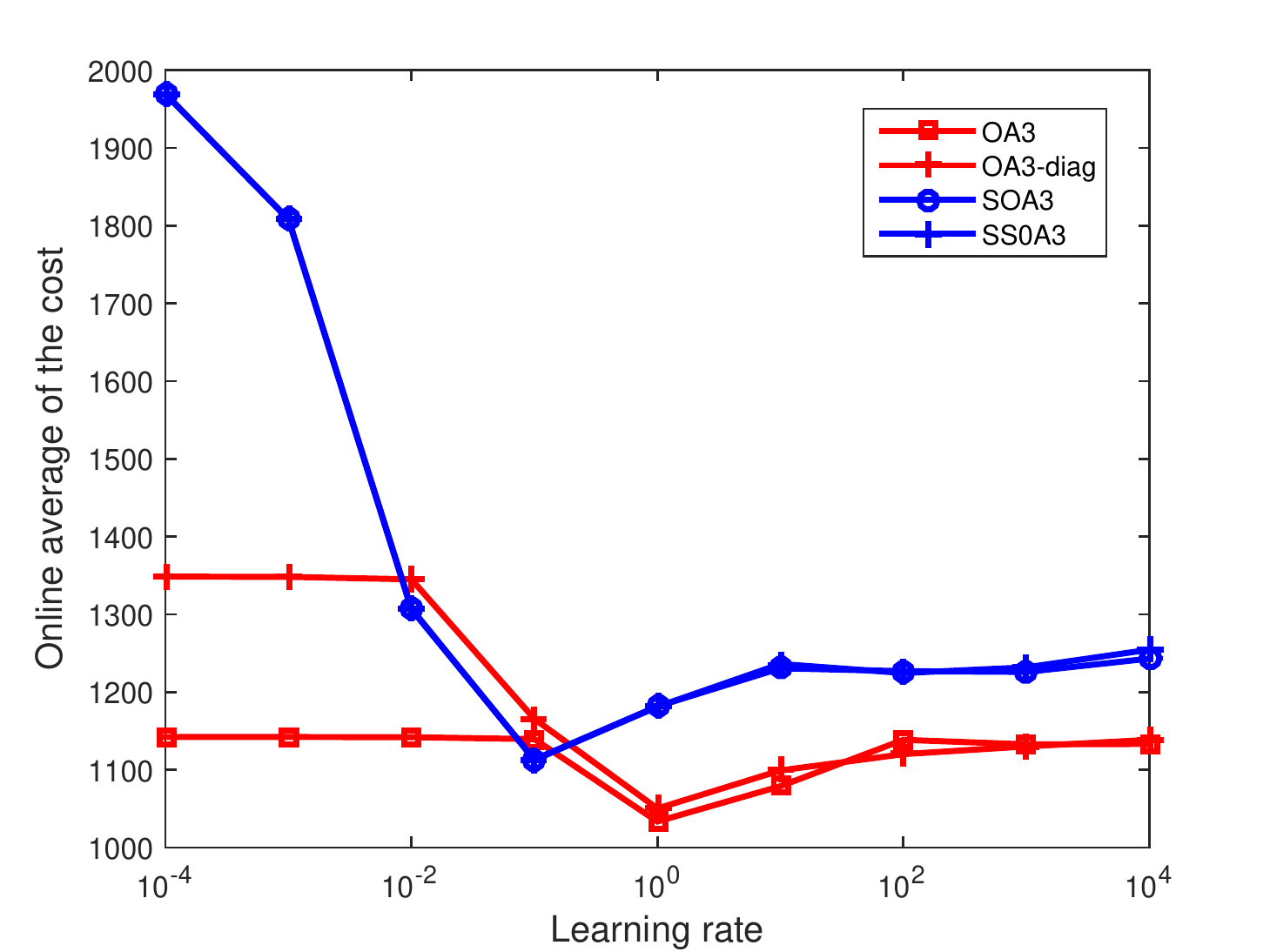}}
      \centerline{(c) w8a, B=32000}
    \end{minipage}
    \hfill
    \begin{minipage}{0.47\linewidth}
      \centerline{\includegraphics[width=4.6cm]{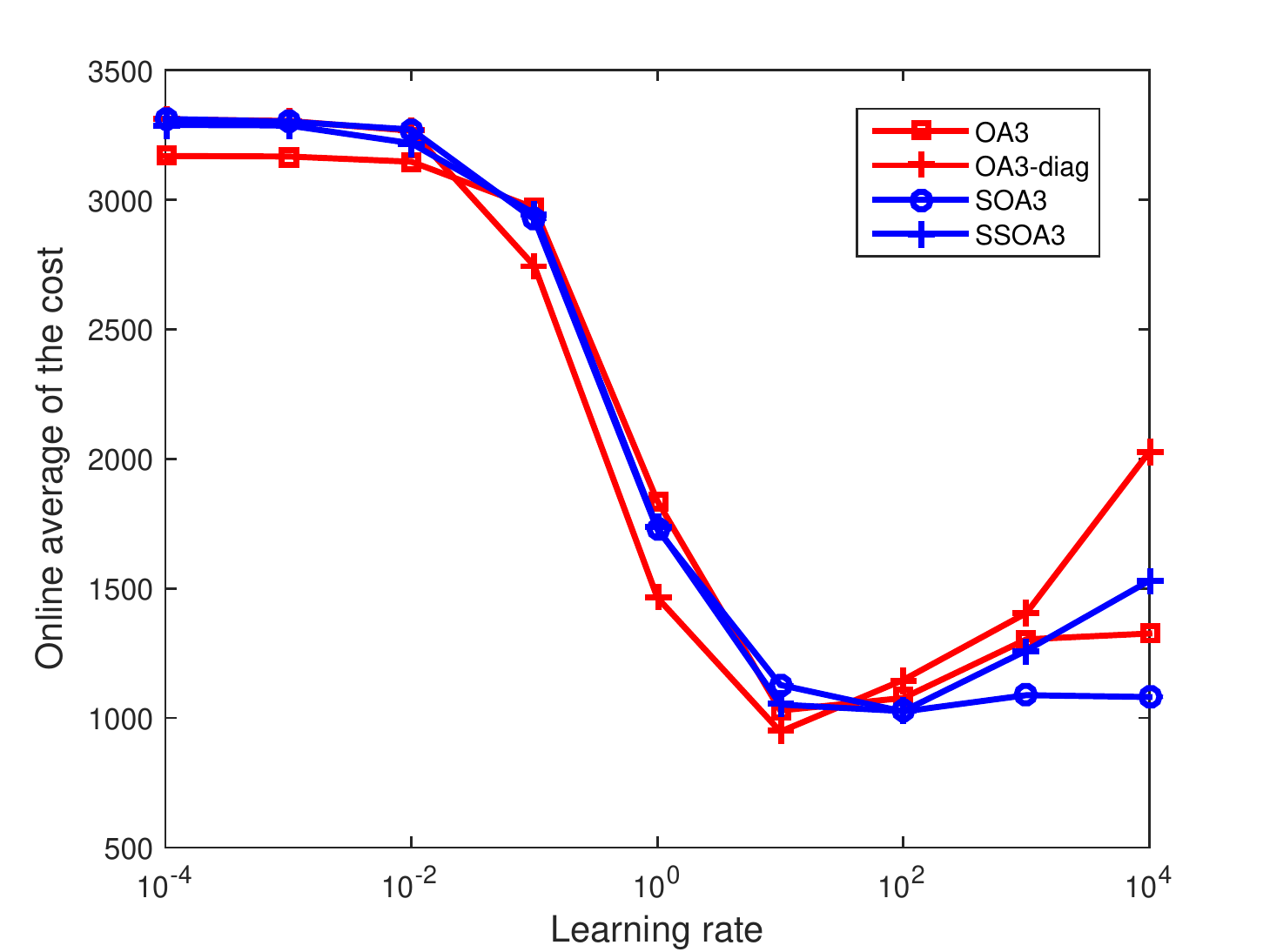}}
      \centerline{(d) KDDCUP08, B=50000}
    \end{minipage}
    \vspace{-0.1in}
    \caption{Performance of cost under varying learning rates.}\label{cost_learning_rates}
    \vspace{-0.15in}
\end{figure}

From Fig.~\ref{cost_learning_rates}, results show the suitable range of learning rates for different datasets. We find OA3 algorithms achieve the best result on most datasets. Moreover, SOA3 and SSOA3 can achieve a relatively good performance on most datasets, and sometimes even better, compared with OA3, which also confirms that the sketched versions of OA3 are good choices to balance the performance and efficiency.

\subsection{Evaluation of Regularized Parameter}

\revise{In previous experiments, the regularization parameter is set to 1 (\ie, $\gamma\small{=}1$) by default. However, the rationality of this default setting has not been verified. In this subsection, we examine the performance of our algorithms with different regularization parameters $\gamma$ from $[10^{-4},10^{-3},...,10^{3},10^{4}]$.}

\begin{figure}
    \begin{minipage}{0.485\linewidth}
      \centerline{\includegraphics[width=4.7cm]{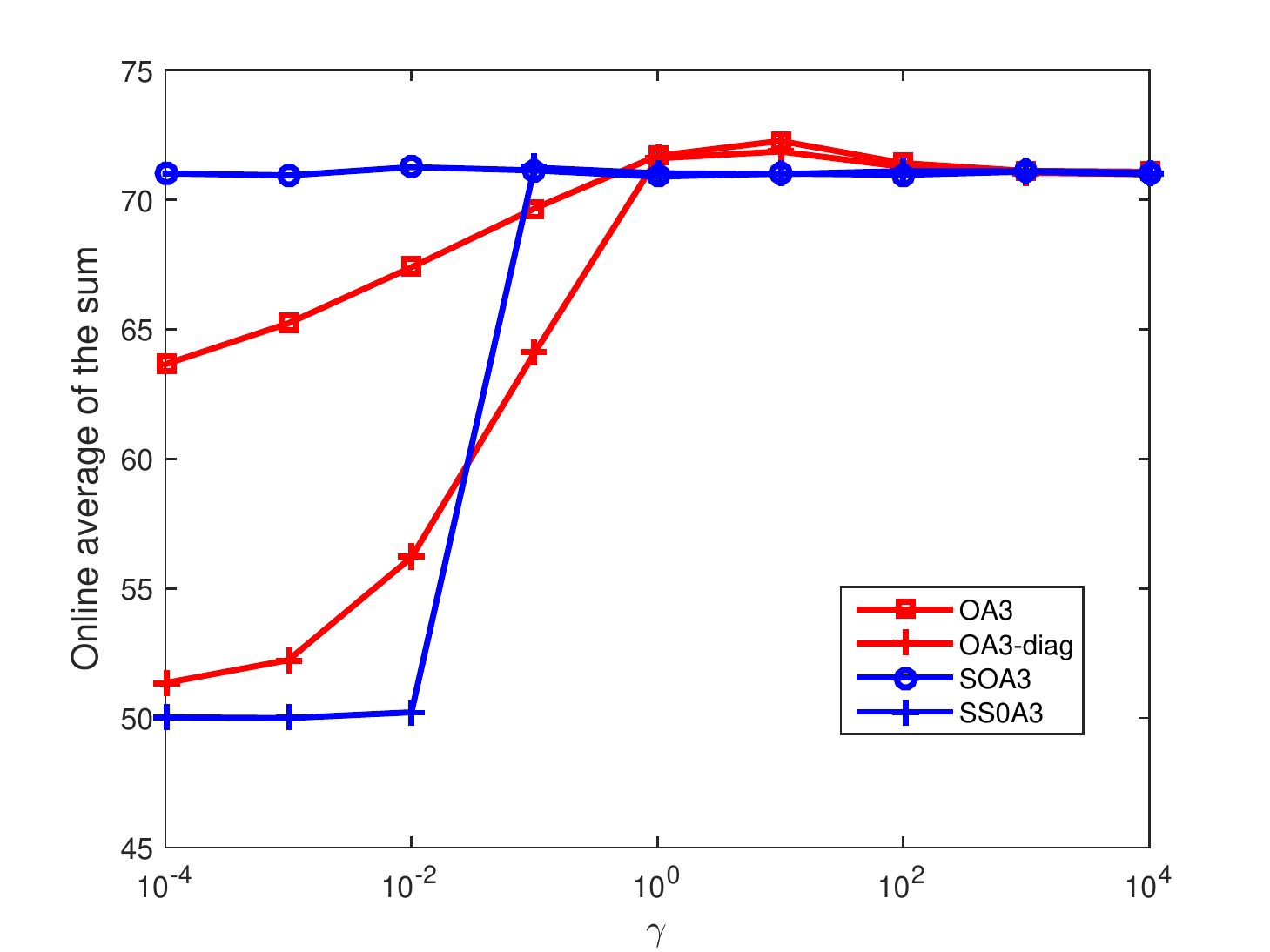}}
      \centerline{(a) protein, B=8500}
    \end{minipage}
    \hfill
    \begin{minipage}{0.485\linewidth}
      \centerline{\includegraphics[width=4.7cm]{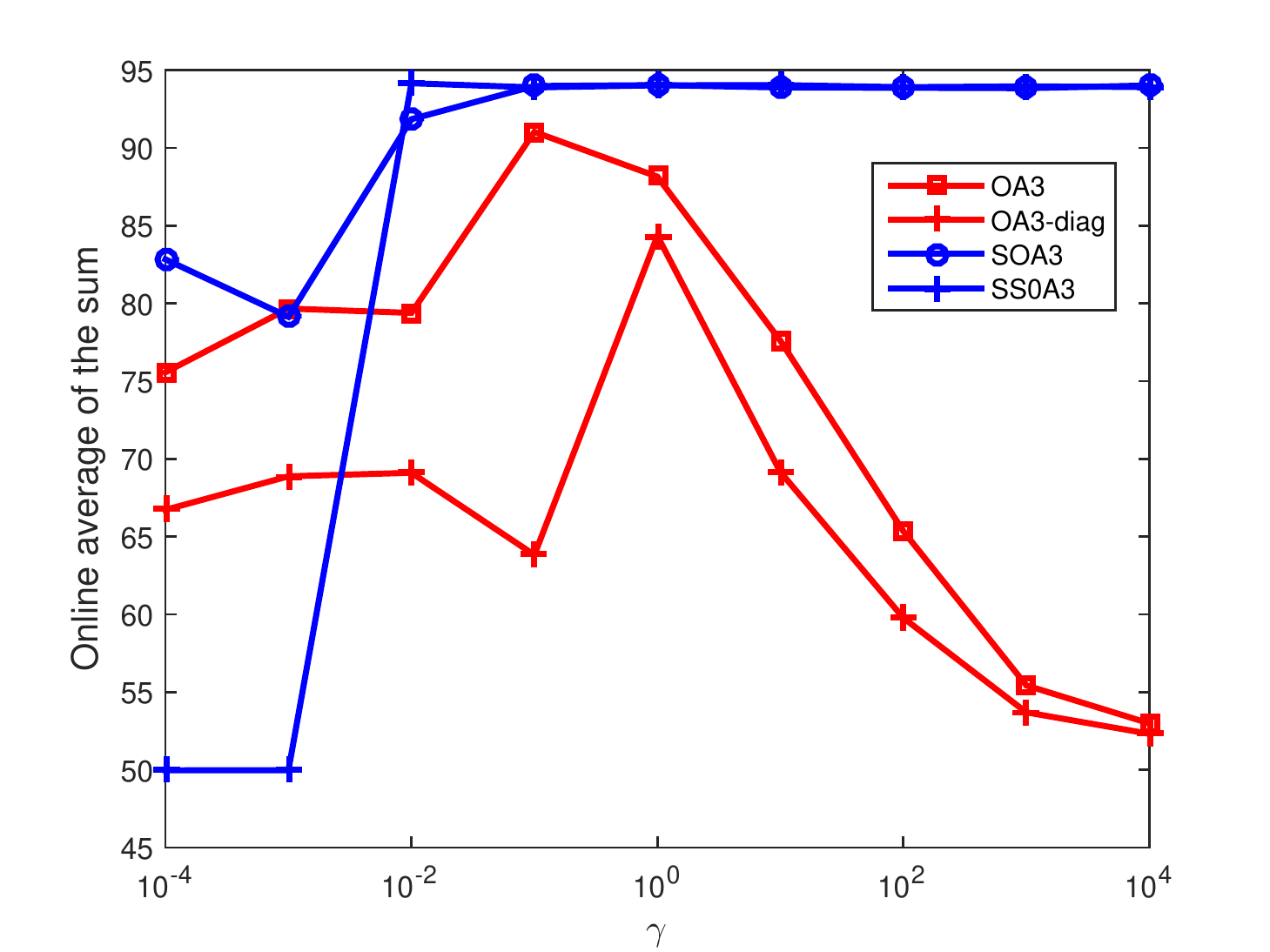}}
      \centerline{(b) Sensorless, B=29000}
    \end{minipage}
    \vfill
    \begin{minipage}{0.485\linewidth}
      \centerline{\includegraphics[width=4.7cm]{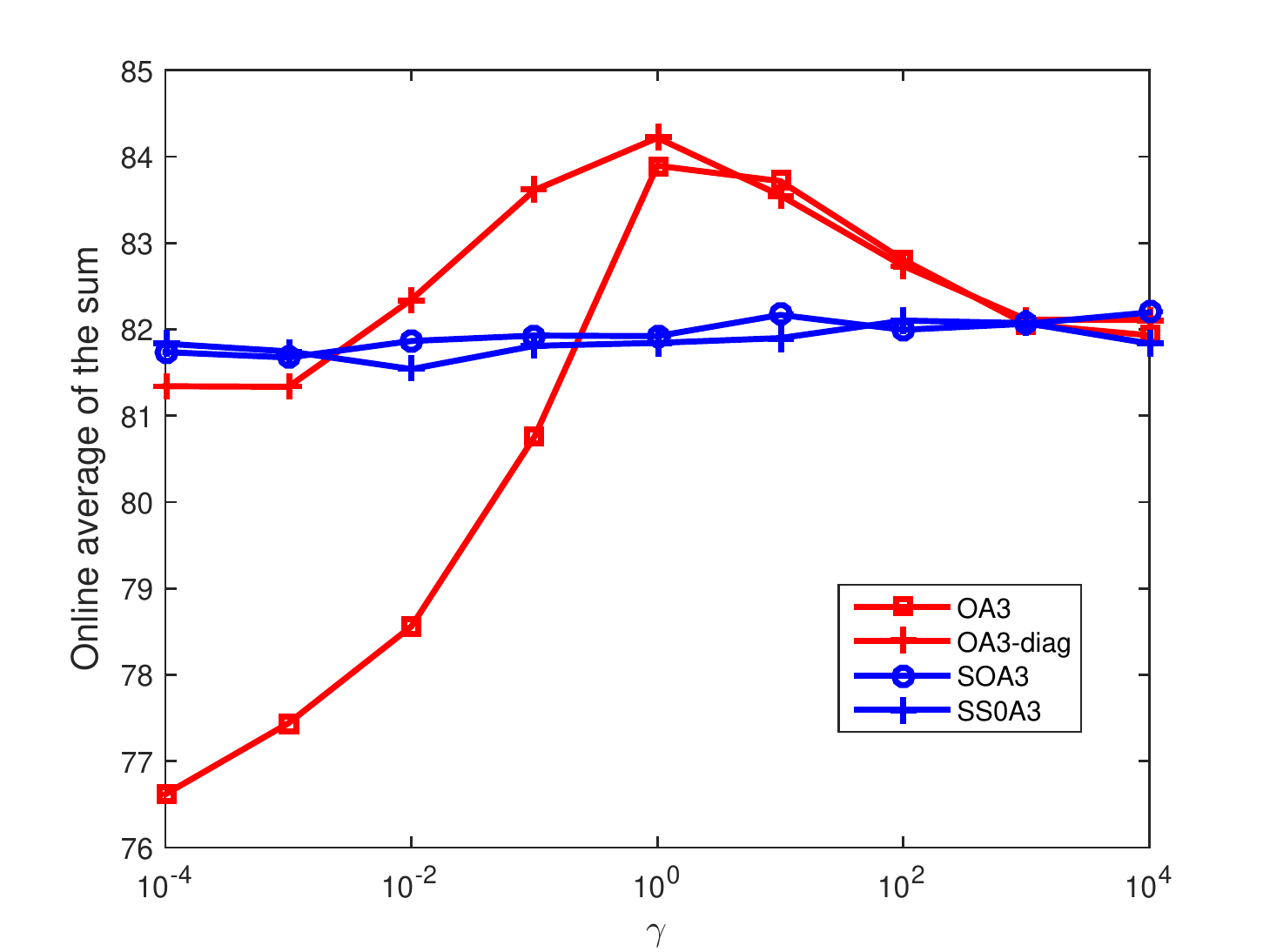}}
      \centerline{(c) w8a, B=32000}
    \end{minipage}
    \hfill
    \begin{minipage}{0.485\linewidth}
      \centerline{\includegraphics[width=4.7cm]{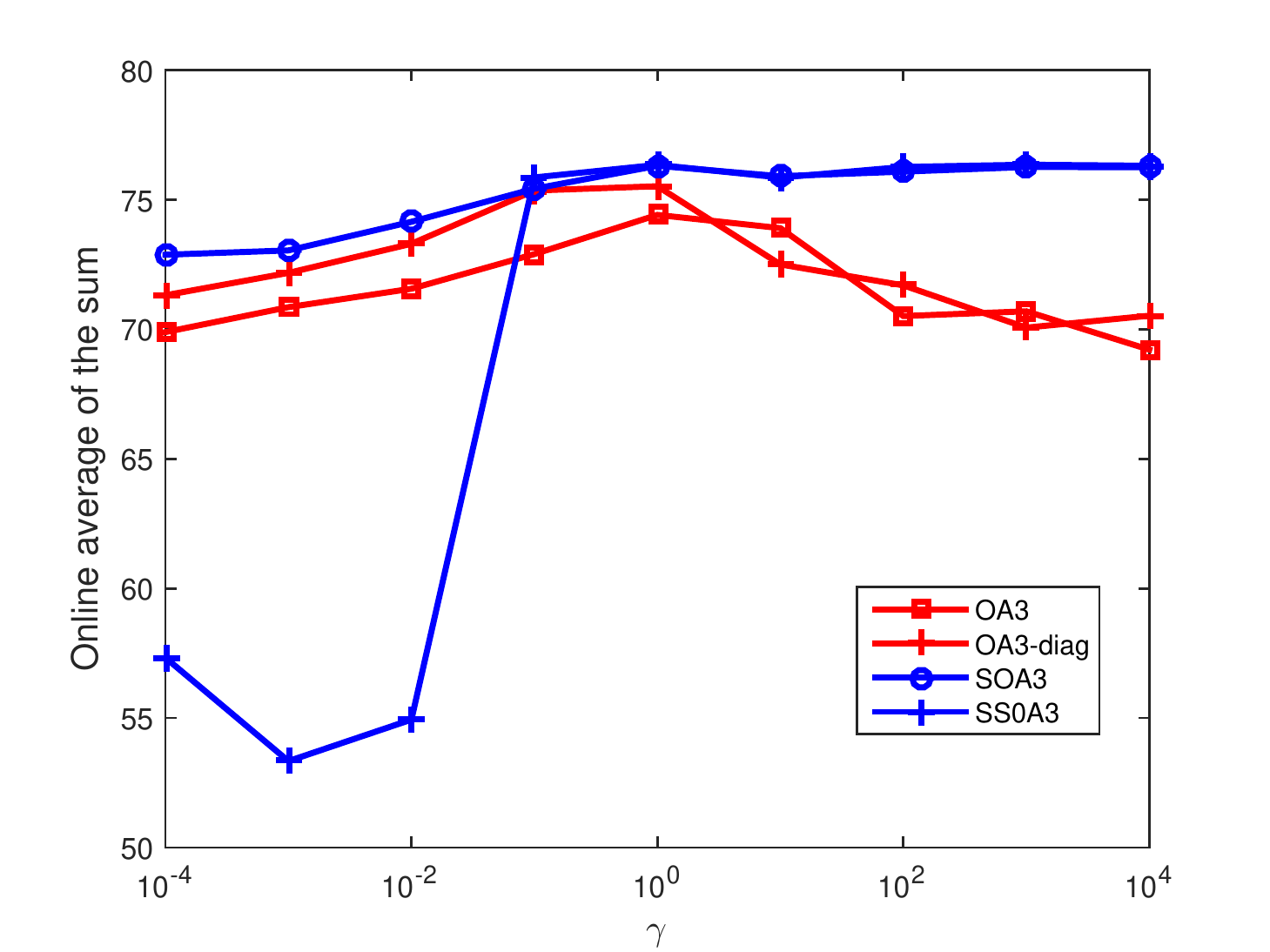}}
      \centerline{(d) KDDCUP08, B=50000}
    \end{minipage}
    \vspace{-0.1in}
    \caption{\yifan{Performance of sum with varying regularized factors.}}\label{sum_gammas}
    \vspace{-0.15in}
\end{figure}

\begin{figure}
    \begin{minipage}{0.47\linewidth}
      \centerline{\includegraphics[width=4.6cm]{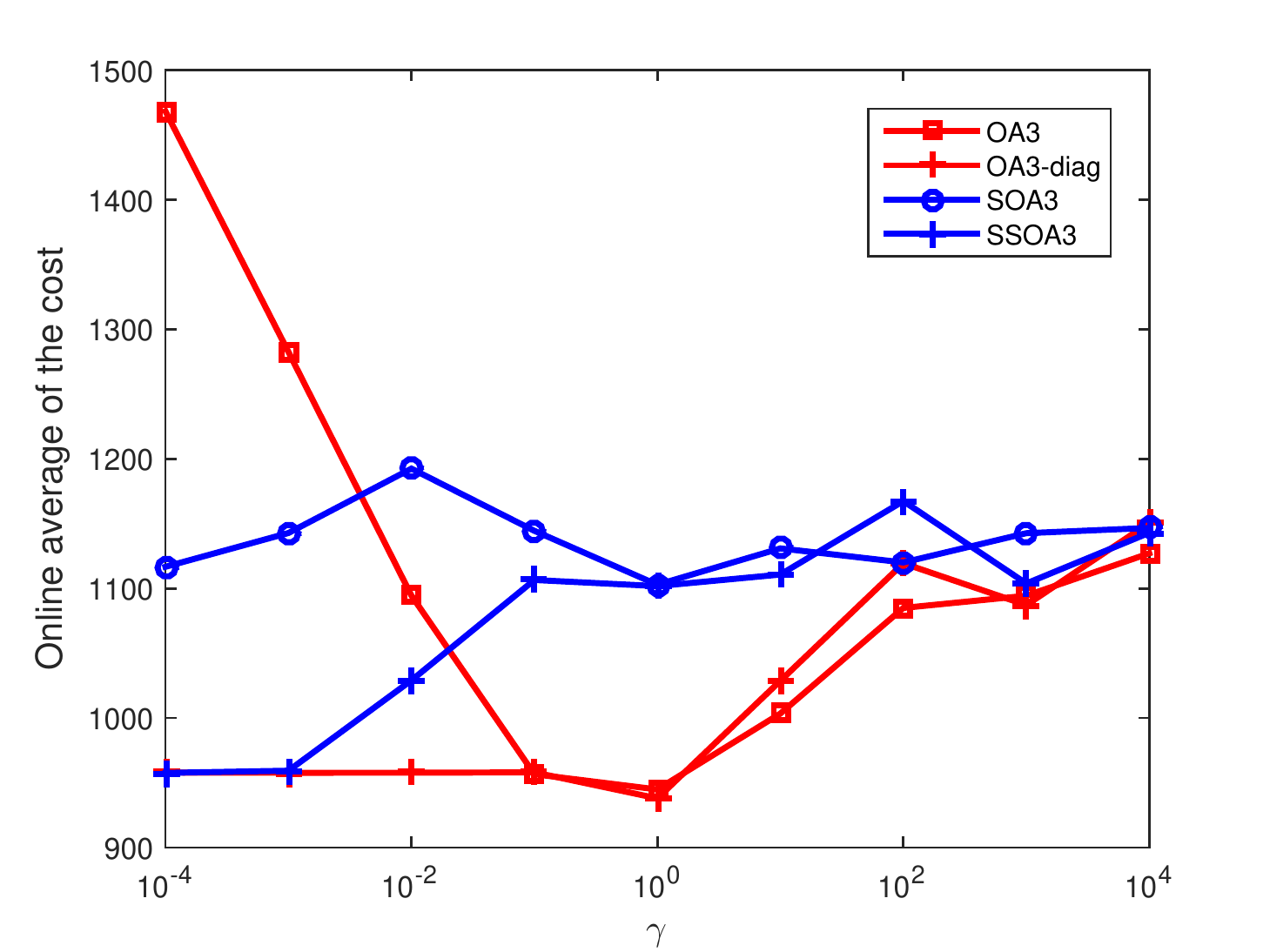}}
      \centerline{(a) protein, B=8500}
    \end{minipage}
    \hfill
    \begin{minipage}{0.47\linewidth}
      \centerline{\includegraphics[width=4.6cm]{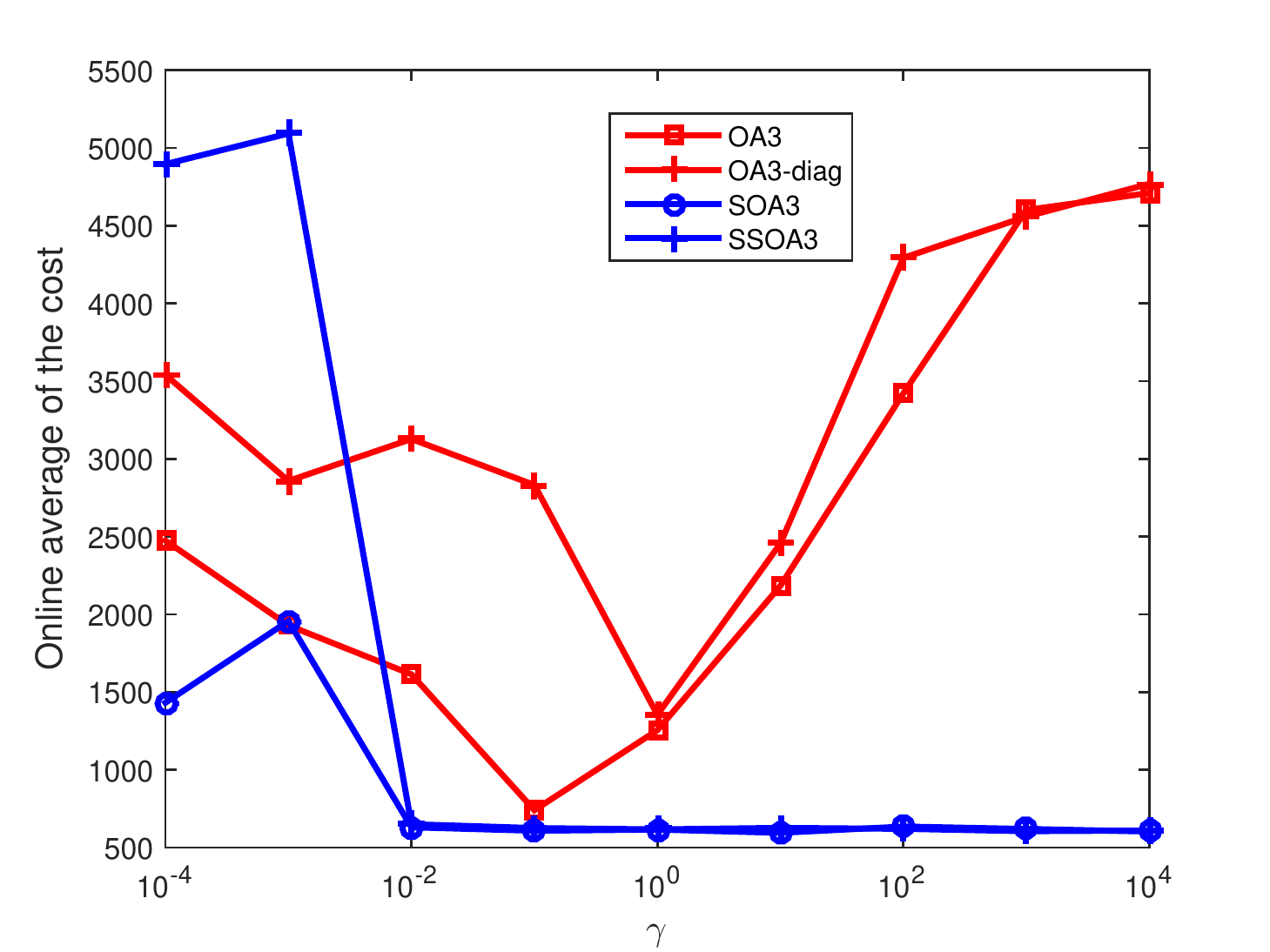}}
      \centerline{(b) Sensorless, B=29000}
    \end{minipage}
    \vfill
    \begin{minipage}{0.47\linewidth}
      \centerline{\includegraphics[width=4.6cm]{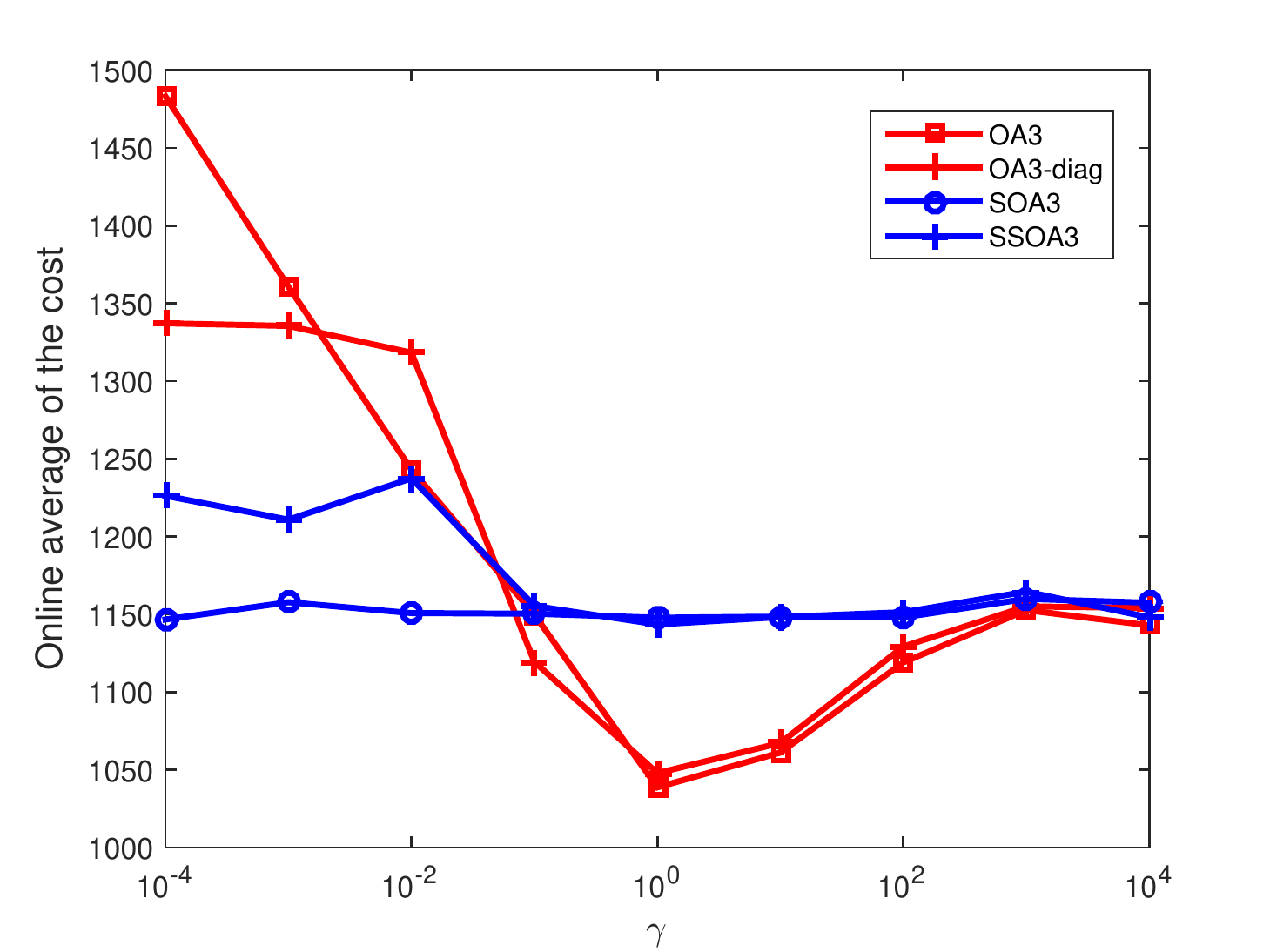}}
      \centerline{(c) w8a, B=32000}
    \end{minipage}
    \hfill
    \begin{minipage}{0.47\linewidth}
      \centerline{\includegraphics[width=4.6cm]{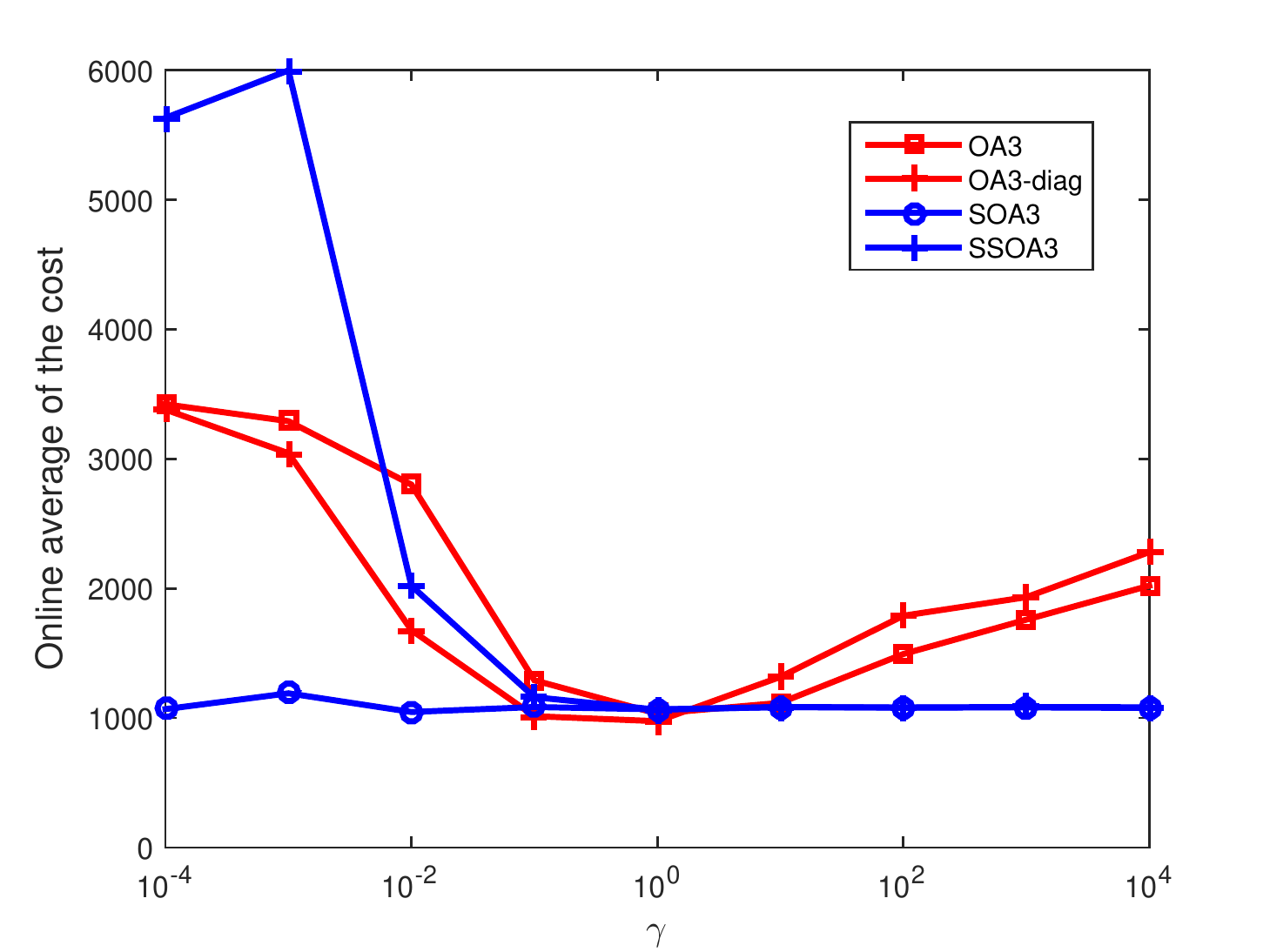}}
      \centerline{(d) KDDCUP08, B=50000}
    \end{minipage}
    \vspace{-0.1in}
    \caption{Performance of cost with varying regularized factors.}\label{cost_gammas}
    \vspace{-0.15in}
\end{figure}

\revise{From Fig.~\ref{sum_gammas} and~\ref{cost_gammas}, we find the optimal selection of $\gamma$ depends on datasets and methods. Nevertheless, in most cases, the setting $\gamma \small{=} 1$ achieves the best or fairly good performance. This observation validates the practical value of our algorithms with the default setting in real-world datasets.}

\subsection{Evaluation of Query Strategies}

In this subsection, we examine our asymmetric query strategy. We compare OA3 with two variants of OA3 with weak query rules: (1) OA3 with the ``First come first served`` strategy (\textbf{OA3-F}), which is the pure updating version without query strategies; (2) OA3 with the random query strategy (\textbf{OA3-R}).

\begin{figure}
    \begin{minipage}{0.47\linewidth}
      \centerline{\includegraphics[width=4.6cm]{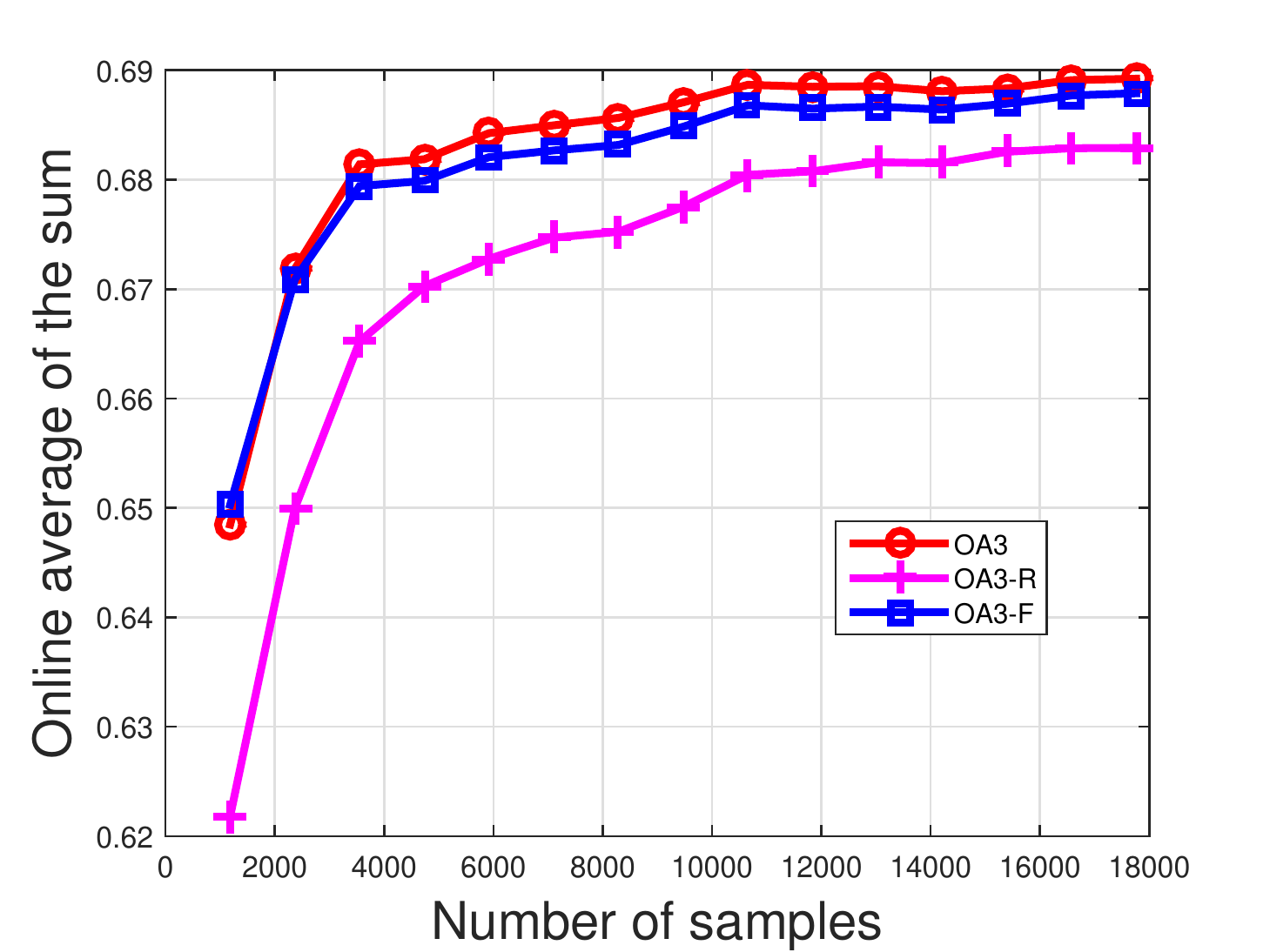}}
      \centerline{(a) protein, \shuai{B=1200}}
    \end{minipage}
    \hfill
    \begin{minipage}{0.47\linewidth}
      \centerline{\includegraphics[width=4.6cm]{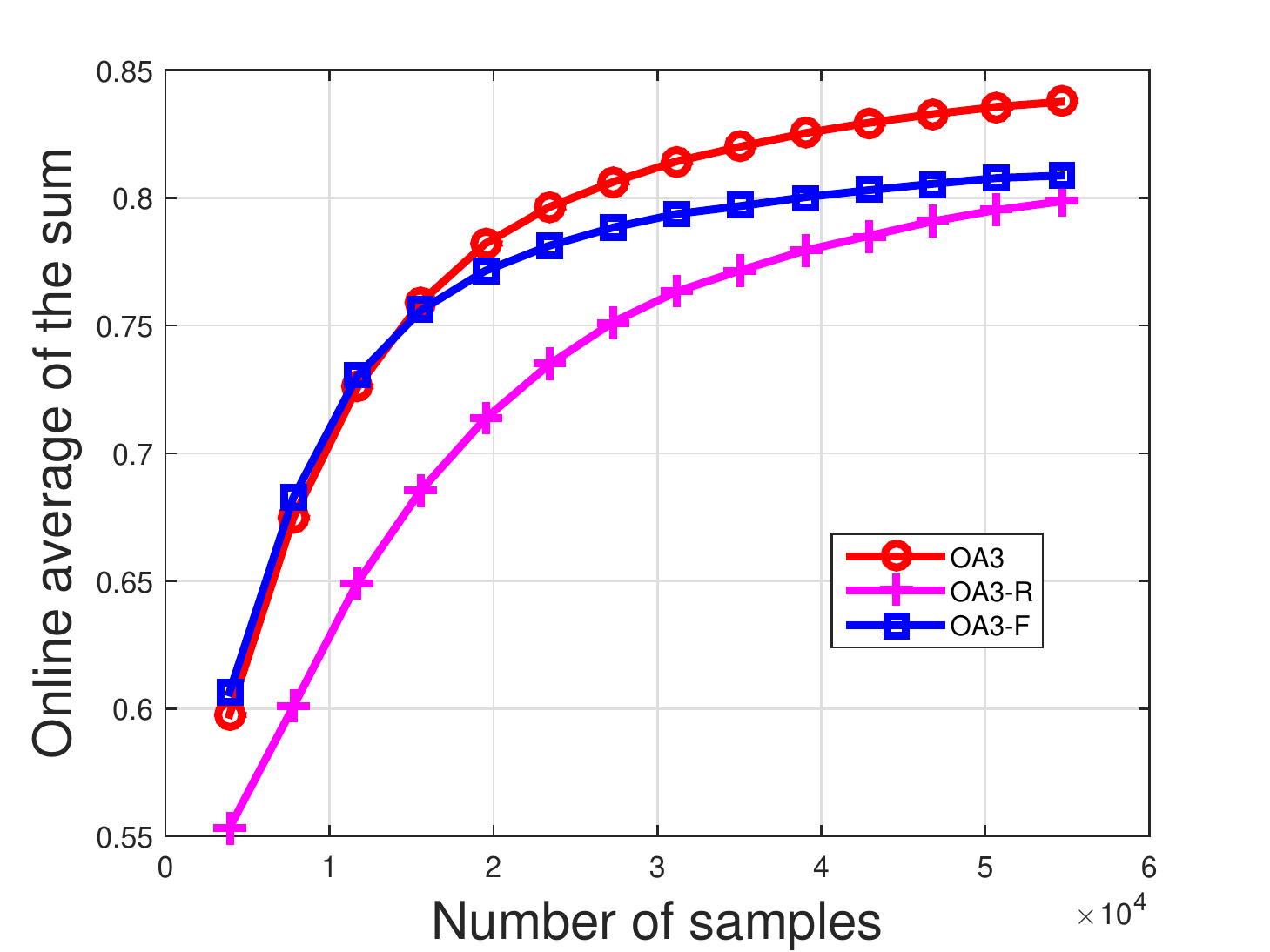}}
      \centerline{(b) Sensorless, \shuai{B=10000}}
    \end{minipage}
    \vfill
    \begin{minipage}{0.47\linewidth}
      \centerline{\includegraphics[width=4.6cm]{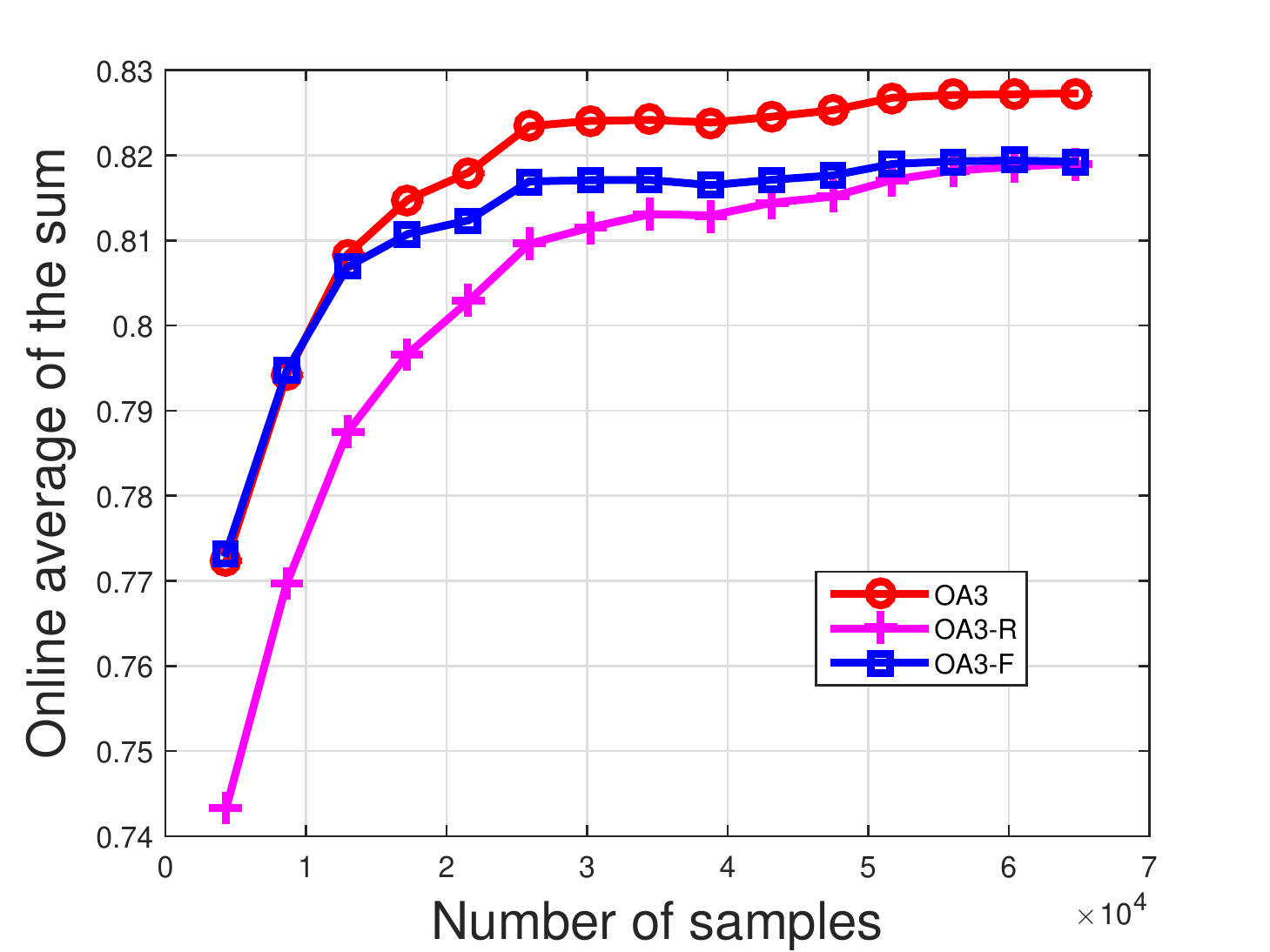}}
      \centerline{(c) w8a, \shuai{B=8000}}
    \end{minipage}
    \hfill
    \begin{minipage}{0.47\linewidth}
      \centerline{\includegraphics[width=4.6cm]{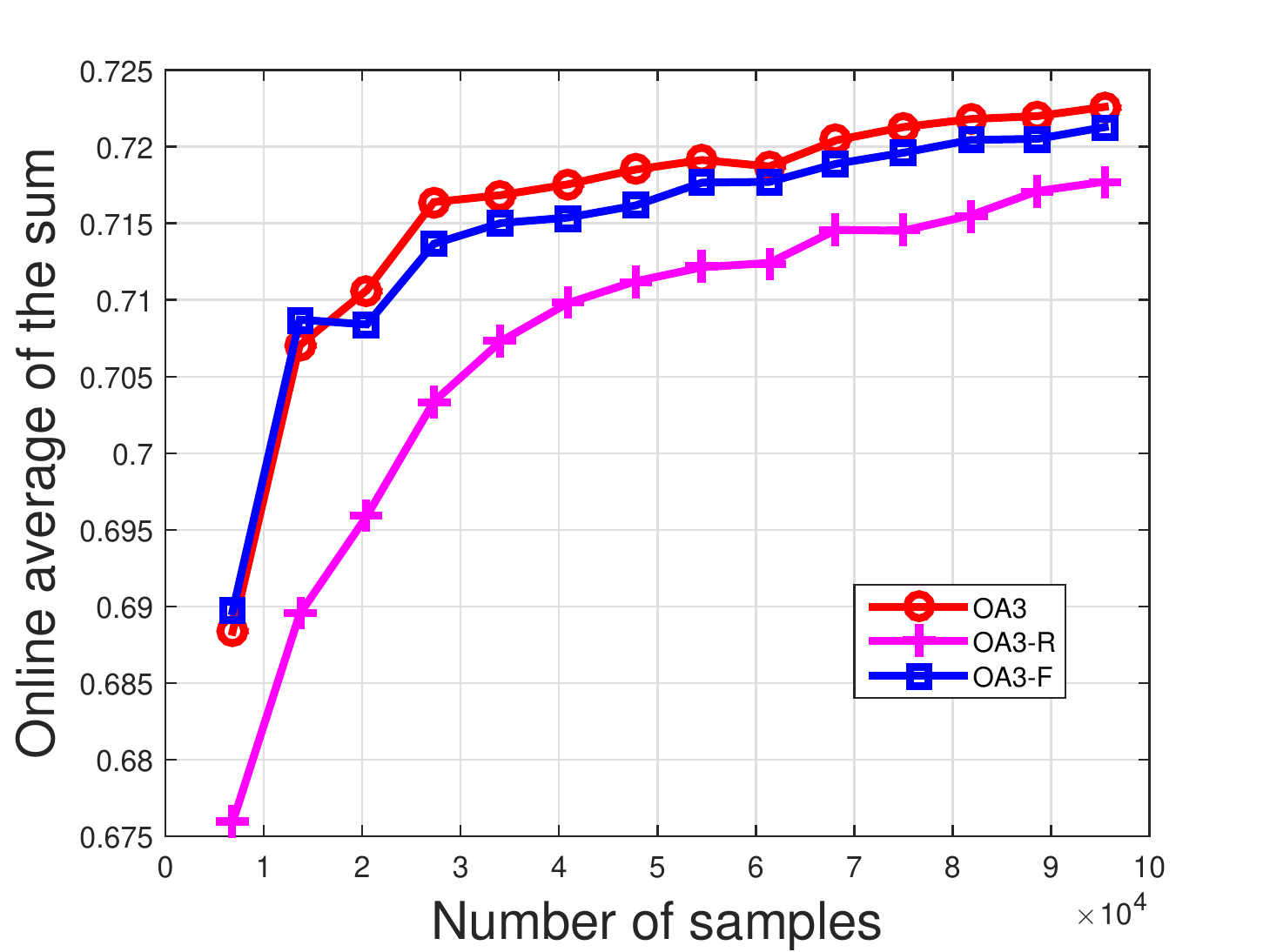}}
      \centerline{(d) KDDCUP08, \shuai{B=10000}}
    \end{minipage}
    \vspace{-0.1in}
    \caption{Sum evaluation of different query strategies.}\label{sum_query_strategy}
    \vspace{-0.15in}
\end{figure}

\begin{figure}
    \begin{minipage}{0.47\linewidth}
      \centerline{\includegraphics[width=4.6cm]{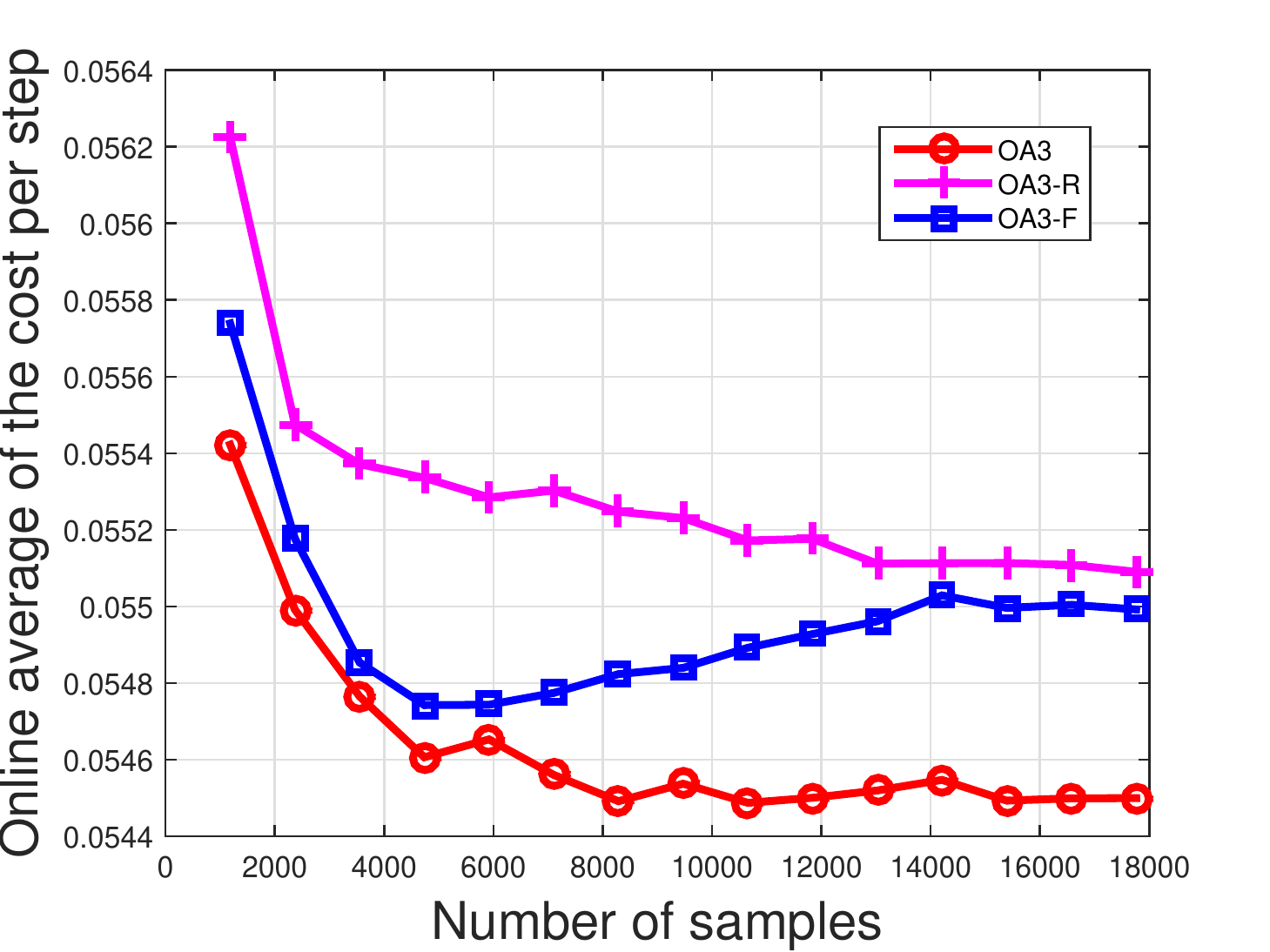}}
      \centerline{(a) protein, \shuai{B=1200}}
    \end{minipage}
    \hfill
    \begin{minipage}{0.47\linewidth}
      \centerline{\includegraphics[width=4.6cm]{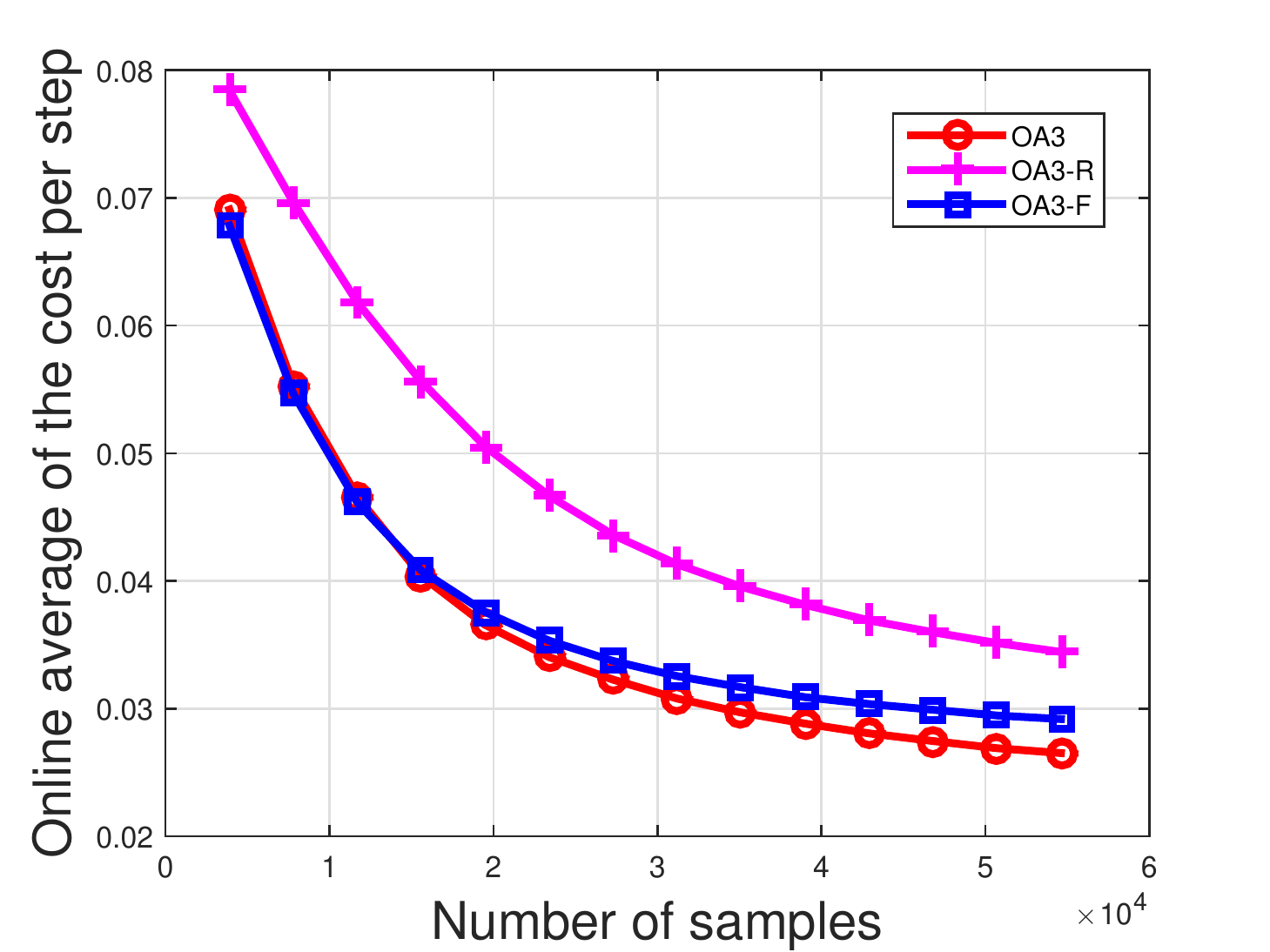}}
      \centerline{(b) Sensorless, \shuai{B=12000}}
    \end{minipage}
    \vfill
    \begin{minipage}{0.47\linewidth}
      \centerline{\includegraphics[width=4.6cm]{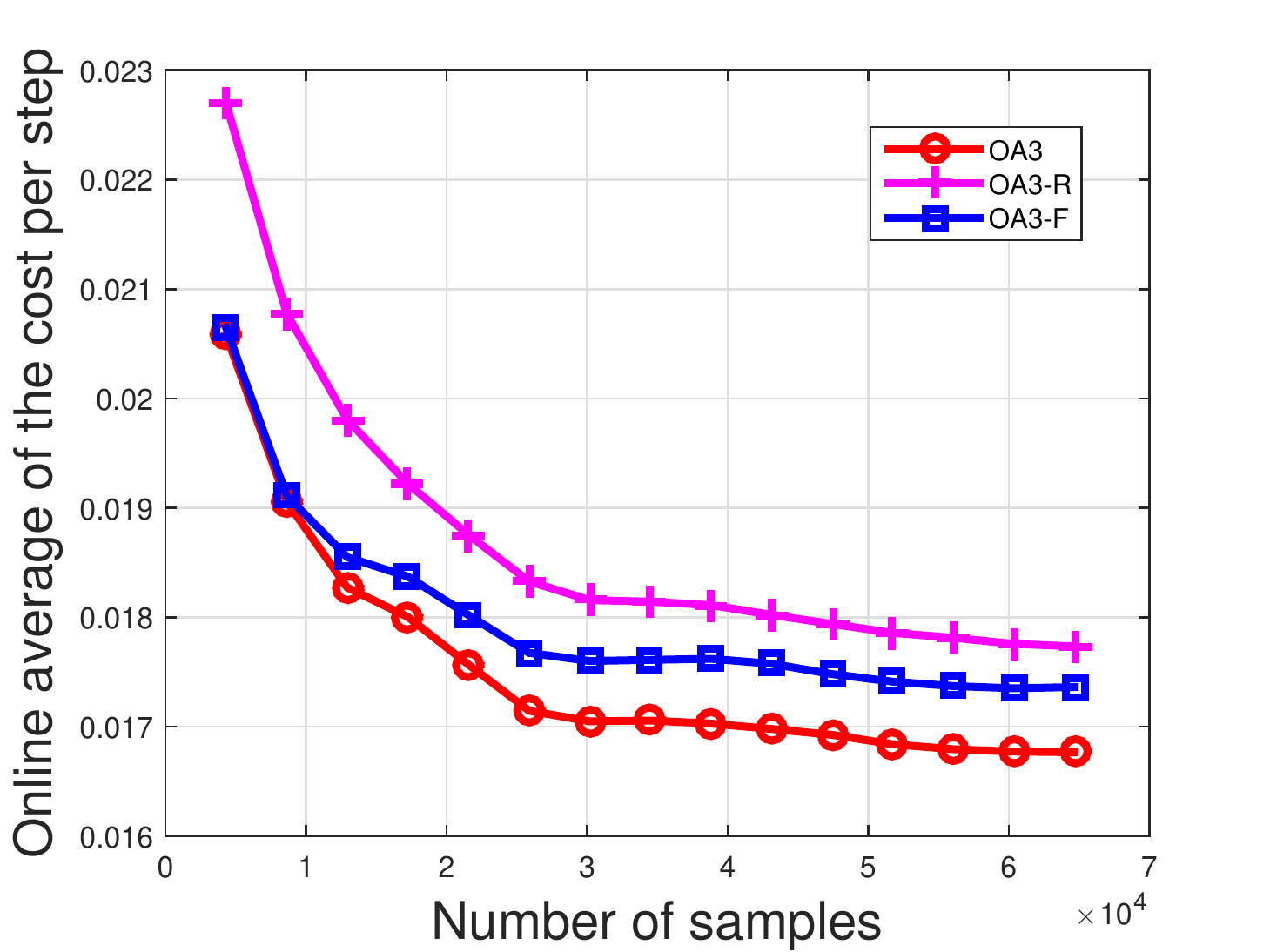}}
      \centerline{(c) w8a, \shuai{B=6500}}
    \end{minipage}
    \hfill
    \begin{minipage}{0.47\linewidth}
      \centerline{\includegraphics[width=4.6cm]{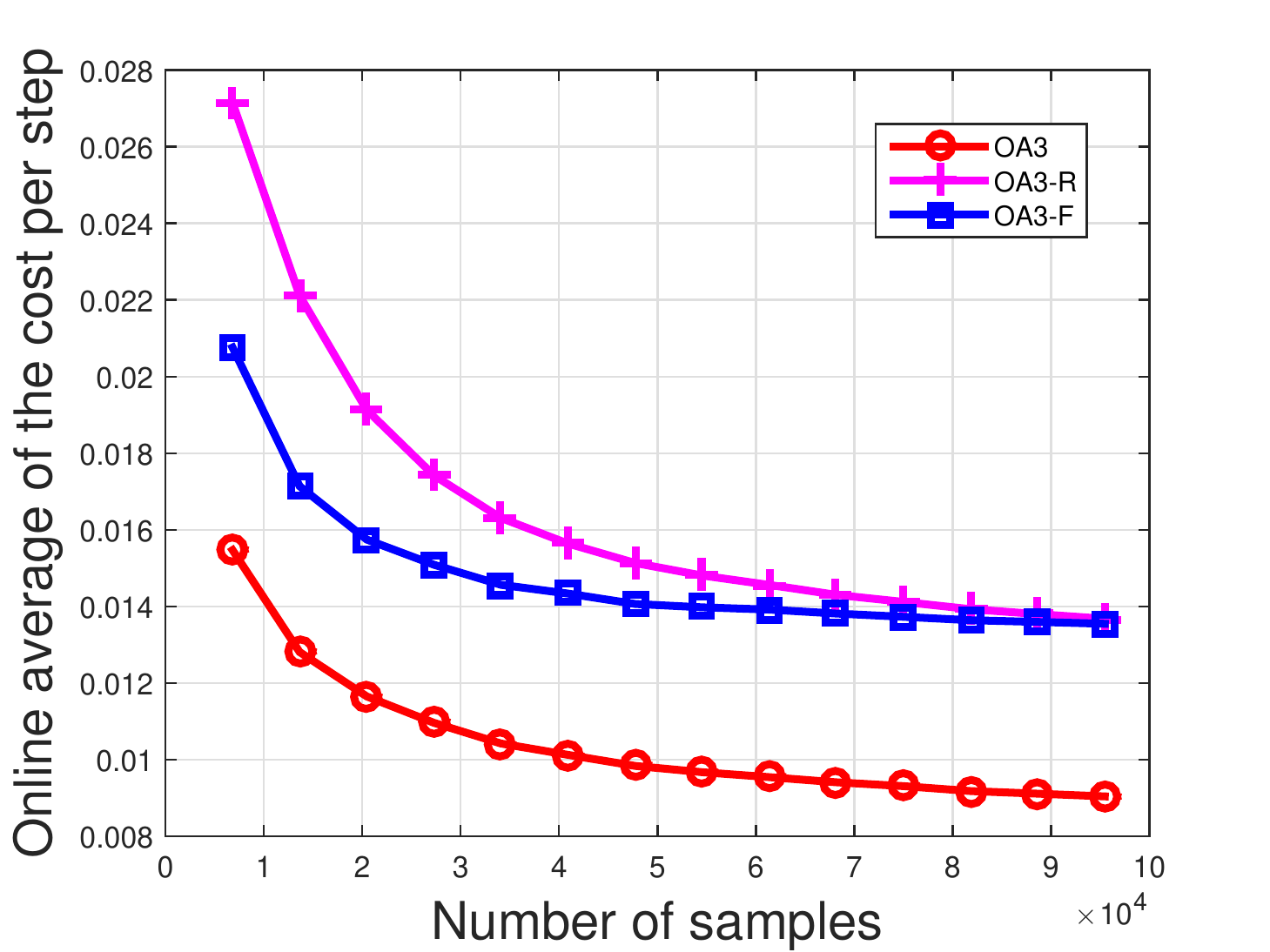}}
      \centerline{(d) KDDCUP08, \shuai{B=10000}}
    \end{minipage}
    \vspace{-0.1in}
    \caption{Cost evaluation of different query strategies.}\label{cost_query_strategy}
    \vspace{-0.15in}
\end{figure}

From the results in Fig.~\ref{sum_query_strategy} and \ref{cost_query_strategy}, our asymmetric query strategy outperforms both OA3-R and OA3-F on most datasets, which verifies the effectiveness of our query strategy. \shuai{Besides, when the number of samples is small, the performance of OA3 and OA3-F are almost same on most datasets. One possible reason is that at the beginning of online learning, each sample is informative and thus is queried by OA3. In this case, OA3 is nearly same to OA3-F.} Note that our query strategy also has theoretical guarantees which we described in Section \ref{theorems}.

\subsection{Evaluation on High Dimensional Datasets}

\begin{figure}
    \begin{minipage}{0.47\linewidth}
      \centerline{\includegraphics[width=4.6cm]{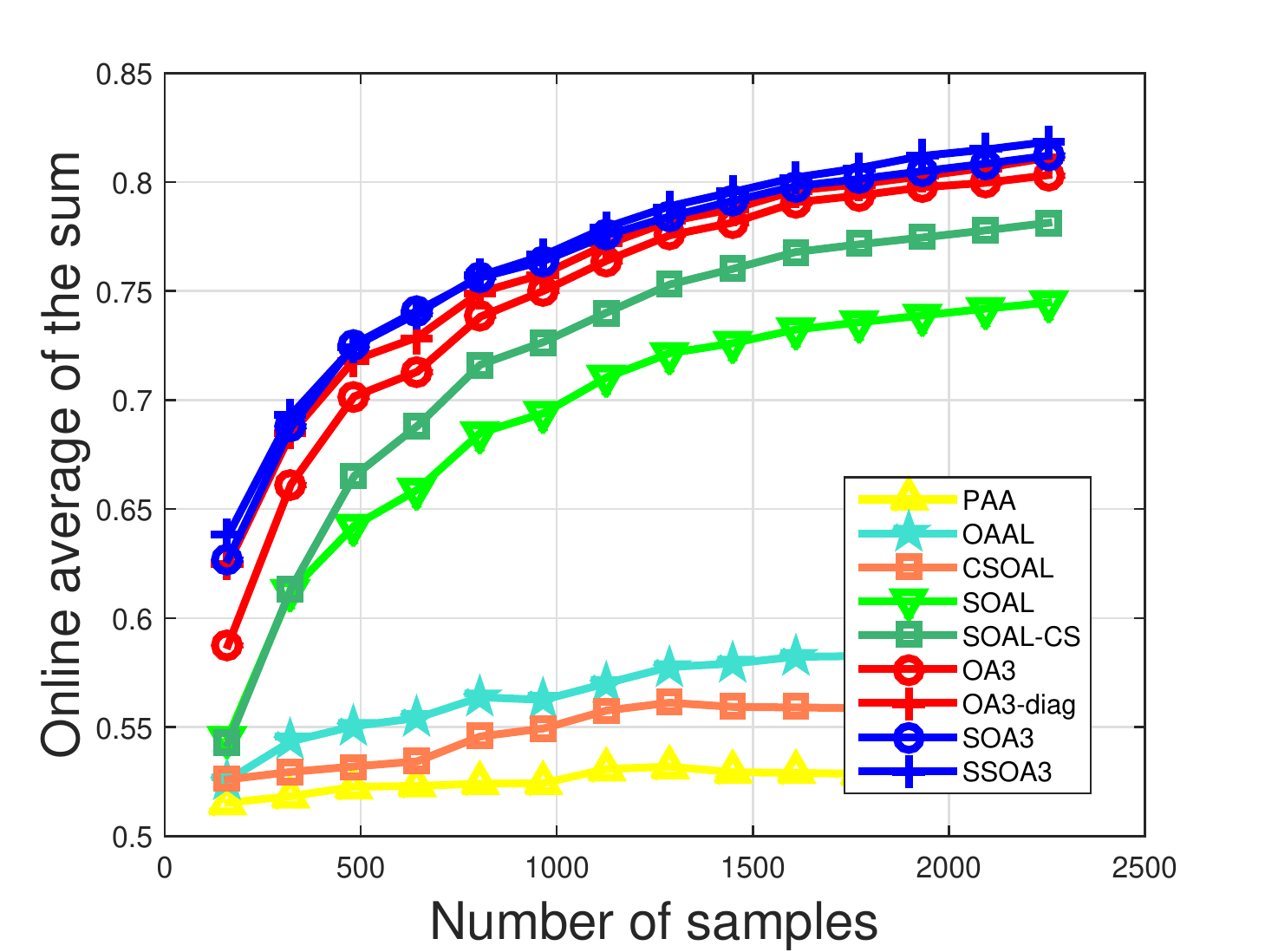}}
      \centerline{(a) sum of IAD, B=1205}
    \end{minipage}
    \hfill
    \begin{minipage}{0.47\linewidth}
      \centerline{\includegraphics[width=4.6cm]{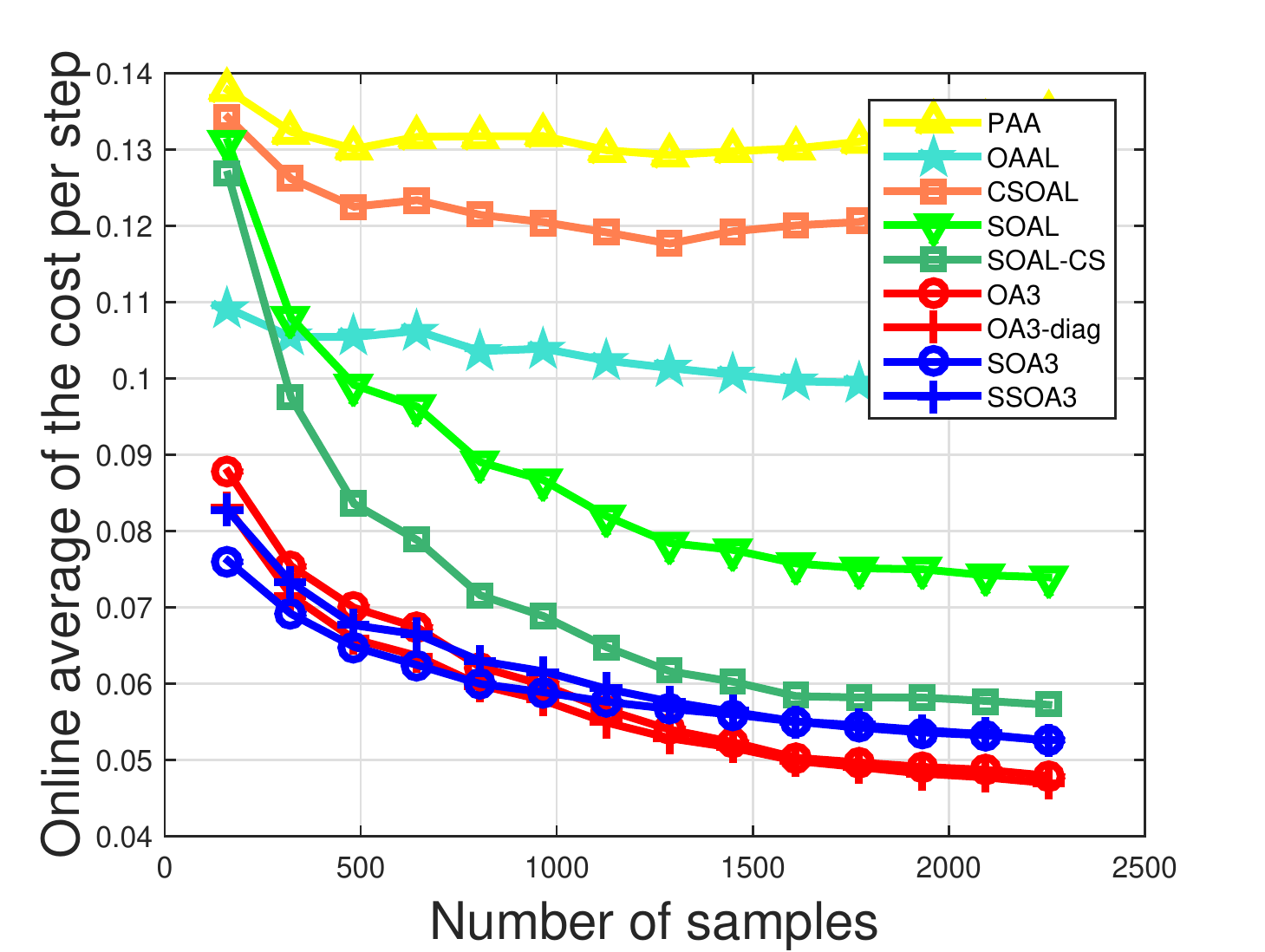}}
      \centerline{(b) cost of IAD, B=1205}
    \end{minipage}
    \vfill
    \begin{minipage}{0.47\linewidth}
      \centerline{\includegraphics[width=4.6cm]{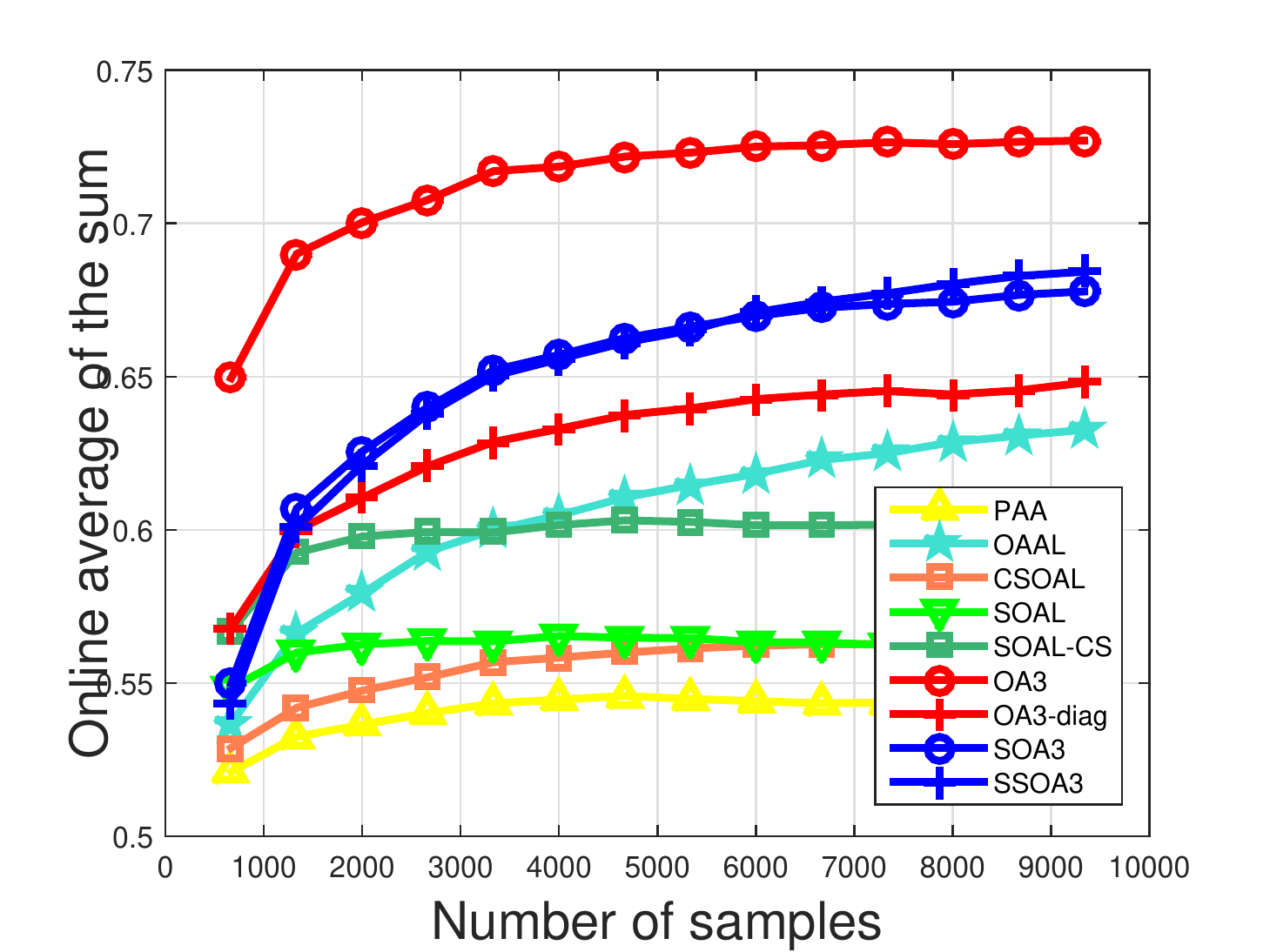}}
      \centerline{(c) sum of CIFAR-10, B=5000}
    \end{minipage}
    \hfill
    \begin{minipage}{0.47\linewidth}
      \centerline{\includegraphics[width=4.6cm]{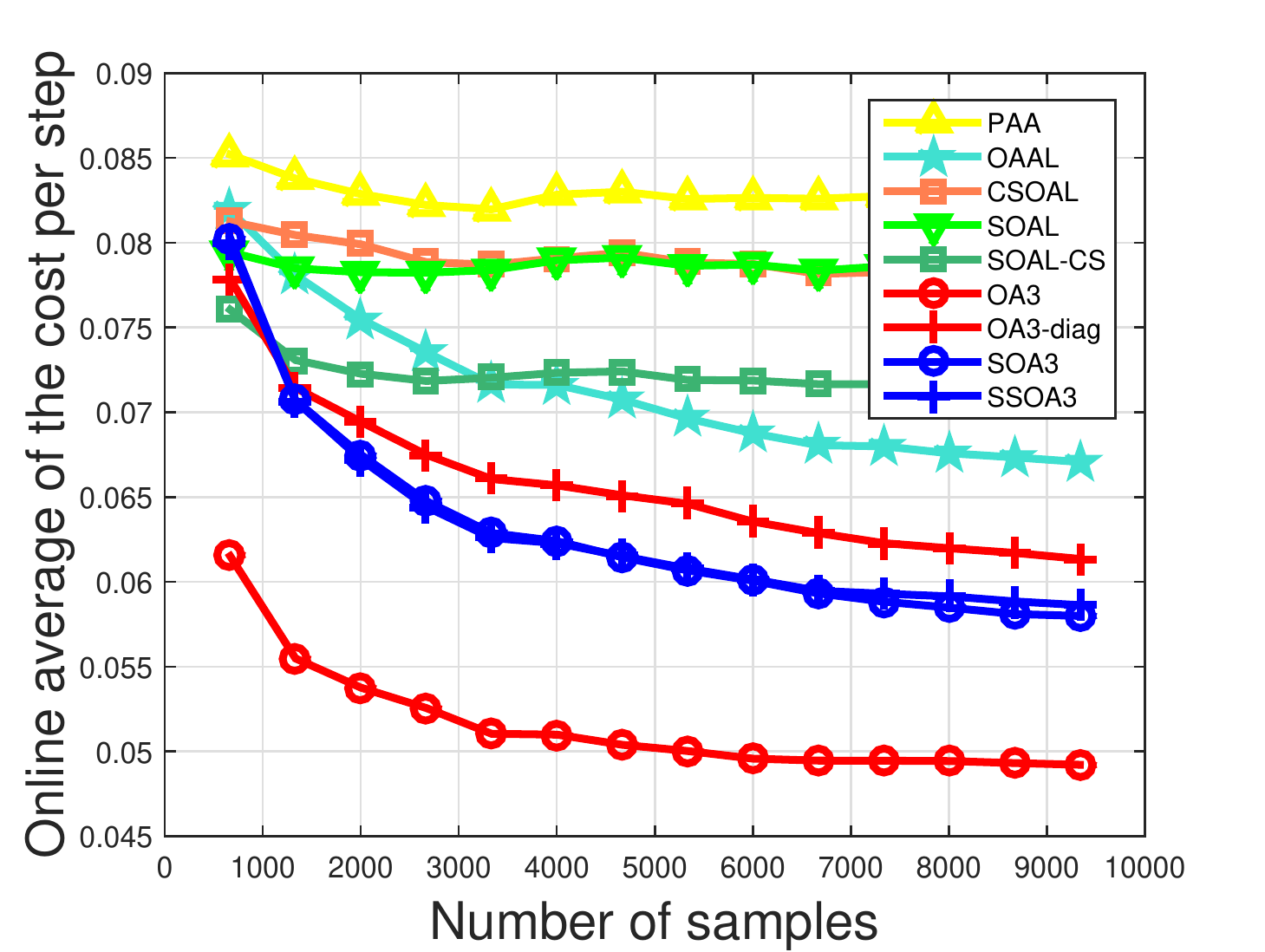}}
      \centerline{(d) cost of CIFAR-10, B=5000}
    \end{minipage}
    \vspace{-0.1in}
    \caption{Evaluation with fixed budget.}\label{sum_cost_high_dimensional}
    \vspace{-0.15in}
\end{figure}

\revise{In the main text, we have reported the results of $sum$ metric on the two higher dimensional datasets. In this subsection, we present the results of $cost$ metric and provide more discussions. Table~\ref{high_dataset} summarizes the statistics of the datasets.  Fig.~\ref{sum_cost_high_dimensional} shows the development of \textit{sum} and \textit{cost} performance. Table~\ref{cost_performance} provides more details in terms of $cost$ metric.}

\revise{From the empirical results, OA3 outperforms all first-order algorithms (PAA, OAAL and CSOAL) and other second-order algorithms (i.e., SOAL and SOAL-CS). However, the computation cost of all second-order methods (i.e., OA3, SOAL and SOAL-CS)  is a huge problem when the dimension of data increases. In contrast, our proposed SOA3 and SSOA3 demonstrate better trade-off between efficiency and performance than OA3. It worth mentioning that, from Table~\ref{speedup_ratio}, our SOA3 and SSOA3 show larger speedup ratios when the dimensionality increase, which further verify the efficiency of the proposed methods.}

\begin{table}[h]
	\caption{\small{Statistics of two high-dimensional datasets}}
	\label{high_dataset}
    \vspace{-0.1in}
    \begin{center}
	\begin{tabular}{|l|c|c|c|}\hline
		Dataset &\#Examples & \#Features & \#Pos:\#Neg  \\\hline \hline
        IAD& 2410 & 1558 & 1:4.9 \\
        CIFAR-10 & 10000 & 3072 & 1:9.0 \\\hline
	\end{tabular}
    \end{center}
\end{table}

\begin{table*}[t]
	\caption{\revise{Cost evaluation on high-dimensional datasets} }\label{cost_performance}
    \begin{center}
    \begin{scriptsize}
    \scalebox{1}{
    \renewcommand*\arraystretch{1.25}
	\begin{tabular}{|l|c|c|c|c|c|c|c|c|}\hline

        \multirow{2}{*}{Algorithm}&\multicolumn{4}{c|}{$``cost"$ on CIFAR-10} &  \multicolumn{4}{c|}{$``cost"$ on IAD}\cr\cline{2-9}
        &Cost&Sensitivity(\%)&Specificity  (\%)&Time(s)   & Cost&Sensitivity(\%)&Specificity  (\%)&Time(s)\cr
        \hline \hline
        PAA   &830.720 	$\pm$ 34.970 	& 17.313 	$\pm$ 15.064 	& 90.793 	$\pm$ 15.308 	& 0.516
        &325.320 	$\pm$ 39.882 	& 19.289 	$\pm$ 19.460 	& 85.539 	$\pm$ 16.220 	& 0.064  \\
        OAAL  &664.890 	$\pm$ 15.590 	& 58.766 	$\pm$ 1.571 	& 67.545 	$\pm$ 0.997 	& 0.472  	
        &241.620 	$\pm$ 7.927 	& 53.333 	$\pm$ 2.513 	& 64.905 	$\pm$ 1.858 	& 0.062  \\
        CSOAL  &778.650 	$\pm$ 58.513 	& 21.124 	$\pm$ 9.960 	& 92.749 	$\pm$ 3.781 	& 0.463
        &301.160 	$\pm$ 48.723 	& 23.088 	$\pm$ 14.965 	& 90.639 	$\pm$ 5.080 	& 0.058  \\
        SOAL  &790.440 	$\pm$ 12.342 	& 18.478 	$\pm$ 1.908 	& \textbf{94.100 	$\pm$ 0.786} 	& 505.121
        &178.790 	$\pm$ 38.081 	& 53.554 	$\pm$ 10.719 	& \textbf{95.884 	$\pm$ 1.130} 	& 21.788  \\
        SOAL-CS  &719.380 	$\pm$ 18.485 	& 27.920 	$\pm$ 2.729 	& 92.505 	$\pm$ 0.861 	& 556.762
        &138.270 	$\pm$ 6.059 	& 68.235 	$\pm$ 1.835 	& 89.196 	$\pm$ 1.654 	& 28.812  \\
        OA3   &\textbf{489.540 	$\pm$ 9.696} 	& \textbf{79.791 	$\pm$ 3.050}  & 65.898 	$\pm$ 3.636 & 1859.225
        &114.680 	$\pm$ 8.430 	& 85.270 	$\pm$ 3.898 	& 69.735 	$\pm$ 5.624 	& 45.005  \\
        OA3$_{diag}$  &607.460 	$\pm$ 39.401 	& 62.169 	$\pm$ 10.385 	& 70.508 	$\pm$ 6.869 	& 105.221
        &\textbf{111.580 	$\pm$ 8.055} 	& 84.118 	$\pm$ 3.506 	& 73.397 	$\pm$ 5.555 	& 6.609  \\
        SOA3  &576.430 	$\pm$ 27.359 	& 71.433 	$\pm$ 8.647 	& 64.643 	$\pm$ 8.730 	& 66.108
        &125.480 	$\pm$ 8.543 	& \textbf{97.745 	$\pm$ 1.032} 	& 41.459 	$\pm$ 4.996 	& 4.409  \\
        SSOA3 &582.720 	$\pm$ 49.878 	& 72.328 	$\pm$ 8.699 	& 63.043 	$\pm$ 10.479 	& 50.397
        &125.160 	$\pm$ 19.941 	& 93.627 	$\pm$ 6.757 	& 49.171 	$\pm$ 20.764 	& 3.177  \\

        \hline
	\end{tabular}}

    \end{scriptsize}
    \end{center}
\end{table*}

\begin{table*}[h]
	\caption{\revise{Speedup ratios of sketched algorithms on different datasets}}
    \label{speedup_ratio}
    \begin{center}
	\begin{tabular}{|l|c|c|c|c|c|c|}\hline
		Time (s)& Sensorless, $d\small{=}48$ & KDDCUP08, $d\small{=}117$ & w8a, $d\small{=}300$& protein,  $d\small{=}357$& IAD, $d\small{=}1558$ & CIFAR-10, $d\small{=}3072$ \\\hline \hline
        OA3 & 0.966 & 7.066 & 14.118 & 12.511 & 44.674 & 1948.292\\
        SOA3 & 0.862 $(\small{\times}1.12)$ & 3.779 $(\small{\times}1.87)$& 4.556 $(\small{\times}3.10)$& 1.395 $(\small{\times}8.97)$& 4.084 $(\small{\times}10.94)$& 66.739 $(\small{\times}29.19)$\\
        SSOA3 & 0.920 $(\small{\times}1.05)$& 3.376 $(\small{\times}2.09)$& 3.398 $(\small{\times}4.15)$& 1.004 $(\small{\times}12.46)$& 3.215 $(\small{\times}13.90)$& 52.126 $(\small{\times}37.38)$\\\hline
	\end{tabular}
    \end{center}
    \vspace{0.15in}
\end{table*}

\section{Related Work}

Online learning has been a hot research topic in machine learning for many years. One pioneering method is Perceptron algorithm \cite{Freund1999Large}, which updates the predictive vector by adding the misclassified sample with a constant weight. Recently, many margin-based methods \cite{Crammer2006Online,wang2014cost} emerged, and show good performance. Despite their superiority, these methods only adopt samples' first-order information, which may result in slow convergence rate. To eliminate this limitation, many second-order methods \cite{Crammer2009Adaptive,zhao2015cost,Cesa-Bianchi2005A,zhao2018cost} were proposed, and significantly improve the performance with faster convergence.

Active learning aims to train a well-performed model by querying the labels of a small subset of informative data. It helps to reduce the labeling cost and attracts wide attention \cite{Aggarwal2014Active,Fujii2016Budgeted,Sheng2008Get,Cesa-Bianchi2006Worst,Fang2012Self-taught}. Particularly, one classic work \cite{Cesa-Bianchi2006Worst} proposes a sampling method by drawing a Bernoulli random variable, which has been validated as effective and inspires many studies \cite{Zhang2016Online, Zhao2013Costd}. Despite the effectiveness, most active algorithms \cite{Abe2006outlier,Chakraborty2015Batchrank,Huang2017Cost,Krishnamurthy2017Active} assume that all the data are provided in advance, which is impractical in real-world problems. To address this limitation, OAL has emerged and raised wide research interest \cite{Lu2016Online,Zliobaite2014Active,Hao2016second,Ferdowsi2013Online}.

Although OAL has been studied widely, only a few studies focus on class-imbalanced issues. One classic method is CSOAL \cite{Zhao2013Costd}, which adopts a cost-sensitive update rule to solve imbalance problems. Another famous method is OAAL \cite{Zhang2016Online}, which uses an asymmetric query rule to handle imbalanced data. However, both pioneering methods only consider the asymmetric strategy from one isolated perspective, which restricts their abilities in imbalance problems. Moreover, both methods only consider first-order information of samples, which limits their performance. To address these limitations, we exploit samples' second-order information and develop a new asymmetric strategy, considering both optimization and label queries, to handle imbalance problems and accelerate convergence rate.

Then, we highlight several differences of OA3 with several conceptually related methods, \textit{i.e.} \cite{Krishnamurthy2017Active,Hao2016second}. In \cite{Krishnamurthy2017Active}, a cost overlapped active learning algorithm is proposed to study the cost problem, but it is not an online method and does not consider class-imbalance issues. A second-order online active learning algorithm (SOAL) is proposed in \cite{Hao2016second}. This method adopts second-order information, but it ignores the budget limitations and class-imbalance problems. A cost-sensitive version of SOAL is presented in \cite{Hao2016second}. This method considers the asymmetric losses, but it lacks theoretical guarantees, and ignores budget limitations and asymmetric query strategies.

Finally, sketching methods are a good choice to balance the performance and efficiency of OA3 algorithms. A well-known family of methods for sketching
is the Random Projection \cite{Vempala2004The, Sarlos2006Improved, Liberty2007Randomized, Achlioptas2003Database, Indyk1998Approximate, Kane2014Sparser}, which relies on the properties of random low dimensional subspaces and strong concentration of measure phenomena. However, the theoretical guarantees of the Random Projection may perform terrible when the rank of approximated matrix is near full-rank \cite{luo2016efficient}. To address the above issue, researchers recently proposed the Frequent Direction Sketch \cite{Ghashami2016Frequent, Liberty2013Simple}, which is a class of deterministic methods that derives from the similarity comparison between matrix sketching problem and item frequency estimation problem. For the Frequent Direction Sketch methods, the regret bound depends on a super-parameter and a square root term, which are not controlled by the sketching algorithms \cite{luo2016efficient}. To better focus on the dominant part of the spectrum for deterministic sketching, an Oja’s sketch algorithm was recently proposed \cite{luo2016efficient} based on Oja’s algorithm \cite{Oja1982Simplified, oja1985stochastic}.

 \revise{A brief version of this paper had been published in SIGKDD conference \cite{zhang2018online}. Compared with it, this journal manuscript makes several significant extensions, including (1) two updated variants with sketching methods, and some theoretical analyses about their time complexity; 
(2) more empirical studies to evaluate the proposed algorithms；.}

\balance
